%% file: main.tex
\documentclass[journal]{IEEEtran}
\usepackage{xcolor}
\colorlet{secondary}{white!25!black}
\usepackage{tabularx, multirow, graphics, url, float, subfigure, amsmath, tikz, hyperref,textcomp}
\hypersetup{
    colorlinks=true,
    citecolor={blue!50!black},
    linkcolor=black,
    urlcolor=cyan,
}
\usepackage[nolist]{acronym}
\usepackage[export]{adjustbox}
\usetikzlibrary{mindmap,trees,fit,decorations.pathreplacing,patterns,decorations.markings}

\newcounter{mynode}
\tikzset{step node/.code={\stepcounter{mynode}
}}

\tikzstyle{pyschologyBar} = [rounded corners=1mm,fill=cyan, draw=white]
\tikzstyle{systemBBar} = [rounded corners=1mm,fill=blue!45!green, draw=white]
\tikzstyle{riseFPBar} = [rounded corners=1mm,fill=blue!40!white, draw=white]
\tikzstyle{risePPBar} = [rounded corners=1mm,fill=red!40!white, draw=white]
\tikzstyle{firstDBBar} = [rounded corners=1mm,fill=orange, draw=white]
\tikzstyle{firstWKBar} = [rounded corners=1mm,fill=green!70!red, draw=white]
\tikzstyle{riseEBar} = [rounded corners=1mm,fill=yellow!90!black, draw=white]
\tikzstyle{educationText} = [rectangle, above=.1cm, align=center,,scale=1.0]
\tikzstyle{experienceText} = [rectangle, below=.1cm, align=center,,scale=1.00]

\makeatletter
\let\NAT@parse\undefined
\makeatother
\usepackage[sort&compress,numbers]{natbib}

\def\checkmark{\tikz\fill[scale=0.33](0,.35) -- (.25,0) -- (1,.7) -- (.25,.15) -- cycle;}

\usepackage[normalem]{ulem}

\begin{document}

\title{Affordances in Robotic Tasks - A Survey}

\author{Paola Ard{\'o}n, \textit{Member IEEE}, {\`E}ric Pairet, \textit{Member IEEE}, Katrin S. Lohan, \textit{Member IEEE}, \\ Subramanian Ramamoorthy, \textit{Member IEEE}, and Ronald P. A. Petrick, \textit{Member IEEE}
\thanks{
\textcopyright 2020 IEEE.  Personal use of this material is permitted.  Permission from IEEE must be obtained for all other uses, in any current or future media, including reprinting/republishing this material for advertising or promotional purposes, creating new collective works, for resale or redistribution to servers or lists, or reuse of any copyrighted component of this work in other works. \textbf{Manuscript currently under review.}

The authors are with the Edinburgh Centre for Robotics. University of Edinburgh and Heriot-Watt University. Edinburgh, UK. \tt \{paola.ardon;eric.pairet;s.ramamoorthy\}@ed.ac.uk; \{k.lohan;r.petrick\}@hw.ac.uk.}

}

 \markboth{IEEE Transaction on Robotics,~Vol.~XX, No.~X, December~2019}%
 {Shell \MakeLowercase{\textit{et al.}}: XXX Survey for IEEE Journals}

\maketitle

\input{acronyms}

\begin{abstract}
Affordances are key attributes of what must be perceived by an autonomous robotic agent in order to effectively interact with novel objects. Historically, the concept derives from the literature in psychology and cognitive science, where affordances are discussed in a way that makes it hard for the definition to be directly transferred to computational specifications useful for robots. This review article is focused specifically on robotics, so we discuss the related literature from this perspective. In this survey, we classify the literature and try to find common ground amongst different approaches with a view to application in robotics. We propose a categorisation based on the level of prior knowledge that is assumed to build the relationship among different affordance components that matter for a particular robotic task. We also identify areas for future improvement and discuss possible directions that are likely to be fruitful in terms of impact on robotics practice.
\end{abstract}


\input{sections/1_introduction.tex}

\input{sections/2_formalism.tex}

\input{sections/3_model.tex}

\input{sections/4_combined_methods.tex}
\input{sections/5_exploratory.tex}

\input{sections/6_limitations.tex}

\input{sections/open_questions.tex}

\footnotesize{
\bibliographystyle{IEEEtranSN}
\bibliography{references.bib}
}

\input{sections/biography.tex}

\end{document}

%% file: acronyms.tex
\begin{acronym}[ransac]
  \acro{LbD}{learning by demonstration}
  \acro{RL}{reinforcement learning}
  \acro{IRL}{inverse reinforcement learning}
  \acro{SVM}{support vector machine}
  \acro{DOF}{degrees-of-freedom}
  \acro{CAD}{computer-aided design}
  \acro{ROI}{regions of interest}
  \acro{MCMC}{Markov Chain Monte Carlo}
  \acro{ECV}{early cognitive vision}
  \acro{IADL}{instrumental activities of daily living}
  \acro{CDR}{cognitive developmental robotics}
  \acro{2-D}{two-dimensional}
  \acro{3-D}{three-dimensional}
  \acro{RANSAC}{Random sample consensus}
  \acro{RGB-D}{red-green-blue depth}
  \acro{IFR}{International Federation of Robotics}
  \acro{CNN}{Convolutional Neural Networks}
  \acro{KB}{Knowledge Base}
  \acro{MSE}{mean square error}
  \acro{GWR}{Geographically Weighted Regression}
  \acro{PCA}{Principal Component analysis}
  \acro{CRF}{Conditional Random Fields}
  \acro{ILGA}{Integrated Learning of Grasps and Affordances}
  \acro{OACs}{object action complexes}
  \acro{HRI}{Human-Robot Interaction}
  \acro{CP}{control program}
\end{acronym}

%% file: sections/1_introduction.tex
\section{Introduction\label{sc:introduction}}

The field of robotic manipulation has enjoyed significant recent success towards scaling up to human-like abilities. However, the gap between human and robot dexterity, especially when it comes to purposeful and temporally extended manipulations, remains large. For instance, consider the familiar scenario of a robot tasked with tidying up a home environment. The robot will find it challenging to naively learn and remember every possible presentation of a diverse collection of objects in such a heterogeneous environment. This makes it hard to achieve a level of autonomy wherein the robotic system can replicate a complex task on previously unseen objects and new environments. Such generalisation is at the very heart of human capabilities. It is argued that concepts like affordances serve the key role of being intermediaries that organise the diversity of possible perceptions into tractable representations that can support reasoning for the purposes of action prediction, manipulation and navigation.

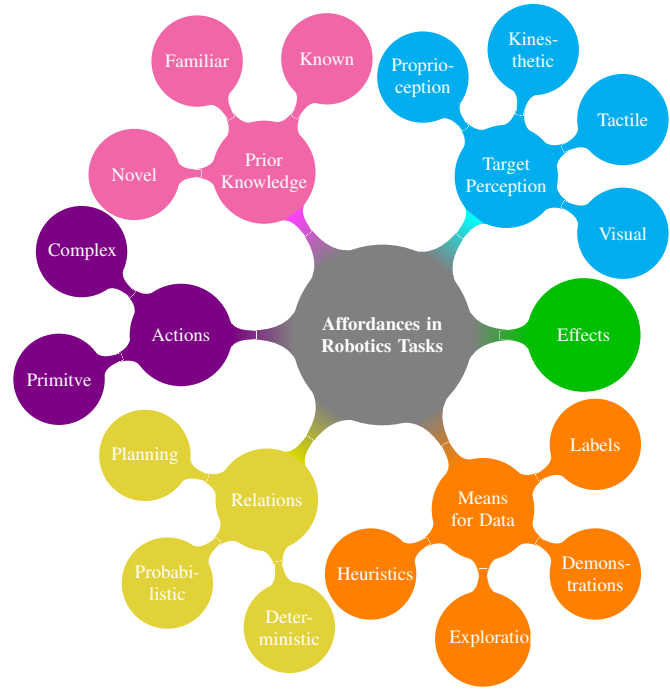
\begin{figure}[t!]
    \centering
    \resizebox{\columnwidth}{!}{
         \begin{tikzpicture}
    	\tikzstyle{every node}=[font=\large]
      \path[mindmap,concept color=gray,text=white]
        node[concept,minimum size=4cm] {\textbf{Affordances in Robotics Tasks}}
        [clockwise from=0]
        child[level distance=4.5cm,concept color=green!75!black] {
          node[concept, minimum size=2.5cm] {Effects}
        } 
        child[level distance=4.5cm, concept color=orange, style={sibling angle=60}] {
          node[concept] {Means for Data}
          [clockwise from=30]
          child { node[concept,minimum size=2cm] {Labels} }
          child { node[concept,minimum size=2cm] {Demons-trations} }
          child { node[concept,minimum size=2.1cm] {Exploration} }
          child { node[concept,minimum size=2cm] {Heuristics} }
        }
        child[level distance=4.5cm,concept color=purple!65!blue, style={sibling angle=90}] {
          node[concept] {Actions}
          [clockwise from=200]
          child { node[concept,minimum size=2cm] {Primitve} }
          child { node[concept,minimum size=2cm] {Complex} }
        }
        child[level distance=4.5cm,concept color=magenta!75!white, style={sibling angle=78}] {
          node[concept] {Prior Knowledge}
          [clockwise from=181]
          child { node[concept,minimum size=2cm] {Novel} }
          child { node[concept,minimum size=2cm] {Familiar} }
          child { node[concept,minimum size=2cm] {Known} }
        }
        child[level distance=4.5cm, concept color=cyan, style={sibling angle=77}] {
          node[concept] {Target Perception}
          [clockwise from=130, level 2 concept/.append style={sibling angle=52}]
          child { node[concept,minimum size=2cm] {Proprio-ception} }
    	  child { node[concept,minimum size=2cm] {Kines-thetic} }
          child { node[concept,minimum size=2cm] {Tactile} }
          child { node[concept,minimum size=2cm] {Visual} }
        }
          child[level distance=4.5cm,concept color=yellow!85!black, style={sibling angle=25}] {
          node[concept] {Relations}
          [clockwise from=-80]
          child { node[concept,minimum size=2cm] {Deter-ministic} }
          child { node[concept,minimum size=2cm] {Probabi-listic} }
          child { node[concept,minimum size=2cm] {Planning} }
        }
        ;
    \end{tikzpicture}
    }
    \caption{The main aspects that influence a robotics task that includes affordances, with methodologies classified according to their generalisation capabilities. This classification is based on the use of prior knowledge that relates \textit{target object}, \textit{action}, and \textit{effects}. Other aspects are also identified, such as how these relationships are built and the means of collecting data given that some methods rely on heuristics to directly relate these elements or on learning models from data collected offline.}
    \label{fig:intro}
\end{figure}


    \subsection{Background}
        The psychologist James J. Gibson, along with Eleanor J. Gibson, introduced the concept of an {\textit{affordance}} \cite{gibson2014ecological}. According to Gibsons' theory, an affordance refers to the ability to perform a certain action with an object in a given environment. This definition attracted controversy among psychologists who attempted to narrow down this concept by defining boundaries on the perception and action elements. For example, \citeauthor{mcgrenere2000affordances} suggested that an object's affordance exists independently of the individual's ability to perceive its possibility of action~\cite{mcgrenere2000affordances}. 
        
        When the concept is transferred to use in a robot, the focus of attention shifts. One is less interested in the intrinsic worth of whether action and perception are to be viewed as jointly essential, and somewhat more interested in whether any such definition can be operationalised. So, to a roboticist, it may be that the agent must rely on some form of processing of sensory inputs in order to identify the object's affordances.
        
        Given such variations in interpretation of the underlying concept, D. Normal in 1988 proposed a definition of affordance by targeting both the actual and perceived properties \cite{norman1988psychology}. Consequently, in Norman's conceptualisation, the affordance of a ball is both: its round shape, material, bounceability among other physical properties, as well as the perceived suggestion as to how the ball should be used~\cite{norman1988psychology}. It was not until 1991 that \citeauthor{gaver1991technology} redefined affordances with the purpose of using it in technology modelling~\cite{gaver1991technology}.
        \citeauthor{gaver1991technology} divided the affordance concept into three categories that reflect the information as perceived by the agent: (i)~\emph{false affordance} which is a perceived affordance that does not have a real function (e.g. a placebo button), (ii)~\emph{hidden affordances} which are those that are not evident for the agent but they exist, and (iii)~\emph{perceptible affordance} which are the ones where the information is available such that the agent can perceive them and act on them \cite{gaver1991technology}. 
        Given the majority of robotic agents depend on their sensors to perceive the environment, \citeauthor{gaver1991technology}'s definition appears best suited to the needs of robotic manipulation.

        Previous reviews such as \citet{chemero2007gibsonian} and \citet{csahin2007afford} present summaries of the different formalisms that attempt to build a bridge between such a controversial concept and mathematical representations. \citeauthor{horton2012affordances} describe the influence of the Gibsonian theory on the physical design of robotic agents \cite{horton2012affordances}. Along the same lines, \citeauthor{zech2017computational} presents an overview of computational models of affordances and discuss how closely relevant they are to the original Gibsonian theory \cite{zech2017computational}. In 2016, \citeauthor{min2016affordance} reviewed the works that implement the affordance concept in developmental robotics, relating the methodologies to how infants learn \cite{min2016affordance}. Also from a developmental point of view, \citeauthor{jamone2018affordances} reviewed the affordance concept across the fields of psychology, neuroscience and robotics \cite{jamone2018affordances}. In contrast to the existing works summarising the affordance literature, in this survey, we identify and discuss different aspects that influence a robotics task requiring nontrivial use of the concept of affordances.

    \subsection{Relevance and outline}
      
        Given that the discussions within the extant literature have been centred on psychological abstractions of the affordance concept, there is not a clear scheme that outlines \textit{what is essential to accomplish a robotics task through the use of affordances?} and \textit{what is needed to evaluate such tasks?} 
        In contrast to earlier reviews on similar topics, mentioned above, our interest in this survey is to clearly outline the components that are being used in tasks that include affordances, with the purpose of serving as a guide for reproducibility in robotics. Regardless of the abstraction of the concept, using affordances in robotics tasks usually refers to the problem of perceiving a \textit{target object}, identifying what \textit{action} is feasible with it and the \textit{effect} of applying this action to assess if the task is replicable. The optimal goal of learning the relation among target object, actions and effects in robotics is to achieve human-like generalisation capabilities. This relation from now on will be referred to as the \textit{affordance relation} and will represent the primary organising principle for structuring this survey. Fig.~\ref{fig:intro} illustrates a mind map of the elements that help create this affordance relation in a robotics task and their influence is detailed in Section~\ref{sc:motivation}. We propose to group different approaches to the affordances concept based on what they assume is known \textit{a priori} about this relation. It is important to note that some of these works might fall into more than one category, therefore, we consider their main contribution to the field as an indicator as to which o. We group the approaches as follows:
  
  \begin{itemize}
      \item Known affordance relation: These approaches assume that they have seen the affordance relation before and thus it exists as a form of a template, i.e., full \textit{a priori} knowledge. In this case, the system has access to a database that is built offline such that once the target object is recognised the goal is to search for the previously modelled action and effects relation to accomplish a task in a given application. Works in this category are summarised in Section~\ref{sc:model}.
      
      \item Familiar affordance relation: These approaches assume that they have seen a similar, thus familiar affordance relation before, i.e., partial \textit{a priori} knowledge, which is stored in an offline database. These relations can be familiar on different levels. We will see throughout this survey that the most common way of considering an affordance as familiar is when the approaches generalise among object features (shape, colour, texture etc) rather than categories. Thus, objects with similar features would afford similar actions and effects relations. Works in this category are summarised in Section~\ref{sc:combined}.
      
      \item Novel affordance relation: These approaches do not assume they have access to a database that has information about the affordance relation, i.e., no \textit{a priori} knowledge. Methods in this category explore based on heuristic rules on the perceived target object to generate a ranking for corresponding actions and effects. Works in this category are summarised in Section~\ref{sc:exploratory}.
  \end{itemize}
  
    To understand the progression of the field, Section~\ref{sc:formalism} in this survey discusses the historical evolution of the different mathematical definitions that have tried to formalise the inclusion of affordance in robotics tasks. We also identify the gaps existing in the field and thoroughly discuss them in Section~\ref{sc:limitations}, alongside a guide of the available datasets that include the affordance relation. 
  The main goal of this survey is to serve as a guide to help researchers interested in including the concept of affordance in their robotics tasks and identify the clear gaps where the field would benefit from contributions. To guide and facilitate the search we make available an online library\footnote{\url{https://paolaardon.github.io/affordance_in_robotic_tasks_survey/}} with the content and classification of this survey, which will be updated in a bi-annual basis to keep it updated with state-of-the-art literature.
  
\section{Connotation of using affordances in robotics \label{sc:motivation}}
 Incorporating affordances in robotics tasks has been a particularly slow-moving-forward field of study. One of the main causes is the abstraction of the concept. As it is going to be presented in Section~\ref{sc:formalism}, there are multiple mathematical formalisms which attempt to replicate different psychological theories. Finding a suitable affordance relation that generalises across objects and environments is a challenging problem that has been of great interest to the robotics community. Methods using affordance as part of their task show that the inclusion of the concept improves the agent's performance on applications such as navigation \cite{hermans2013decoupling,ugur2015staged,sun2010learning}, action prediction for collaborative tasks \cite{zhu2014reasoning,castellini2011using,aldoma2012supervised} and manipulation \cite{fallon2015architecture,antunes2016human,stark2008functional}. Nonetheless, these methods do not reach the desired level of generalisation capability and, thus, this remains an open area of research for robotics tasks.
  
   In an attempt to guide future contributions in the field, we pinpoint a number of factors that influence a robotics task that incorporates the affordances concept, as shown in Fig.~\ref{fig:intro}. In this summary, as in the mind map illustration, we focus on identifying how the affordance elements \textit{relate} to each other. Namely, the relation can be built as follows: (i)~a probabilistic approach, where there is more than one possible outcome, (ii)~a deterministic one, where the methods offer an outcome, and (iii)~a planning approach, where the relation is built as a task planning problem thus allowing the system to do multi-step predictions on different affordance relations. 
  
  The way in which the works collect data is also considered. There exist methods that rely on labels from images, post-labelling from demonstrations and the system experience as well as on heuristic functions to then learn a model.
  
  The \textit{perception of the target object} is also essential, and refers to the sensory input medium to recognise all the physical and visual qualities that suggest a set of actions in the scene. For example, a handle in a mug contains the visual and physical features that suggest the affordance `to be grasped'. They can be perceived with a different set of sensors. An extensive summary of perception interaction is presented in \citet{bohg2017interactive}. For the purposes of this survey we consider \textit{visual, tactile, kinesthetic} (motion) and \textit{proprioception} (torque and pressure) sensing.
  
  The \textit{actions} can be those that the object affords such as a ball is `rollable' and those that can be applied to the object to achieve an action, such as `pushing' the ball so it rolls. For the latter case, \citet{asada2009cognitive} summarises works that imitate the cognitive development of an infant, dividing it into $12$ stages according to the difficulty of the motions. In this survey, we group those $12$ stages in two sets: \textit{primitive} actions such as reach or poke an object, or \textit{complex} ones such as pouring from one object to another, thus focusing on what the agent could do with the object.
  
  The outcome or \textit{effect} of this exerted action on the object serves to evaluate the level of success of the task. Some of the most common application tasks where the affordances concept has been used are navigation for wheeled or locomotion systems, action prediction for collaborative tasks and manipulation.
  
  Even though the field lacks a common understanding of the meaning of affordances in robotics, we will see throughout this survey that all the works in the literature include at least the affordance relation elements with a variation on the level of prior knowledge on how this relation is built. Building this affordance relation has proven to achieve better generalisation capabilities in different robotic applications. Thus, our objective is to establish the non-written common ground, to provide a clear guide for replicability in robotics tasks that include affordances and ease the contribution of new research in the field.

%% file: sections/2_formalism.tex
\section{Affordances from abstraction to formalisms \label{sc:formalism}}

  \begin{figure*}[tbh!]
     \centering
     \subfigure[\citet{kruger2011object} ]{\label{fig:kruger_d} \includegraphics[width=3.5cm,trim = 0cm -0.50cm 0cm 0cm]{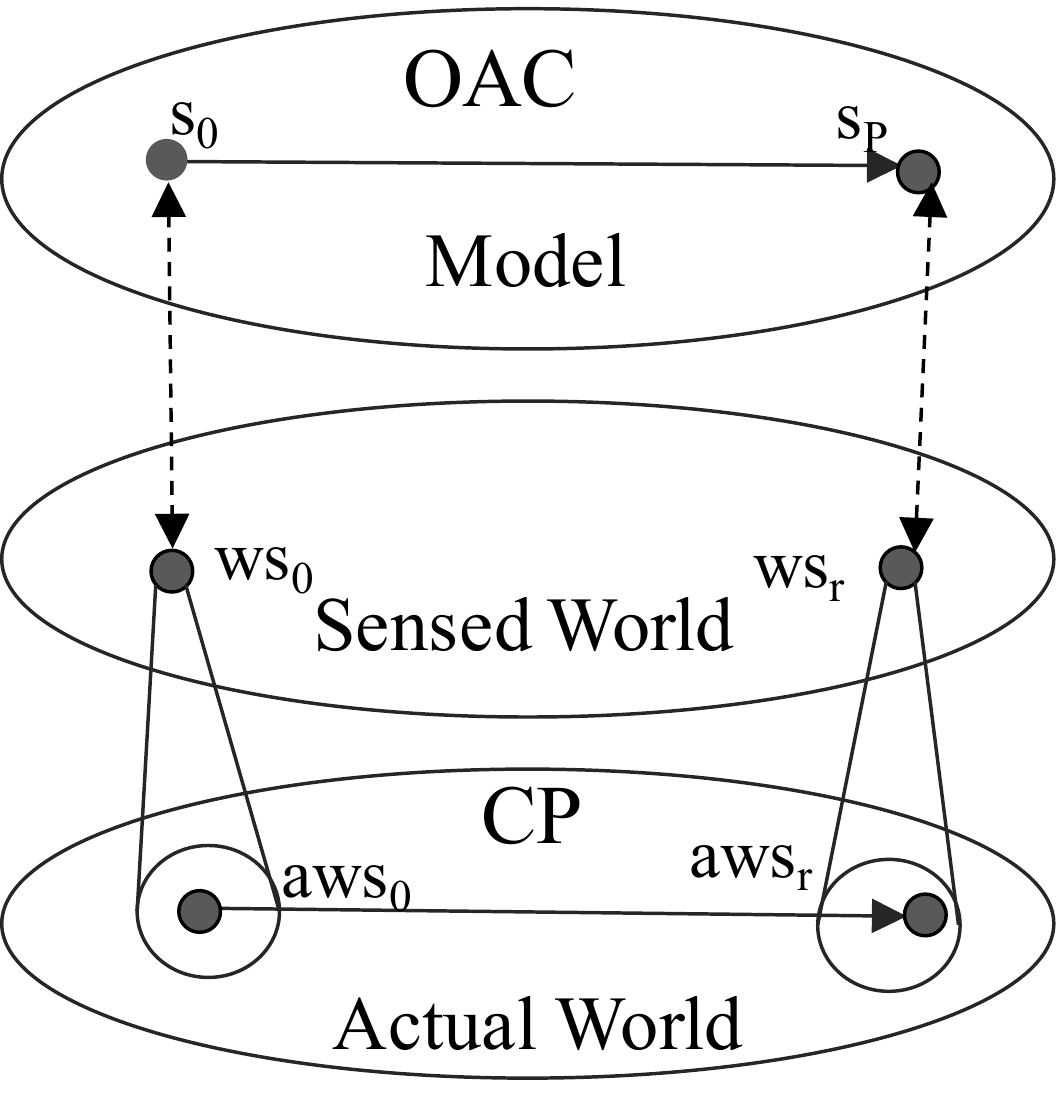}}
      \subfigure[\citet{montesano2007affordances} ]{\label{fig:montesano_d} \includegraphics[width=3.5cm,trim = 0cm -3cm 0cm 0cm]{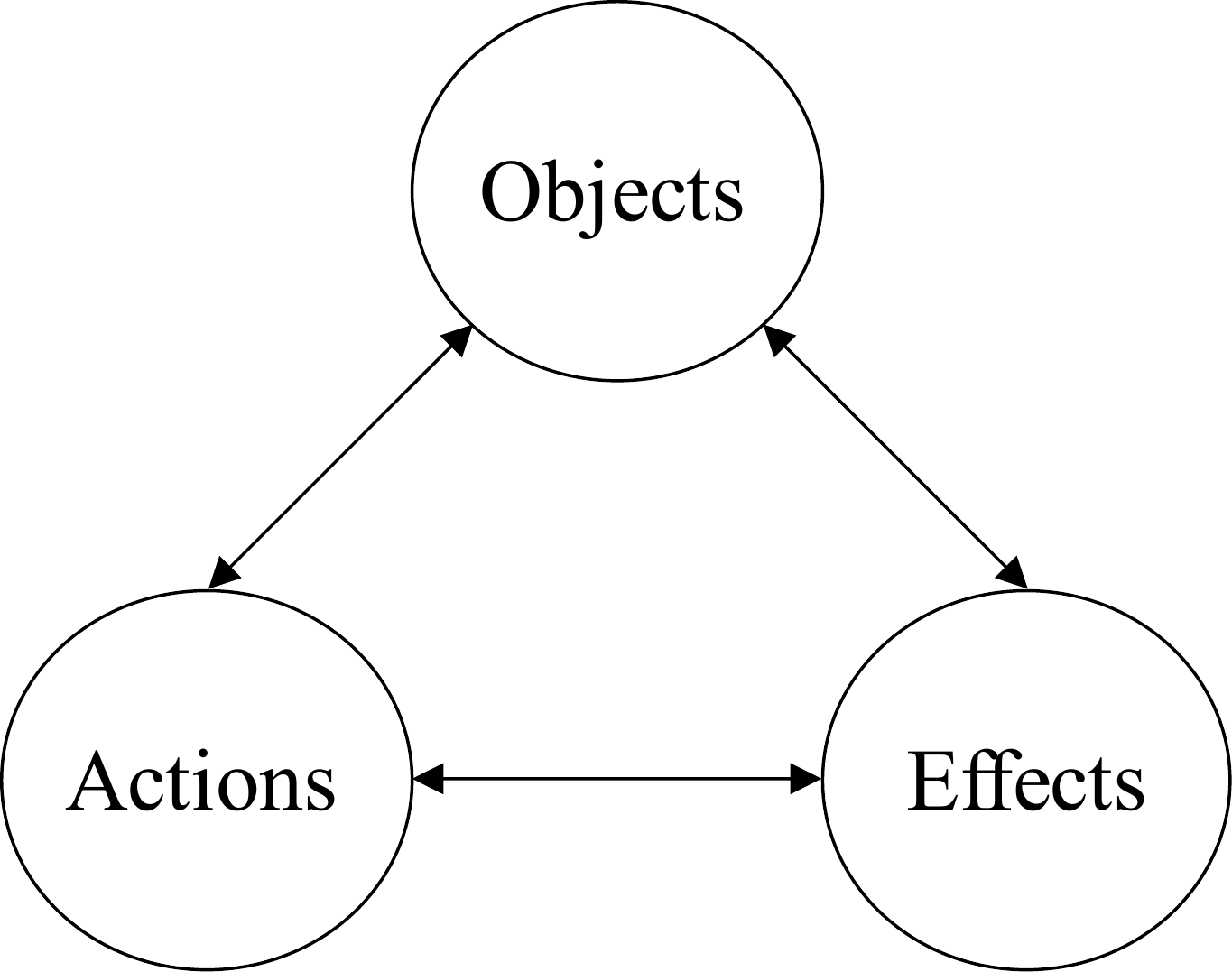}} 
      \subfigure[\citet{cruz2016training} ]{\label{fig:cruz_d} \includegraphics[width=3.3cm,trim = 0cm -3cm 0cm 0cm]{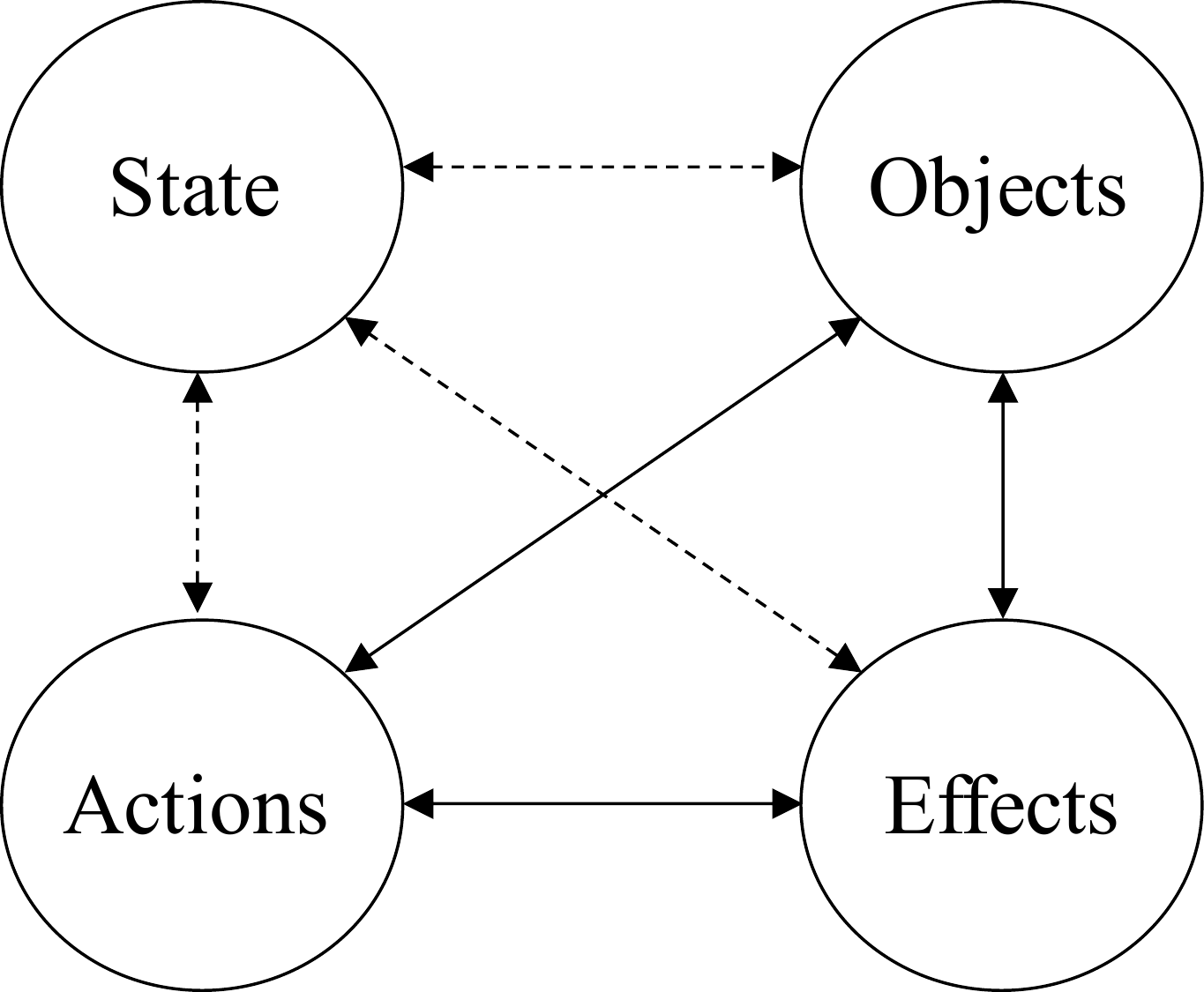}}
      \subfigure[\citet{barck2009learning} ]{\label{fig:barck_d} \includegraphics[width=7cm]{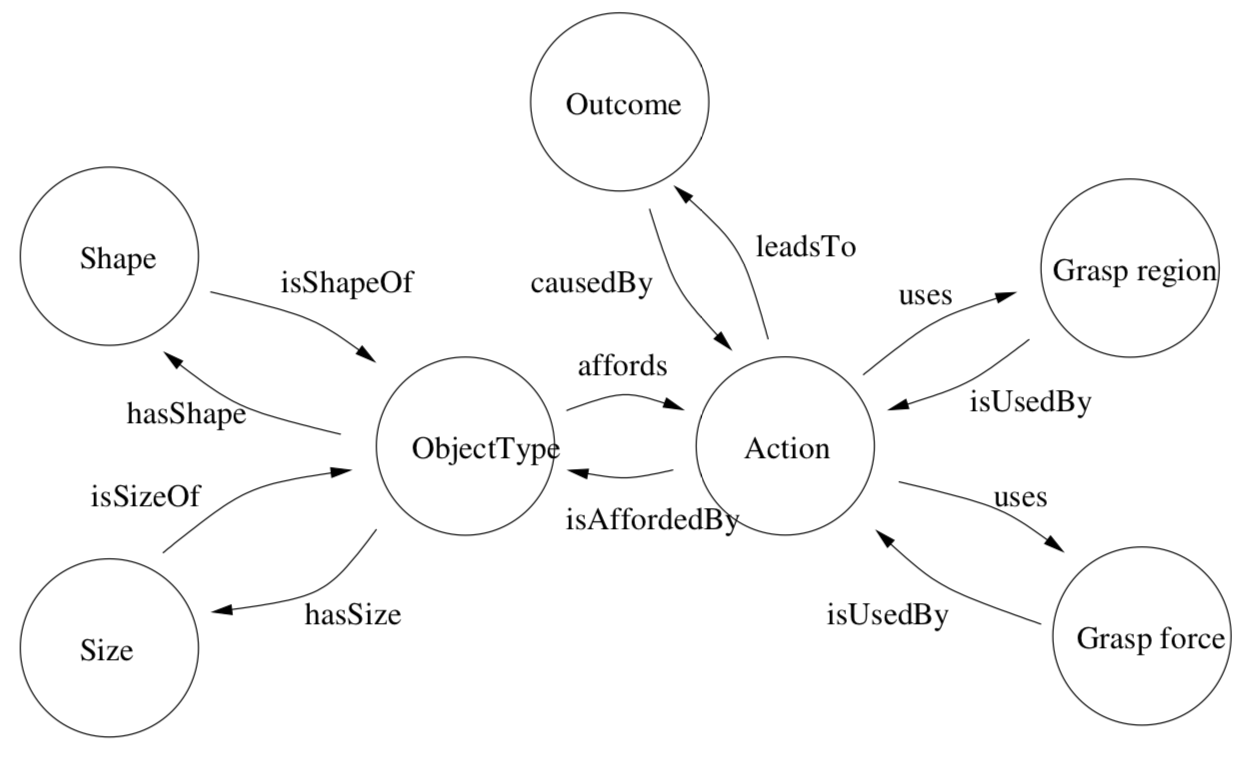}}

     \caption{Approaches that propose a mathematical formalism to the robotics affordance relation. a) \citet{kruger2011object} propose \ac{OACs} as the relationship between the sensed and actual world ($s_0$ is the initial state and $s_p$ is the predicted state). b) \citet{montesano2007modeling} represent affordances as a relation between objects, actions, and effects. Objects are entities which the agent is able to interact with, actions represent the behaviour that can be performed with the object, and effects are the results caused by applying an action. c) \citet{cruz2016training} represent the relations between state, objects, actions and effects, where the state is the current agent's condition and different effects could be produced for different occasions. d) \citet{barck2009learning} present an ontology affordance formalism for grasping. It is a composition of the three basic elements alongside the object properties and grasping criteria.}
     \label{fig:formalisms}
  \end{figure*}

    Historically, in the robotics field, the progression of the study of affordances is marked in two stages. The first stage corresponds to those works that try to mathematically conceptualise affordance in robotics as an extension from psychology theories. The second stage corresponds to those works that propose formalisms focusing on the capabilities of the artificial system rather than recreations from the psychology field.
    
    In spite of the differences in approaching the problem, for both stages, the purpose remains to achieve human-like generalisation performance. Thus, the notorious efforts in the field to translate from human psychology theories to robotics. Interestingly, works across both stages build the affordance relation using the same previously discussed three elements, with equivalents for the target object such as person, environment, entity and features and for the actions variations such as ability or behaviour. 
    \subsection{Early stage formalisms of affordance in robotics}
    
    In the early stages of the field, rather than focusing on perception, the literature focused on perspective. Namely, these works emphasised \textit{how} and \textit{where} the affordance resided following psychology theories. \citet{chemero2007gibsonian} and \citet{csahin2007afford} present a summary of the works using affordances in robotics up to 2007. In these works the discussion of how the different psychology perspectives can be translated into the robotics field is significant. \citeauthor{csahin2007afford} classified the early literature in affordance into three different parallel perspectives \cite{csahin2007afford}: 
    \begin{itemize}
      \item Agent perspective: The concept of affordance resides inside the agent's possibilities to interact with the environment.
      \item Environmental perspective: This concept includes the perceived and hidden affordances in the environment. This is the most abstract of the perspectives.
      \item Observer perspective: This is when the interaction is observed by a third party to further learn these affordances.
    \end{itemize}
    
   The works discussed in \cite{chemero2007gibsonian,csahin2007afford} are summarised in Table~\ref{tb:formalisms}. In these works, the discrepancy in the used terminology is noticeable. Moreover, all the works include the most philosophical perspective of affordances, the environmental one, except for \citet{csahin2007afford} that focused the formalism on the agent perspective.

    Contrary to these early stage formalisms, the current methodology including affordances in robotic application tasks considers the agent and observer perspective. Thus, highlighting the importance of the perception and then action routine rather than the conceptualisation of abstract hidden affordances.

    \begin{table}[t!]
        \centering
        \begin{tabular}{p{1.7cm}|p{3.3cm}| p{2.5cm}}
        \hline
        \textbf{Work} & \multicolumn{2}{c}{\textbf{Formalism of Affordance in Robotics}} \\ \hline
       
        \citet{turvey1992affordances} &  In this example, the stairs that hint are climbable and the effectiveness of the agent to climb them creates a person-climbing system. The affordance is explicitly attached to the environment. & \multirow{2}{*}{\centering \parbox{2.5cm}{\vspace{1.5cm}\textbf{Example}: \\ consider a person climbing the stairs as a person-climbing-stairs system.}} \\ \cline{1-2}
        
        \mbox{\citet{stoffregen2003affordances}} & In this example, the properties of the agent and system are co-defining. The affordance is then considered as agent-environment system rather than just on the environment.  & \\ \hline
        
        \citet{chemero2003outline} & \multicolumn{2}{p{6cm}}{ It is the first approach that sees affordances as relations between abilities of the agent and features of the environment rather than properties. \mbox{Affords-$\phi$} (feature, ability), where $\phi$ is the afforded behaviour.}  \\\hline
        
        \citet{steedman2002formalizing} & \multicolumn{2}{p{6cm}}{As an association of actions with a particular object and what consequences these actions yield. It is the first one to define an \textit{affordance-set}. It also considers the environmental perspective to associate actions.}  \\ \hline 
        
        \mbox{\citet{csahin2007afford}} & \multicolumn{2}{p{6cm}}{As an acquired relation between an \textit{effect} and an \textit{(entity, behaviour)} tuple such that when the agent applies the behaviour on the \textit{entity}, the \textit{effect} is generated: (\textit{effect}, (\textit{entity, behaviour})). First approach to focus only on the agent perspective.} \\ \hline 
        \end{tabular}
        \caption{Summary of early stage formalisms of affordances in robotics tasks.\label{tb:formalisms}}
    \end{table}
    \subsection{Current formalisms of affordance in robotics}
    
Newer proposals such as \citet{montesano2007modeling} and \citet{kruger2011object} define affordances in robotics as using symbolic representations obtained from sensory-motor experience and behaviours. In \cite{kruger2011object} they define an affordance task as a combination of \ac{OACs}. \ac{OACs} capture the interactions of the agent with the world via a low-level \ac{CP}. Fig.~\ref{fig:kruger_d} illustrates the \ac{OACs} formalism with the behaviour of a \ac{CP} functioning in the real world to move an agent to perform an affordance task. The general idea is that the \ac{CP} causes changes in the actual world that transform the actual initial state of the world denoted by $aws_0$ (and sensed by the agent as $ws_0$), to the resulting actual world state (sensed by the agent as $ws_r$). In \cite{montesano2007affordances}, \citeauthor{montesano2007affordances} define affordance in robotics as a formalisation that captures the essential world and object properties, in terms of the actions the robot can perform with the perceived information. This model is shown in Fig~\ref{fig:montesano_d}, where the relation is one that can be used to predict the effects of planned actions to achieve a specific goal, or to select an object to produce a certain effect if acted upon in a certain way. Thus the elements in Fig~\ref{fig:montesano_d} are co-defining. 

A similar approach to the one presented by \citet{montesano2007affordances} is \citet{cruz2016training}. \citeauthor{cruz2016training} consider that if an affordance exists and the agent has knowledge and awareness of it, the next step is to determine if it is possible to utilise it given the agent's current state. For example, a cup affords grasping, as does a die, but in case we have an agent with one die in each hand, then the agent cannot grasp the cup anymore \cite{cruz2016training}. In other words, the affordance is temporarily unavailable. This model is shown in Fig.~\ref{fig:cruz_d}.

\citet{barck2009learning} base their formalism on ontological and probabilistic approaches. An ontological approach in robotics is commonly referred to as a reasoning engine used for learning where important properties of the task are stored with specific definitions. \citet{barck2009learning} examine how ontological approaches can be used to model affordances and how they may produce more successful outcomes when performing an affordance task. Among others, they use object and action properties to model an ontology as shown in Fig.~\ref{fig:barck_d}.

\begin{figure*}
    \centering
    \resizebox{2\columnwidth}{!}{
        \begin{tikzpicture}
        \shade[rounded corners=0cm, top color=gray!20,bottom color=gray!80] (0,5) -- (0,5.80) -- (25.5,5.80) -- (27,5.45) -- (27,5.35) -- (25.5,5.00);
        
        \draw (1,5.40) node[circle,fill=secondary!45!white,draw]{2007};
        \draw (3,5.40) node[circle,fill=secondary!45!white,draw]{2008};
        \draw (5,5.40) node[circle,fill=secondary!45!white,draw]{2009};
        \draw (7,5.40) node[circle,fill=secondary!45!white,draw]{2010};
        \draw (9,5.40) node[circle,fill=secondary!45!white,draw]{2011};
        \draw (11,5.40) node[circle,fill=secondary!45!white,draw]{2012};
        \draw (13,5.40) node[circle,fill=secondary!45!white,draw]{2013};
        \draw (15,5.40) node[circle,fill=secondary!45!white,draw]{2014};
        \draw (17,5.40) node[circle,fill=secondary!45!white,draw]{2015};
        \draw (19,5.40) node[circle,fill=secondary!45!white,draw]{2016};
        \draw (21,5.40) node[circle,fill=secondary!45!white,draw]{2017};
        \draw (23,5.40) node[circle,fill=secondary!45!white,draw]{2018};
        \draw (25,5.40) node[circle,fill=secondary!45!white,draw]{2019};

        \filldraw[pyschologyBar] (0,6.20) rectangle (1.66,6.40);
        \node[educationText] at (0.8,6.20){Popularity of\\psychology- \\ based \\ formalisms};
        
        \filldraw[systemBBar] (1.85,6.20) rectangle (9.00,6.40);
        \node[educationText] at (5.28,6.20){Popularity of system-based formalisms};
        
        \filldraw[riseFPBar] (4,7.60) rectangle (6.8,7.80);
        \node[educationText] at (5.2,7.60){Rise of the field\\with full prior knowledge literature};
        
        \filldraw[riseFPBar] (13,7.60) rectangle (17.8,7.80);
        \node[educationText] at (15.5,7.60){Full prior knowledge\\ works gain popularity};
        
        \filldraw[firstDBBar] (12.2,6.20) rectangle (13.8,6.40);
        \node[educationText] at (13,6.20){First available\\ dataset};
        
        \filldraw[riseEBar] (16.2,6.20) rectangle (21.8,6.40);
        \node[educationText] at (19,6.20){Rise of works with no prior knowledge};
        
        \filldraw[firstDBBar] (18,7.60) rectangle (25.8,7.80);
        \node[educationText] at (21.8,7.60){Five more datasets};
        
        \filldraw[risePPBar] (7,4.40) rectangle (17.2,4.60);
        \node[experienceText] at (11.875,4.60){Rise of works exploiting familiar affordance relations};
        
        \filldraw[firstWKBar] (15,3.20) rectangle (16.70,3.40);
        \node[experienceText] at (16,3.40)%
        {Proposal of first \\ workshop in the field};
        
         \filldraw[firstWKBar] (22.7,4.40) rectangle (25.70,4.60);
        \node[experienceText] at (24.3,4.60)%
        {Organisation of two\\ more workshops};
    \end{tikzpicture}
    }
  \caption{Timeline of significant events and research directions in the field of affordances for robotics tasks.
 }
      \label{fig:timeline}
 \end{figure*}
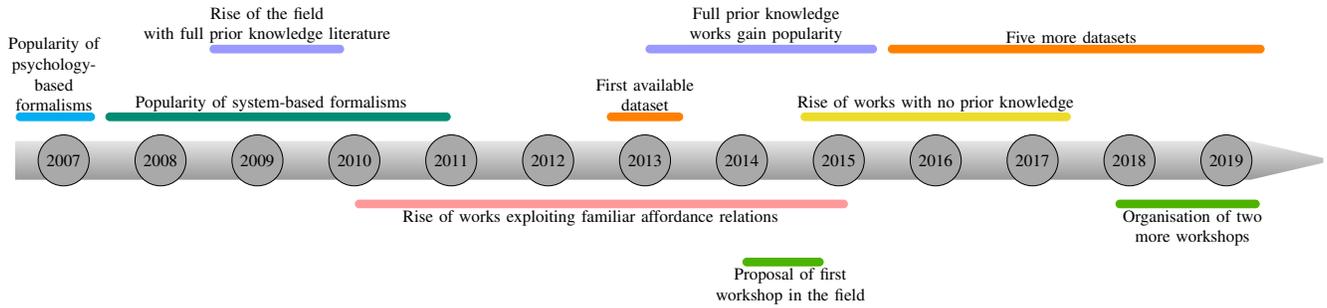

As observed in this section, the perspectives on how affordance should be included in robotics vary among stages. Even though variations of the affordance relation components are used across formalisms in both stages, historically all of the proposals have noticeably focused on how well the methods generalise to new situations. Fig.~\ref{fig:timeline} is a visualisation of the significant events in the field that include affordances in robotics. From this timeline, the evolution of the field towards more human-like generalisation capabilities is evident, as well as the early influence from psychological theories. The areas of improvement are also notable, such as the need for more datasets and activities to gather researchers and discuss progress in the field.


%% file: sections/3_model.tex
\section{Known Affordance Relation \label{sc:model}}

    In robotics, the term affordance has been generalised as a relationship of the three components previously introduced: target, actions and effects. This section is focused of those works that have full \textit{a priori} knowledge on the best possible matches that will make an affordance task successful. The works in this section have access to a list of possible actions and effects given a system's knowledge of the target object. These methods match possible actions onto the known objects following a template. In some cases, the underlying template is updated with new action-object relations acquired through robot self-experience. Thus, in this category, all the approaches have a mapping for the affordance relation.
    
    The nature of the template varies across proposals: (i)~some of them start with a template and keep it as it is (\textit{Section~\ref{ssc:not_update}}), (ii)~others update this template along the way using trial and error (\textit{Section~\ref{ssc:update}}) and, (iii) some others might update this template and include variants such as object-object relation and other agent actions (\textit{Section~\ref{ssc:third_party}}). Nonetheless, in this category, all the works have in common the following:
    \begin{itemize}
        \item They have a grounding of all the components in the affordance relation, i.e., they know the target, the corresponding action and the consequential effect. 
        \item These models require little data to train their model, nonetheless they lack generalisation to novel scenarios and, hence, to variations of the robotics task.
    \end{itemize}
      \begin{figure}[bh!]
        \centering
        \includegraphics[width= 8.7cm]{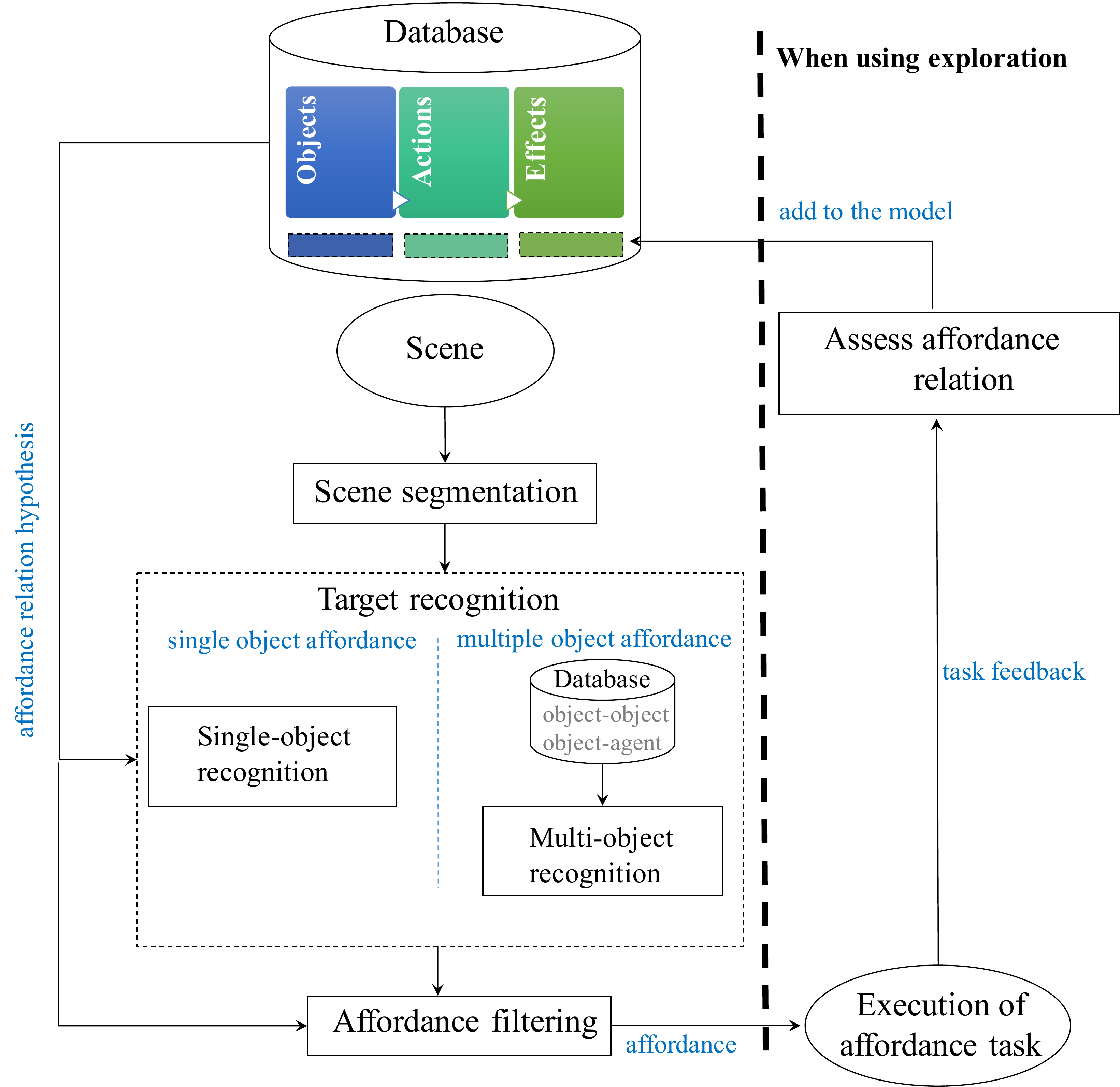}
        \caption{Flowchart for a typical system that has full \textit{a priori} information on the relation of the affordance task components. The relation of the target object, actions and effects serve as input from a database that contains a template \cite{barck2009learning,pandey2013affordance,wang2013robot,Hart2015TheAT}. In some cases the systems update the already known template by matching known objects, actions and possible effects \cite{antunes2016human,wang2013robot,thomaz2009learning}. Particularly, in this category the expansion to new relations such as object-object, object-agent are eased and part of the target recognition.}
        \label{fig:known_data}
      \end{figure}
    A general workflow of the methods in this category is illustrated in Fig.~\ref{fig:known_data}. The diagram illustrates that by knowing a template with the affordance relations the methods can fix this template and query the database that stores this relation, learn it once and add new relations of known instances through exploration or have in storage object-object and object-agent relations to associate an affordance and accomplish a robotics task. Additionally, Table \ref{tb:known_data} summarises the works presented in this section. 
  
    \begin{table*}
        \centering
       \setlength\tabcolsep{5pt}
        \begin{tabular}{p{3.3cm}|l|c c c|c c c c|c c|c c c c|c c c}
        \hline
        \multirow{6}{*}{\multirow{6}{*}{{\textbf{Work}}}}&
        \multicolumn{1}{c|}{\multirow{6}{*}{{\textbf{Robotics Platform}}}}
        &\multicolumn{3}{c|}{\textbf{Robotics Task}}
        &\multicolumn{4}{c|}{\textbf{Perception}}
        &\multicolumn{2}{c|}{\textbf{Actions}}
        &\multicolumn{4}{c|}{\textbf{Means of Data}}
        &\multicolumn{3}{c}{\textbf{Relations}}\\
        \cline{3-18} & &
        \rotatebox[origin=c]{90}{Manipulation}&
        \rotatebox[origin=c]{90}{Navigation}&
        \rotatebox[origin=c]{90}{Action Prediction}&
        \rotatebox[origin=c]{90}{Visual}&
        \rotatebox[origin=c]{90}{Proprioception}&
        \rotatebox[origin=c]{90}{Kinesthetic}&
        \rotatebox[origin=c]{90}{Tactile}&
        \rotatebox[origin=c]{90}{Primitive}&
        \rotatebox[origin=c]{90}{Complex}&
        \rotatebox[origin=c]{90}{Labels}&
        \rotatebox[origin=c]{90}{Demonstrations}&
        \rotatebox[origin=c]{90}{Trial and Error}&
        \rotatebox[origin=c]{90}{Heuristics}&
        \rotatebox[origin=c]{90}{Deterministic}&
        \rotatebox[origin=c]{90}{Probabilistic}&
        \rotatebox[origin=c]{90}{Planning}
        \\
        \hline
            
            \citet{kjellstrom2011visual} & None--theoretical work & \checkmark & & & \checkmark & & & & \checkmark & & &\checkmark & & &\checkmark & &\\ \hline
            
            \mbox{\citet{diana2013deformable}} & Khepera III & &\checkmark &  & \checkmark & & & &  & \checkmark & \checkmark & & & & \checkmark\\ \hline
            
            \mbox{\citet{pandey2013affordance}}  & PR2 & & &\checkmark & \checkmark &  & & & &\checkmark & & \checkmark &  &  &\checkmark &  \\ \hline
            
            \mbox{\citet{fallon2015architecture}}  & Atlas  & \checkmark &\checkmark &  & \checkmark & &  & &  \checkmark &\checkmark& \checkmark& & & &\checkmark\\ \hline
            
            \mbox{\citet{antunes2016human}} & iCub & \checkmark & &  &\checkmark & & & & \checkmark&& \checkmark&&\checkmark&&\checkmark&&\checkmark\\ \hline

            \mbox{\citet{thomaz2009learning}}  & Bioloid & \checkmark & &  &\checkmark & & & & \checkmark&&\checkmark&&&&\checkmark \\ \hline
            
            \citet{wang2013robot}  & NAO & & & \checkmark & \checkmark & & & & \checkmark&&\checkmark&&\checkmark&&\checkmark \\ \hline
            
            \citet{cruz2016training}  & iCub & \checkmark & & &\checkmark & &  & & & \checkmark&&&\checkmark&\checkmark&&\checkmark \\ \hline
            
            \mbox{\citet{do2017affordancenet}} & None--theoretical work  & & &\checkmark & \checkmark & & & & \checkmark & \checkmark&\checkmark&&&\checkmark \\ \hline
            
            \mbox{\citet{myers2015affordance}} & None--theoretical work  & & &\checkmark & \checkmark & & & & \checkmark & \checkmark&\checkmark&&&&\checkmark \\ \hline
            
            \mbox{\citet{barck2009learning}} & Kuka \& Barret hand & \checkmark & & &  \checkmark& & & & \checkmark&&\checkmark&&&&\checkmark&\checkmark\\ \hline
            
            \citet{sun2010learning} & \mbox{PeopleBot and Rovio} & & \checkmark & & \checkmark & &  & &  & \checkmark&&\checkmark&&&&\checkmark \\ \hline
            
            \citet{sun2014object} & Fanuc \& Barret hand & \checkmark & & & \checkmark & &  & & \checkmark & \checkmark&\checkmark&&&&&\checkmark \\ \hline
            
            \mbox{\citet{moldovan2014occluded}}  & None--theoretical work & & & \checkmark & \checkmark & & & & & \checkmark &\checkmark&&&&&\checkmark\\ \hline
            
            \citet{Hart2015TheAT} & Valkyrie \& \mbox{Robonaut2} & \checkmark & & & \checkmark & & & & \checkmark & \checkmark&\checkmark&&&&\checkmark\\ \hline
            
            \citet{zhu2014reasoning} & None--theoretical work & & & \checkmark & \checkmark & & & & \checkmark &\checkmark&\checkmark&&&&&\checkmark \\ \hline
            
            \citet{saxena2014robobrain} & PR2 & \checkmark & & & \checkmark &\checkmark&  &  & & \checkmark&\checkmark&&\checkmark&&&\checkmark\\ \hline
            
            \citet{song2015task} & Armar III & \checkmark & & & \checkmark & & & \checkmark &  &\checkmark &  & \checkmark & & & &\checkmark & \\ \hline
            
            \citet{bekiroglu2013predicting} & ATI Mini45 & \checkmark & & & \checkmark &\checkmark & & & \checkmark & & &  &\checkmark & \checkmark &&\checkmark & \\ \hline
            
            \citet{song2013predicting} & Tombatossals & \checkmark & & & \checkmark & & & \checkmark &  & \checkmark& &\checkmark & &  & &\checkmark & \\ \hline
            
             \citet{gijsberts2010object} & None--theoretical work & \checkmark & & & \checkmark & & & \checkmark &  & \checkmark& &\checkmark & &  &\checkmark & & \\ \hline
             
             \citet{nishide2009modeling} & HRP-2 & \checkmark & & & \checkmark &\checkmark & &  & \checkmark & & & &\checkmark &  &&\checkmark  & \\ \hline
             
             \citet{song2015learning} & PR2 & \checkmark & & & \checkmark & & &  &  &\checkmark & \checkmark & & &  &\checkmark &  & \\ \hline
             
             \citet{szedmak2014knowledge} & None--theoretical work &  & &\checkmark  &\checkmark  & & &  &  &\checkmark & \checkmark & & &  &\checkmark &  & \\ \hline
             
             \citet{dogar2007primitive} & Kurt 2 & & \checkmark &  &\checkmark  & & &  &  &\checkmark &  & &\checkmark &  &\checkmark &  &\checkmark \\ \hline
             
             \citet{lewis2005foot} & Own bipedal design & &\checkmark  &  &\checkmark  & & &  & \checkmark & & \checkmark & & &  &\checkmark &  & \\ \hline
             
             \citet{song2011embodiment} & Armar III &\checkmark &  &  &\checkmark  & & &\checkmark  & & \checkmark& \checkmark & & & & &\checkmark  & \\ \hline
             
              \citet{kostavelis2012collision} & Kurt 2 & &\checkmark & &\checkmark &\checkmark & & & & \checkmark & \checkmark & & & &\checkmark &  & \\ \hline
              
              \citet{omrvcen2009autonomous} & Armar III &\checkmark & & &\checkmark & & & & \checkmark&  & & & & \checkmark&\checkmark &  & \\ \hline
              
              \citet{price2016affordance} & Kuka KR5 &\checkmark & & &\checkmark &\checkmark & & & & \checkmark &\checkmark & & & & &\checkmark &\checkmark \\ \hline
              
               \citet{sweeney2007model} & Gifu hand &\checkmark & & &\checkmark & & & &\checkmark & & \checkmark& & & & &\checkmark & \\ \hline
                
                \citet{kroemer2011flexible} & Dexter &\checkmark & & &\checkmark & \checkmark& & &\checkmark & & \checkmark& & & & \checkmark & &\checkmark \\ \hline
                
                \citet{cutsuridis2013cognitive} & None--theoretical work &\checkmark & & &\checkmark & \checkmark& & &\checkmark & & \checkmark& & & &  &\checkmark &\checkmark \\ \hline

        \end{tabular}
        
        \caption{Summary of works that have full \textit{a priori} knowledge of the relation among affordance components to perform a task. \label{tb:known_data}}
    \end{table*}

      \begin{figure}[bh!]
         \centering
          \subfigure[\citet{antunes2016human} ]{\label{fig:antunes2016human} \includegraphics[width=3.8cm]{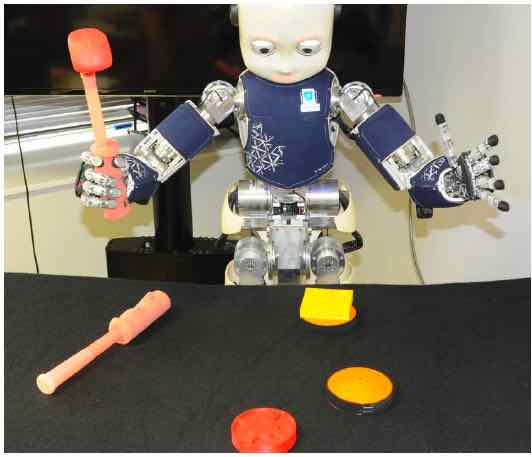}}
          \subfigure[\citet{Hart2015TheAT} ]{\label{fig:Hart2015TheAT} \includegraphics[width=4.4cm]{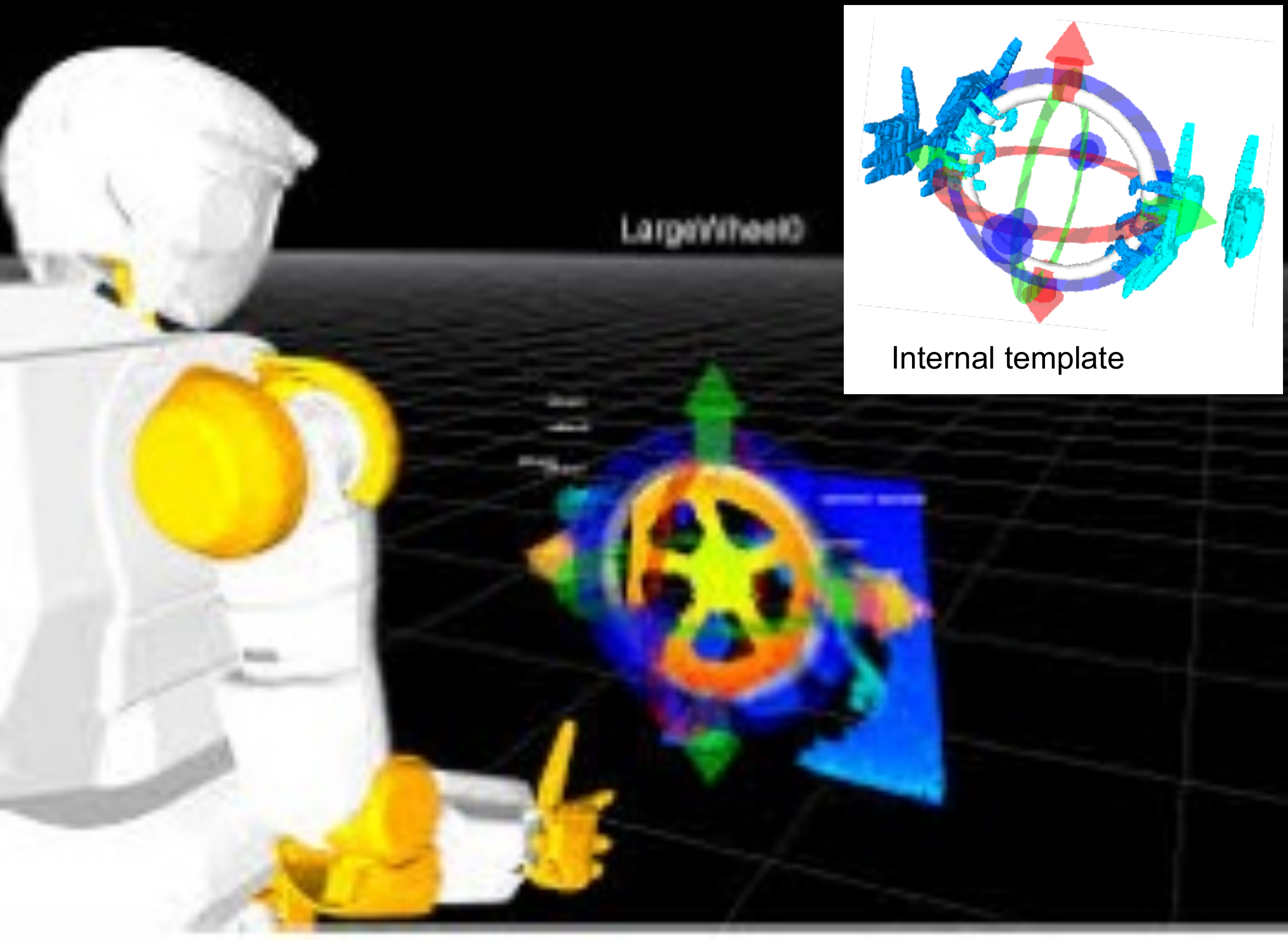}}
         
         \caption{Examples of works that build affordance relations with complete prior knowledge by using templates. a) \citet{antunes2016human} use the iCub robot to follow indications to prepare a sandwich. The robot needs to get to the unreachable ingredients using the provided tools (rake and stick). b) \citet{Hart2015TheAT} propose an affordance template. For example, in a wheel the template uses different waypoints to pre-grasp, grasp, turn-goal, etc.}
         \label{fig:model_based}
      \end{figure}

    \subsection{Learning the affordance relations model without update \label{ssc:not_update}}

        Works in this sub-category learn a single template of matching affordance relations and have it fixed for the rest of the robotics task. They learn the best match of target, action and effects from labelled data using probabilistic approaches such as Gaussian processes and Bayesian Networks \cite{barck2009learning,sun2010learning,sun2014object,antunes2016human,sweeney2007model,moldovan2014occluded} or deterministic rule-based models \cite{pandey2013affordance,fallon2015architecture,wang2013robot,barck2009learning,diana2013deformable}. Thus, the actions become probabilistic rules that are queried at execution time. In general, rule-based approaches only require a small amount of data. Nonetheless, these approaches are deterministic and cannot account for uncertainty, while probabilistic methods can. \citet{barck2009learning} combine both deterministic and probabilistic models. The authors in \cite{barck2009learning} compare a probabilistic method based on Bayesian Networks with a rule-based model of axioms to reason about grasp selection for an object. The Bayesian Network performs well under uncertainty, by using axioms to build an ontological approach. This allows their system to learn an affordance relation much faster with fewer samples.
        
        Approaches that attempt to perform multi-step predictions based on known affordance relations, such as \cite{kroemer2011flexible,price2016affordance,cutsuridis2013cognitive}, often add a planning layer that allows them to achieve goal-oriented tasks by learning either primitive or complex actions.
       
        Nonetheless, when uncertainty is introduced in the scenario the success of the task is compromised and probabilistic models are preferred to rule-based ones. A clear example of uncertainty is when the system needs to perform in cluttered environments. \cite{sun2010learning,lewis2005foot,kostavelis2012collision} and \cite{moldovan2014occluded,antunes2016human} perform navigation and grasping application tasks, respectively. They do so in a scenario with many objects where the purpose is to identify the most suitable affordance relation. In \cite{sun2010learning,kostavelis2012collision,dogar2007primitive}, the goal is to arrive at a destination while pushing or nudging objects, while in \cite{moldovan2014occluded}, the goal is to find an object that might be occluded in a shelf to achieve a queried action.
       
        To predict actions for collaborative tasks, \citet{pandey2013affordance} introduce the concept of affordance graphs for decision making in human-robot interaction tasks, while \citet{wang2013robot} use a table of interpretable triplets, containing the affordance components. In \cite{pandey2013affordance}, the graph encodes the actions an agent might be able to take with an object. These actions are ranked according to an effort level from which the agent optimally chooses the lowest effort demanding action that achieves a successful effect. \citet{wang2013robot}'s system is presented with a knowledge table relating the affordance components and is updated using a short term memory with a simple forgetting mechanism.
        Other approaches use
        affordance templates based on \acp{KB} or \acp{CNN} to build databases of affordance relations \cite{fallon2015architecture,Hart2015TheAT,saxena2014robobrain,zhu2014reasoning,do2017affordancenet,myers2015affordance}. In \cite{fallon2015architecture,Hart2015TheAT}, the system is commanded by an operator to perform a manipulation task and the templates provide information about the optimal grasping regions of the objects. An example is shown in Fig.~\ref{fig:Hart2015TheAT}. 
        
        \citet{song2015task} developed a framework by stages composed of \cite{song2011embodiment,song2013predicting,song2015task} that makes sure a robotic end-effector properly hands over an object. In this framework the objects are known as well as their corresponding actions and effects. The focus is to grasp the objects in a way that ensures the other agent has enough surface left on the object to be able to grasp it.
        
    \subsection{Updating the learned affordance relations \label{ssc:update}}
        
        Methods in this sub-category learn a model and update it using trial and error to find new combinations of known target, action and effect relations. These works ground the actions through exploration by predicting the likelihood of a sub-set of effects on known objects.
        Clear examples of building such models are \cite{bekiroglu2013predicting,omrvcen2009autonomous,nishide2009modeling} that work with the same object throughout the task knowing the action that they need to perform upon it but observing the effects to learn the most efficient way to perform the indicated task.
    
        \citet{wang2013robot,antunes2016human,bekiroglu2013predicting,nishide2009modeling,dogar2007primitive,saxena2014robobrain} combine pre-learned models with exploration to assess the effects of an action. In \citet{antunes2016human}, the action grounding list that has been learned using Bayesian Networks is modified using the robot's knowledge about the object and its surrounding. This knowledge is acquired through exploration, as in the example shown in Fig.~\ref{fig:antunes2016human}\, where \textit{grasp\_object\_with\_hand} becomes \textit{grasp\_spoon\_with\_left\_hand}. Similarly, \citet{wang2013robot} start with a table of affordance triplets that is expanded using self-experience, e.g., if another action achieves the same or a better effect on a target object then that triplet is added. 
        
        \citet{dogar2007primitive} combine the learning mechanism for navigation purposes where the system has learned a set of primitive actions that then tests on previously seen environments to build complex goal-oriented motions to learn the transversability of the objects.
        
        Along the same lines, \citet{kjellstrom2011visual} improve the classification rate by learning from human demonstration to grasp objects in a household task. Their method categorises manipulated objects and human manipulation action in the context of each other, thus creating an affordance relation. Using demonstrations, \citet{thomaz2009learning} hand label different sample objects with the most obvious affordance category and combine self-exploration with learning from demonstration to learn the effects of the hand labelled actions.
        
         \citet{cruz2016training} complete a cleaning task where the simulated robot uses reinforcement learning and a predefined set of contextual affordances with few starting actions. Having this prior information enables the system to reach higher rates of success, which is the case for \cite{cruz2016training,wang2013robot}.
        \citet{gijsberts2010object} improve the categorisation task of objects by adding a map with grasp information related with the affordances of the objects.

    \subsection{Considering a third party \label{ssc:third_party}}
    
        One possible strategy to achieve better performance on collaboration and human-robot interaction tasks is to identify external elements to the basic affordance relation. Works in this category consider the affordance task as a relation of not only the fundamental target, action, and effect but also as a relation among different targets and other agents in the scene. 
        Namely, in this sub-category, we discuss those works that consider relations such as (i)~object-object and (ii)~agent-object-agent.
        
\begin{figure}[t!]
         \centering
          \subfigure[\citet{moldovan2014occluded} ]{\label{fig:moldovan2014occluded} \includegraphics[width=4.5cm]{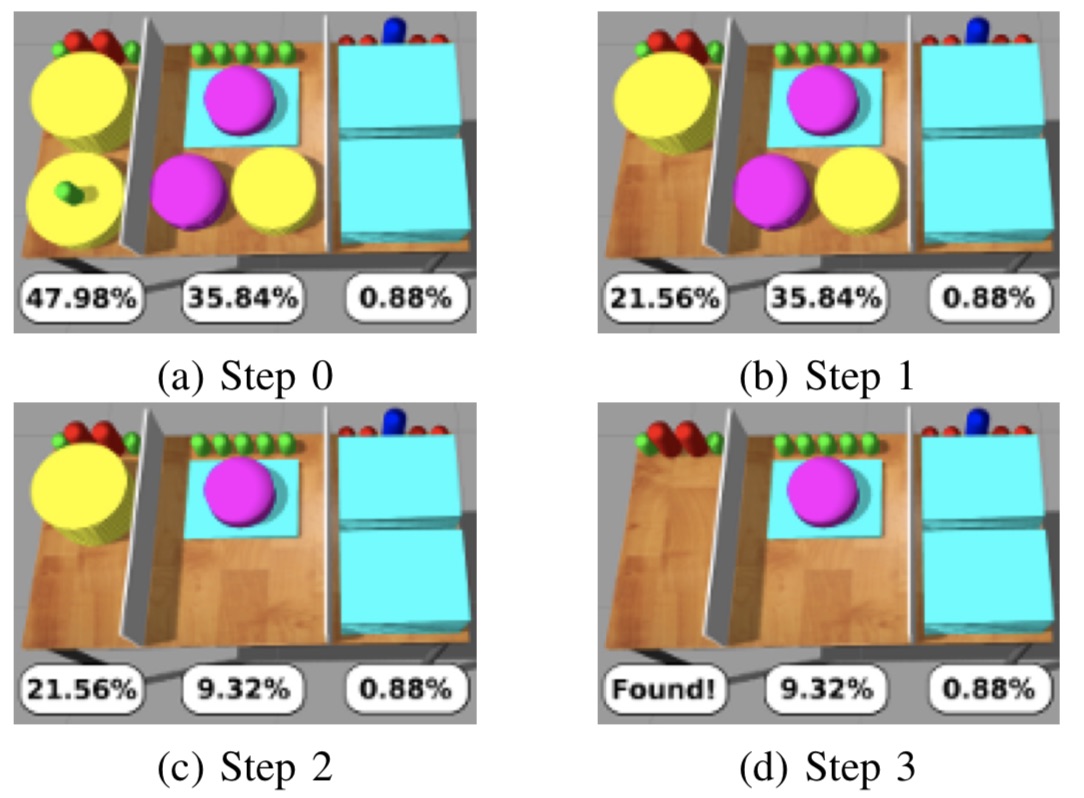}}
          \subfigure[\citet{sun2014object} ]{\label{fig:sun2014object} \includegraphics[width=4cm, trim = -2.5cm -2.7cm 0cm 0cm]{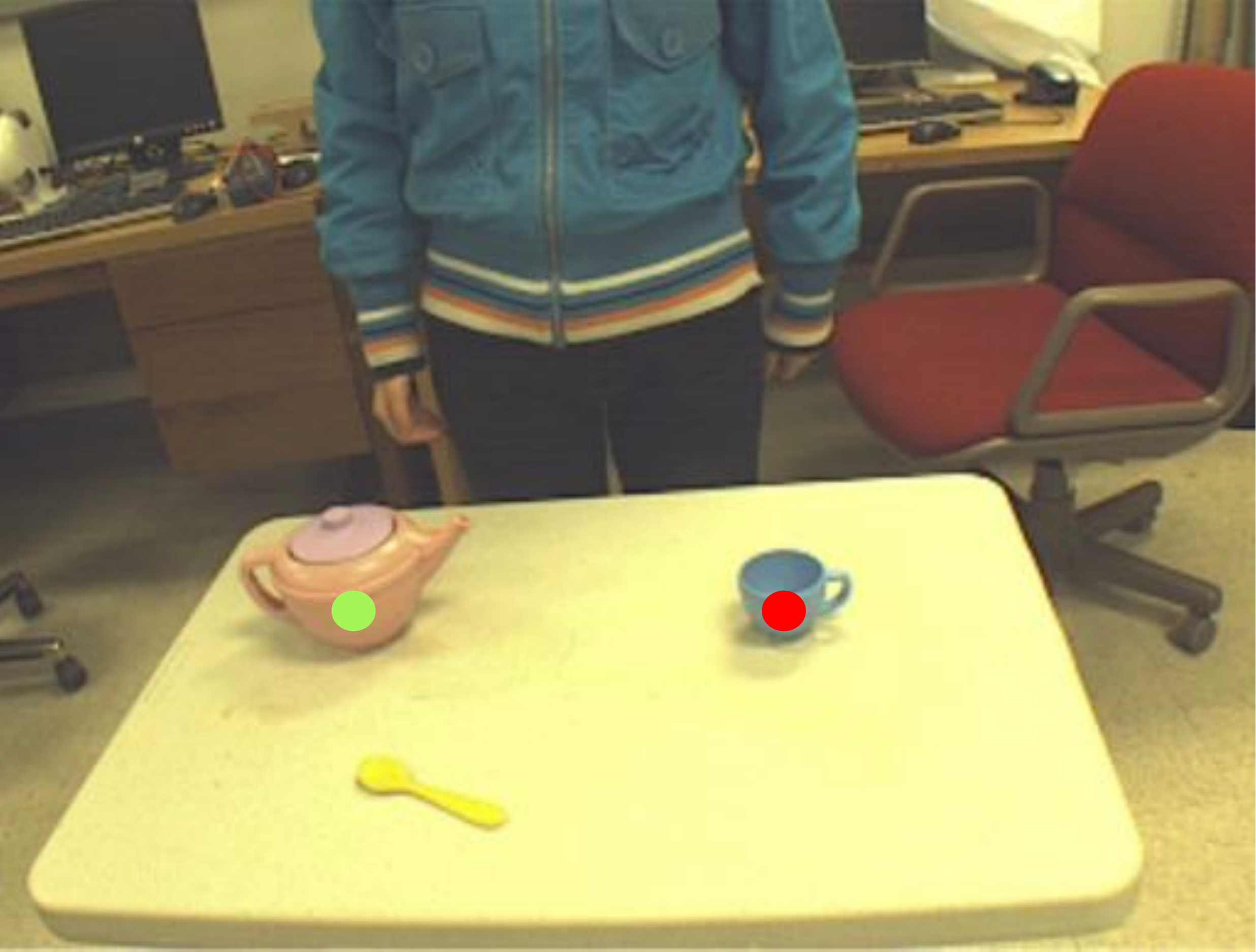}}
        
         \caption{Examples of methods with complete prior knowledge that use templates for the multi-object affordance relation. Both examples show the pouring affordance relation. a) \citet{moldovan2014occluded} relate the objects among shelves, and b) \citet{sun2014object} relate two objects (objects with green and red dots are related) to find the most likely affordance relation (to pour).}
         \label{fig:model_based_third_party}
      \end{figure}

        In the object-object affordance relation, \cite{moldovan2014occluded} uses labelled images of shelves in a kitchen to learn a realistic distribution of objects in that environment, thus being able to find an object in a shelf. To achieve this goal, they model the concept of object co-occurrence by calculating the probability of an object on a shelf being of a particular type and having a specific affordance, given that on the same shelf there are objects of a certain type.
        \citet{sun2014object} and \citet{szedmak2014knowledge} also present an object-object affordance learning approach for the system to learn the interactive functionality of objects. In \cite{sun2014object}, the graphical model is designed to intuitively represent the functional connectivity of the objects, such as a teapot and a cup, or a book and schoolbag, and extends that connectivity to manipulation motions. Examples of such methods are shown in Fig.~\ref{fig:model_based_third_party}.
       
        In the agent-object-agent and agent-agent relation, \citet{pandey2013affordance} and \citet{price2016affordance} consider the actions of more agents that might interact in the same environment. Both works design a system where they consider the action capabilities of manipulating the objects among different agents and across places. 
      
        In this section we explored those works that rely on templates to identify an affordance relation, thus having full \textit{a priori} knowledge of the relating affordance components. Although they are able to efficiently identify known affordances, they are not able to perform on previously unseen variations of the robotics task. 
        
%% file: sections/4_combined_methods.tex
\section{Familiar Affordance Relations\label{sc:combined}}

 Including affordances in robotic tasks can also be done by correlating features from previously known scenarios to new situations that share similar properties. In particular, affordances are most likely to be associated with features such as geometrical shape and texture of an object than they are with its object class \cite{bohg2010learning}. Often, the classification of an object is determined by its function. In contrast to Section~\ref{sc:model}, this category aims at deducing the affordance relation by learning the target object features that represent an affordance, rather than associating the target object class into an affordance relation. Therefore, these works have only partial \textit{a priori} information to perform the robotics task. Works in this category exhibit the following common factors:
    \begin{itemize}
        \item They are able to generalise to familiar robotics tasks that include affordances given that their grounding is not attached to target categories.
        \item Even though they do not learn target categories, these approaches have a prior mapping of features that hint at an action and the corresponding effects, thus not being completely exploratory.
    \end{itemize}
         
    Throughout this section we will see that the works are not only able to generalise to familiar tasks but also that their generalisation capabilities extend to: (i)~help improve other tasks such as object detection (\textit{Section~\ref{ssc:detection}}), (ii)~generalise over detecting affordances for multiple objects in cluttered environments (\textit{Section~\ref{ssc:cluttered}}), and (iii)~translating the affordance relation from human demonstrations to similar objects (\textit{Section~\ref{ssc:developmental}}).
    
\begin{figure}[t!]
      \centering
      \includegraphics[width= 8.7cm]{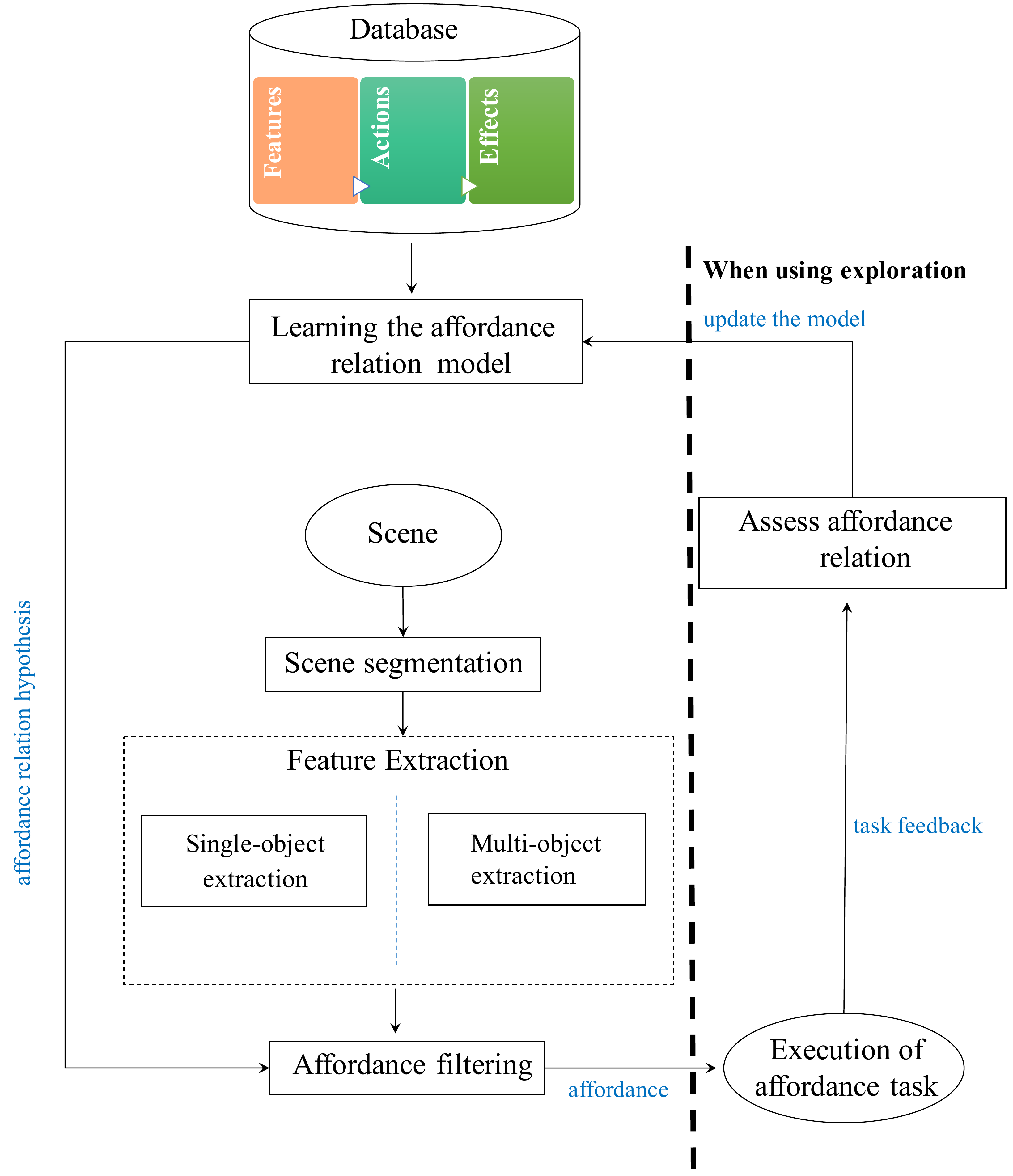}
      \caption{Flowchart for a typical system that has partial \textit{a priori} information on the relation of the affordance task components. In this category, the relation is built from object target features rather than object categories \cite{mar2015multi,song2010learning,bohg2017interactive}. In some cases, the features correspond to multiple objects, for instance in cluttered scenes \cite{aldoma2012supervised,hermans2013decoupling,stark2008functional}. Alternatively, the systems use self-exploration to identify new relations of known features, actions and possible effects to then update the model \cite{cos2004using,detry2011learning,gonccalves2014learning}.}
      \label{fig:familiar_data}
    \end{figure}

    A summary of the typical workflow for methodologies in this category is shown in Fig.~\ref{fig:familiar_data}. The diagram illustrates that by knowing a relation between features, actions and effects, the works in this category are able to generalise their learned affordance relation model to detect the affordance on one or multiple objects in the scene. Moreover, similarly to Section~\ref{sc:model}, some of them update this learned model by matching actions-effects to known features in an attempt to achieve a more generic system. Table~\ref{tb:combined_online_actions} shows a summary of the works in this category.

    \begin{table*}
        \centering
       \setlength\tabcolsep{5pt}
        \begin{tabular}{p{3.3cm}|l|c c c|c c c c|c c|c c c c|c c c}
        \hline
        \multirow{6}{*}{\multirow{6}{*}{{\textbf{Work}}}}&
        \multicolumn{1}{c|}{\multirow{6}{*}{{\textbf{Robotics Platform}}}}
        &\multicolumn{3}{c|}{\textbf{Robotics Task}}
        &\multicolumn{4}{c|}{\textbf{Perception}}
        &\multicolumn{2}{c|}{\textbf{Actions}}
        &\multicolumn{4}{c|}{\textbf{Means of Data}}
        &\multicolumn{3}{c}{\textbf{Relations}}\\
        \cline{3-18} & &
        \rotatebox[origin=c]{90}{Manipulation}&
        \rotatebox[origin=c]{90}{Navigation}&
        \rotatebox[origin=c]{90}{Action Prediction}&
        \rotatebox[origin=c]{90}{Visual}&
        \rotatebox[origin=c]{90}{Proprioception}&
        \rotatebox[origin=c]{90}{Kinesthetic}&
        \rotatebox[origin=c]{90}{Tactile}&
        \rotatebox[origin=c]{90}{Primitive}&
        \rotatebox[origin=c]{90}{Complex}&
        \rotatebox[origin=c]{90}{Labels}&
        \rotatebox[origin=c]{90}{Demonstrations}&
        \rotatebox[origin=c]{90}{Trial and Error}&
        \rotatebox[origin=c]{90}{Heuristics}&
        \rotatebox[origin=c]{90}{Deterministic}&
        \rotatebox[origin=c]{90}{Probabilistic}&
        \rotatebox[origin=c]{90}{Planning}
        \\
        \hline
    
            \citet{cos2004using} & Khepera &  &\checkmark&& \checkmark & & &  & &\checkmark&\checkmark&&&&\checkmark\\ \hline
            
            \citet{stark2008functional} & None--theoretical work & \checkmark&& & \checkmark & & && \checkmark & \checkmark&&\checkmark&&&\checkmark\\ \hline
            
            \citet{castellini2011using} & None--theoretical work & \checkmark&& & \checkmark & & & \checkmark & \checkmark&&&\checkmark&&&\checkmark\\ \hline
            
            \citet{aldoma2012supervised} & None--theoretical work& & & \checkmark & \checkmark & & & & & \checkmark&\checkmark&&&&&\checkmark\\ \hline
            
            \citet{kim2014semantic}  & PR2 & & &\checkmark & \checkmark & & & & \checkmark&&\checkmark&&&&&\checkmark\\ \hline
            
            \citet{katz2014perceiving} & Barrett & \checkmark& & & \checkmark & &\checkmark & & \checkmark&&\checkmark&&&&\checkmark\\ \hline
            
            \citet{gonccalves2014learning} & iCub & \checkmark&& & \checkmark && & & \checkmark&&&&\checkmark&&&\checkmark\\ \hline
            
            \citet{fritz2006learning} & Kurt2 & &\checkmark && \checkmark & & & &  & \checkmark&\checkmark&&&&&\checkmark \\ \hline
            
            \citet{song2010learning}  & Barret hand & \checkmark&& & \checkmark &  & \checkmark & \checkmark&\checkmark&&&\checkmark&&&&\checkmark \\ \hline
            
            \mbox{\citet{ugur2009affordance}}  & Gifu Hand III  &  \checkmark& & & \checkmark && & & \checkmark&&&&\checkmark&&\checkmark&&\checkmark \\ \hline
            
            \mbox{\citet{hermans2011affordance}}  & Pioneer 3 DX & &\checkmark&  & \checkmark  & & & & \checkmark&&&&\checkmark&&\checkmark\\ \hline
            
            \mbox{\citet{bohg2010learning}} & Armar head \& Kuka arm & \checkmark& & & \checkmark & & & &\checkmark&&\checkmark&&&&\checkmark\\ \hline
            
            \citet{mar2015multi} & None--theoretical work & \checkmark& & & \checkmark &  & & & \checkmark&&&\checkmark&&&&\checkmark \\ \hline
            
            \citet{nguyen2017object} & WALK-MAN &&&\checkmark & \checkmark &&  & & & \checkmark&\checkmark&&&&&\checkmark \\ \hline
            
            \mbox{\citet{moldovan2012learning}} & iCub & \checkmark& & & \checkmark & & && \checkmark&&\checkmark&&&&&\checkmark \\ \hline
            
            \citet{tikhanoff2013exploring} & iCub & \checkmark& & & \checkmark & & &  & \checkmark&&\checkmark&&&\checkmark&&\checkmark \\ \hline
            
            \mbox{\citet{de2006learning}} & P5 glove & \checkmark&& &  & && \checkmark & \checkmark&&&\checkmark&&&&\checkmark\\ \hline
            
            \citet{kraft2009learning} & Staubli & \checkmark& & & \checkmark & & & \checkmark & \checkmark&&&&\checkmark&&&\checkmark\\ \hline
            
            \citet{detry2011learning} & Barret hand & \checkmark& & & \checkmark & & & \checkmark & \checkmark&&&&\checkmark&&&\checkmark\\ \hline
            \citet{ruiz2018can} & PR2 & & &\checkmark & \checkmark & & &&& \checkmark & \checkmark&&&&\checkmark\\ \hline
            
            \citet{ye2017can} & Baxter & & &\checkmark & \checkmark & & &&& \checkmark & \checkmark&&&&\checkmark\\ \hline
            
            \citet{chu2019learning} & None--theoretical work & & &\checkmark & \checkmark & & &&& \checkmark & \checkmark&&&&\checkmark\\ \hline
            
            \citet{kaiser2014extracting} & Armar III &\checkmark & & & \checkmark &\checkmark & &&& \checkmark & &&&\checkmark&\checkmark\\ \hline
            
            \citet{baleia2015exploiting} & Their own design & &\checkmark & & \checkmark & & & \checkmark &\checkmark &  & &&\checkmark&\checkmark&\checkmark\\ \hline
            
             \mbox{\citet{ugur2007learning}} & Kurt3D & \checkmark&& & \checkmark & &  & &\checkmark &&&&\checkmark&\checkmark&\checkmark  \\ \hline
             
             \citet{ugur2007curiosity} & Kurt3D  & \checkmark&& & \checkmark &  & & & \checkmark&&&&\checkmark&\checkmark&\checkmark\\ \hline
            
            \citet{aksoy2011learning}  & None--theoretical work & & &\checkmark &\checkmark & & &  & &\checkmark &\checkmark &&&\checkmark&\checkmark\\ \hline
            
            \citet{aksoy2015model}  & None--theoretical work & & &\checkmark &\checkmark & & &  & &\checkmark &\checkmark &&&\checkmark&\checkmark&&\checkmark\\ \hline
            
            \citet{erkan2010learning}  & Barret hand &\checkmark & & & \checkmark & & & & \checkmark & & &&\checkmark&\checkmark&\checkmark\\ \hline
            
             \citet{chu2016learning}  & Curi & \checkmark & & & \checkmark & & & & \checkmark & & &\checkmark&\checkmark&\checkmark&&\checkmark\\ \hline
             
             \citet{hermans2013learning}  & PR2 & \checkmark & & & \checkmark & & & & \checkmark & & &&\checkmark&&&\checkmark\\ \hline
             
             \citet{dag2010learning}  & None--theoretical work & & &\checkmark & \checkmark & & & & \checkmark & & \checkmark&&&&\checkmark&\\ \hline
             
             \citet{kim2015interactive}  & PR2 &\checkmark & & & \checkmark & & & & \checkmark & & &&\checkmark&&&\checkmark\\ \hline
             
             \citet{chan2014determining}  & Their own design &\checkmark & & & \checkmark & & & & & \checkmark & &\checkmark&&&&\checkmark\\ \hline
             
              \citet{dehban2016denoising}  & Icub &\checkmark & & & \checkmark & & & &\checkmark &  & &&\checkmark&&&\checkmark\\ \hline
              
              \mbox{\citet{varadarajan2012afrob}}  & None--theoretical work & & &\checkmark & \checkmark & & & & &\checkmark  &\checkmark &&&&\checkmark&\\ \hline
              
              \citet{ridge2013action}  & None--theoretical work & \checkmark & & & \checkmark & & & &\checkmark &  & &\checkmark&&&\checkmark&\\ \hline
              
              \citet{abelha2016model}  & None--theoretical work & & &\checkmark & \checkmark & & & & \checkmark&&\checkmark &&&&\checkmark&\\ \hline
              
              \citet{griffith2011behavior}  & WAM by Barret & \checkmark& & & \checkmark & &&&\checkmark & && &\checkmark&&\checkmark&\\ \hline
              
              \citet{lopes2007affordance}  & Baltazar & \checkmark& & \checkmark& \checkmark & &&&\checkmark & && &\checkmark&&\checkmark&\\ \hline
              
               \citet{montesano2009learning}& Baltazar & \checkmark&& & \checkmark & & &  & \checkmark&&&\checkmark&&&\checkmark\\ \hline
               
                 \citet{koppula2013learning} & PR2 & & &\checkmark & \checkmark &  && &\checkmark & \checkmark&&\checkmark&&&&\checkmark&\\ \hline
                 
                  \citet{ardon2019learning} & PR2 &\checkmark & & & \checkmark &  && &\checkmark & \checkmark&\checkmark&&&&&\checkmark&\\ \hline

        \end{tabular}
        
        \caption{Summary of methods that have partial \textit{a priori} information on the affordance components relation to achieve a task. \label{tb:combined_online_actions}}
  \end{table*}
 

\subsection{Improving object detection and categorisation \label{ssc:detection}}

\begin{figure*}
         \centering
          \subfigure[\citet{aldoma2012supervised} ]{\label{fig:aldoma2012supervised} \includegraphics[width=5.5cm]{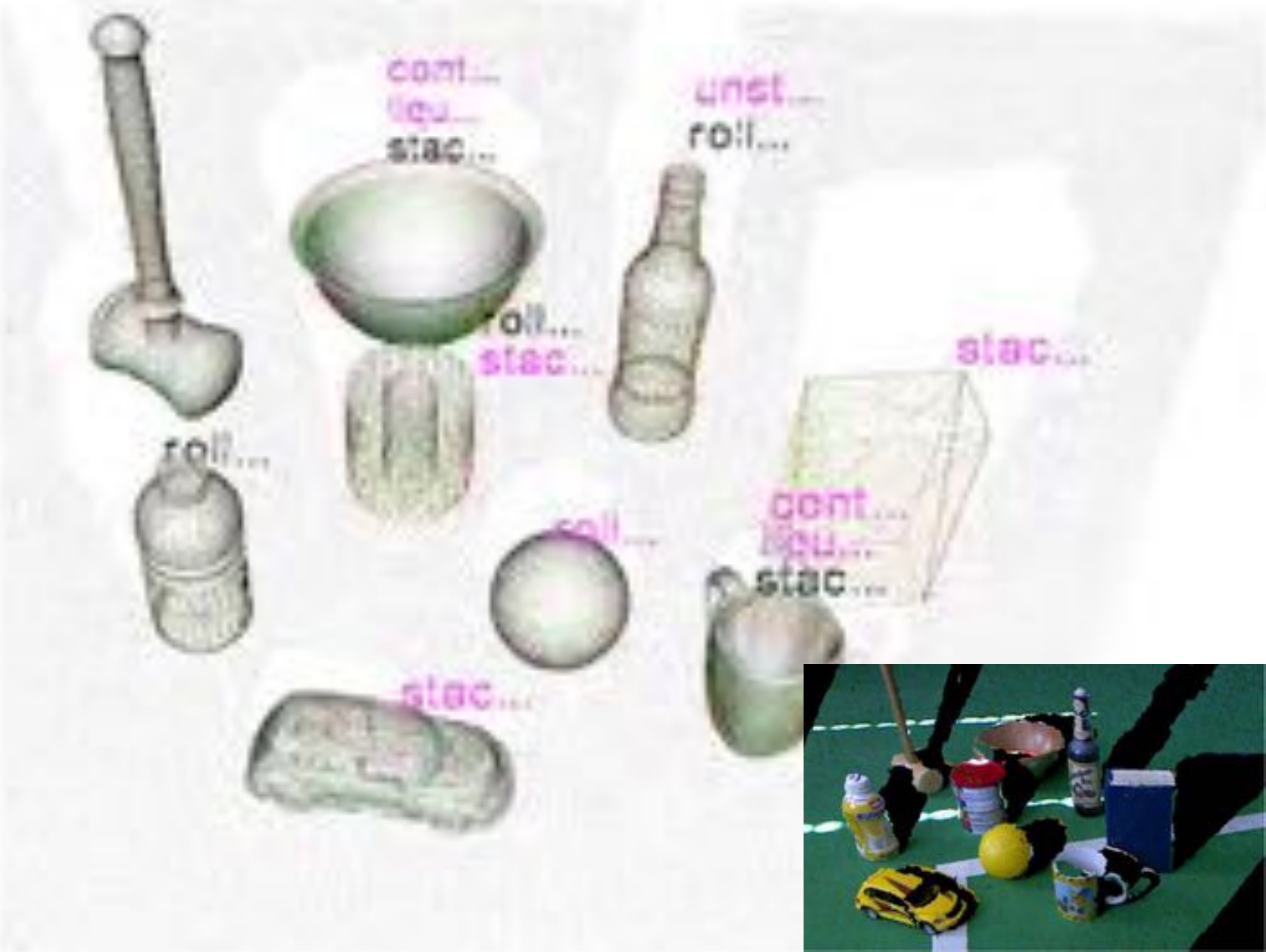}}
          \subfigure[\citet{stark2008functional} ]{\label{fig:stark2008functional} \includegraphics[width=5.4cm]{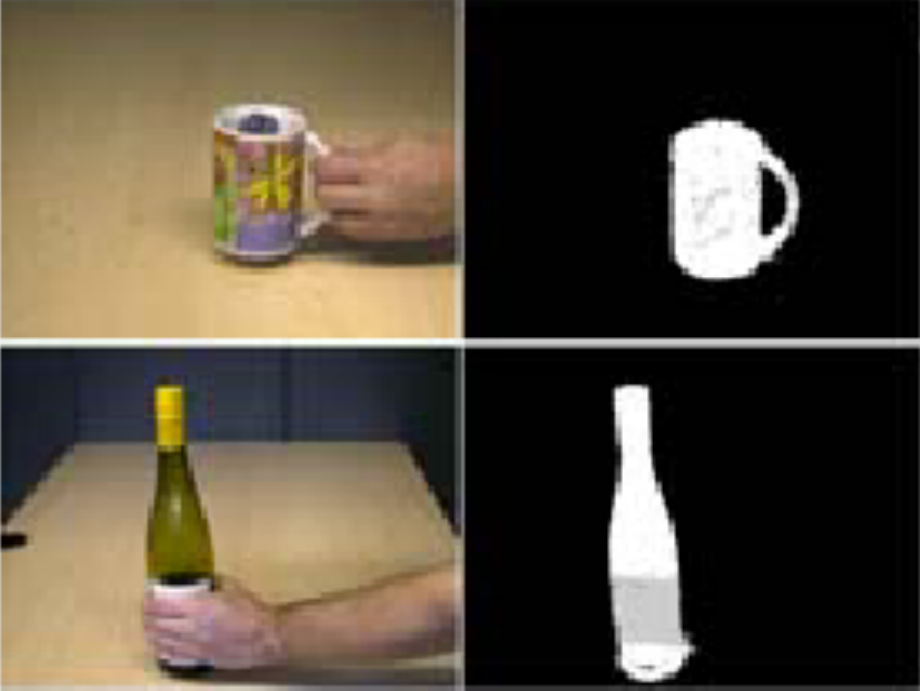}}
          \subfigure[\citet{detry2011learning} ]{\label{fig:detry2011learning} \includegraphics[width=6cm,trim = 0cm -2cm 0cm 0cm]{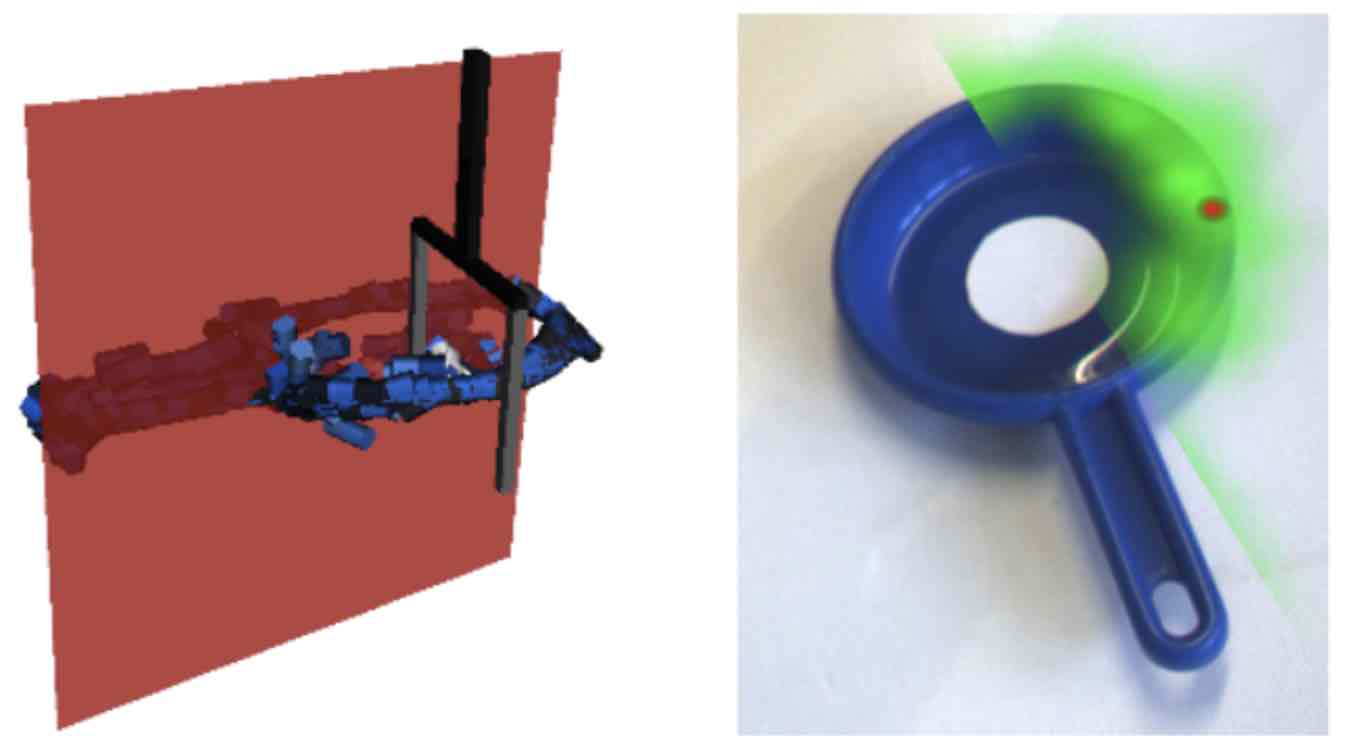}}
        
         \caption{Examples of methods that have partial knowledge on the affordance relation. a) \citet{aldoma2012supervised} work on hidden and non-hidden affordances using features in cluttered environments. Hidden affordances are shown in black and non-hidden in magenta. b) \citet{stark2008functional} show the interacting regions of the objects in grey. c) \citet{detry2011learning} use features to hypothesise grasp affordance region of the objects, where the green area shows the mask where the end-effector can reach (as calculated by the simulation with the red background plane) and where the red dot on the object is the most reachable spot.}
         \label{fig:combined}
      \end{figure*}

Among many other robotics task applications, including the concept of affordance has been particularly useful in solving object recognition and categorisation problems in vision problems. 
\citet{stark2008functional,hermans2011affordance,kim2014semantic,castellini2011using,dag2010learning,griffith2011behavior,aksoy2011learning,aksoy2015model} are examples of methods that argue that object features provide a more appropriate mid-level representation for object prediction than mere object classes, while \cite{aksoy2011learning,ye2017can} share the same argument but for indoor scene detection. Moreover, \cite{stark2008functional,hermans2013decoupling,kim2014semantic,dag2010learning} organise these features by their `functionality', i.e., the affordance these features are related to, such as handle affords grasping, as does the surface of a bottle as shown in Fig.~\ref{fig:stark2008functional}.
In these cases, learning the affordance relation of the features with actions results in a superior generalisation performance for object categorisation. 
\citet{stark2008functional}, distinguish objects according to functional aspects to enable manipulation. \citet{kim2014semantic} use objects' parts \ac{3-D} point clouds as geometric features. These features are then examined, classified and linked to previously seen affordance relations. In both of these methods, including affordances in their robotics task showed to improve the detection and recognition task with respect to those test scenarios where the objects were organised in classes.

\citet{mar2015multi,abelha2016model} enable more precise affordance predictions for tool use scenarios. In \citet{mar2015multi}, instead of learning a single model that tries to relate all the possible variables in an affordance, the robot should learn a separate affordance model for each set of tools and corresponding grasping poses sharing common functionality, thus categorising tool handles and poses. Along the same lines, \cite{castellini2011using} propose the use of grasping motor data (i.e., kinematic grasping data obtained from human demonstrations) to encode the affordances of an object, and then to use this representation on similar objects to improve object recognition.

\subsection{Affordances for multiple objects in cluttered environments \label{ssc:cluttered}}

    Generally, in robotics, a cluttered environment refers to a scene in which the order changes over time, thus being non-static.
    It is natural that once the methodologies are able to generalise better, they explore their performance on more complex surroundings. This is the case for \cite{aldoma2012supervised,katz2014perceiving,hermans2011affordance,nguyen2017object,stark2008functional,ardon2019learning} that perform their affordance detection on table-top scenarios containing more than one object. These methodologies are able to detect all the objects' affordances simultaneously by extracting the features and generalising over previously seen similar scenarios. An example is illustrated in Fig.~\ref{fig:aldoma2012supervised}. 
    
    In works such as \cite{kim2014semantic,ruiz2018can,chu2019learning,kaiser2016towards,varadarajan2012afrob}, we can see an interesting combination of generalisation among cluttered environments and object-object relations. Instead of detecting objects on table-top scenarios, \cite{kim2014semantic,kaiser2014extracting} focus on analysing a scene and finding the affordances of objects in relation to other objects. For example, a box detected behind a table cannot afford being pushed forward. 
  
\subsection{Using experience and demonstrations to model relations \label{ssc:developmental}}

    Methods in this sub-category focus on a more shared-autonomy type of approach where they can learn a model from a tutor, or trial and error, to create a relation between object features and the affordance relation elements.
    Especially for the robotics task of grasping, we find methods that exploit the benefits of \ac{LbD} to build the model for this relation \cite{castellini2011using,stark2008functional,song2010learning,de2006learning,chan2014determining,ridge2013action,koppula2013learning}. \citet{castellini2011using} propose a model that learns from how a human would grasp an object, by matching the extracted object features to previously seen affordance relations. They represent this knowledge as a mapping of visual features of an object to the kinematic features of a hand. The hand features are extracted during grasp demonstrations using a glove. Also using hand demonstrations, but this time extracting \ac{2-D} data, \citet{stark2008functional} distinguish features of object parts that are optimal for manipulation by detecting the segments of human hands on such objects. An example is shown in Fig.~\ref{fig:stark2008functional}.
    \citet{song2010learning} define constraint functions given a set of object features and action attributes learned from tutors. For example, pouring water from a cup requires first opening the cup, and using a knife to cut means that the robot needs to grasp it from the handle part. 
    
    Another group of methods prefer a trial and error process to learn relations between similar features and actions-effects.
    This process allows the system to explore the success or failure of matching known actions on features that share some similarity with previous seen scenarios, thus being able to learn a model \cite{cos2004using,gonccalves2014learning,tikhanoff2013exploring,moldovan2014occluded,ugur2009affordance,kraft2009learning,detry2011learning,erkan2010learning,chu2016learning,hermans2013learning,dehban2016denoising,lopes2007affordance,baleia2015exploiting,kim2015interactive,ugur2007learning,ugur2007curiosity,montesano2009learning}. An example of this trial and error procedure is shown in Fig.~\ref{fig:detry2011learning}. In \citet{detry2011learning}, the system has previously seen objects that share similar shape features and the task is to detect the optimal grasping point on the familiar object, to which they refer to as an affordance. In this example, the green area represents the end-effector reachable space, while the red dot is the optimal affordance point of the object. These works then approach the exploration of objects as an action selection problem, i.e., they choose from a prior of actions-effects related to the invariants of the objects, thus creating a mapping from invariant features to actions and effects. This interaction with the environment allows the robot to learn object affordance relations by assessing the change in the state of the object when a certain behaviour is applied.
    
     Even though the works presented in this section collect data in a more human-like manner and generalise among familiar affordance tasks, they are still not able to explore new scenarios and form a conceptualisation of completely novel affordance relations. 

%% file: sections/5_exploratory.tex
\section{Novel Affordance Relations \label{sc:exploratory}}

Approaches in this categorisation can generalise affordance relations to unknown objects or scenarios. These works use a set of heuristics to learn the effects of different actions on unknown target objects from exploration and demonstration, thus building the affordance relation. In contrast to methodologies in Section~\ref{sc:model} and Section~\ref{sc:combined} that are built on prior information, the methods in this category have no prior on the best combination of target, action and effect that will guarantee the success of the robotics task. Moreover, the works in this category consider more than one action in their heuristic function thus allowing the system to create an abstract reasoning of the affordance relation. In this category, works have the following common features:
    \begin{itemize}
        \item To prove robustness, all the methods in this section experiment in cluttered environments.
        \item They generalise to unseen scenarios.
        \item They are inspired by developmental robotics and thus test their model in a more human-like manner.
    \end{itemize}
    
    The purpose of these works is to model complex learning using physical and social interaction to reach a more generalisable method. Thus, in this category we divide the methods in two: (i)~those that use \ac{RL} to learn the affordance relation through physical exploration (\textit{Section~\ref{ssc:RL}}), and (ii)~those that learn this relation through human demonstrations, \ac{LbD} (\textit{Section~\ref{ssc:lbd}}).
    
\begin{figure}[t!]
    \centering
    \includegraphics[width= 8.7cm]{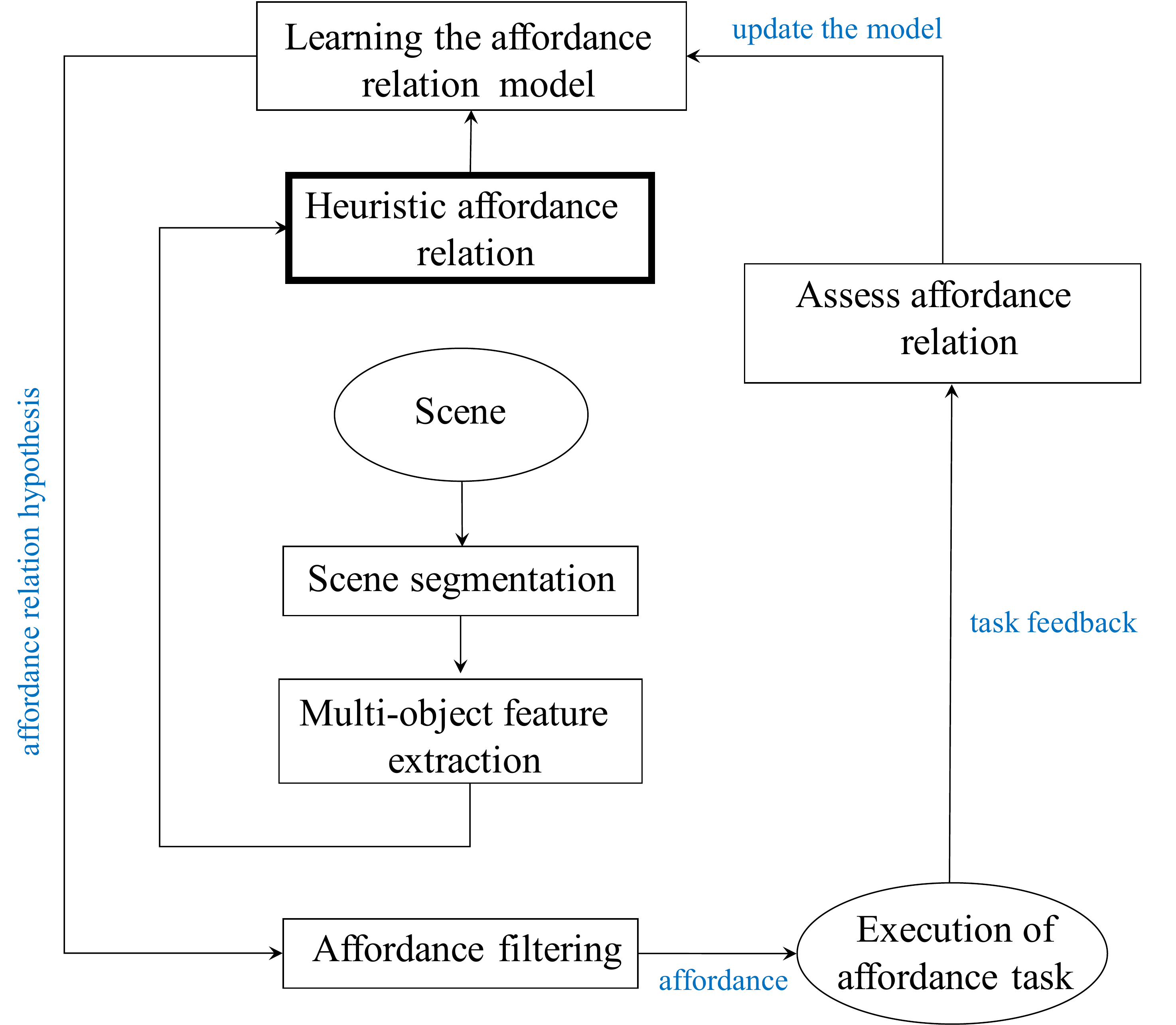}
    \caption{Flowchart for a typical system that has no \textit{a priori} information on the relation of the affordance task components. In this category, the relation is built using trial and error techniques together with imitation from human demonstrations \cite{ugur2007curiosity,koppula2016anticipatory,pieropan2014recognizing,kaiser2016towards,montesano2008learning,hermans2013decoupling}. The system starts by exploring from a heuristic model which is then updated with developmental techniques.}
    \label{fig:exploratory_data}
  \end{figure}

    Most of the methods in this category have a work flow similar to the one illustrated in Fig.~\ref{fig:exploratory_data}, where the system builds the affordance relation model through trial and error techniques and constant updates. Table \ref{tb:exploratory} shows a summary of the methods in this category.

\begin{table*}
    \centering
       \setlength\tabcolsep{5pt}
        \begin{tabular}{p{3.3cm}|l|c c c|c c c c|c c|c c c c|c c c}
        \hline
        \multirow{6}{*}{\multirow{6}{*}{{\textbf{Work}}}}&
        \multicolumn{1}{c|}{\multirow{6}{*}{{\textbf{Robotics Platform}}}}
        &\multicolumn{3}{c|}{\textbf{Robotics Task}}
        &\multicolumn{4}{c|}{\textbf{Perception}}
        &\multicolumn{2}{c|}{\textbf{Actions}}
        &\multicolumn{4}{c|}{\textbf{Means of Data}}
        &\multicolumn{3}{c}{\textbf{Relations}}\\
        \cline{3-18} & &
        \rotatebox[origin=c]{90}{Manipulation}&
        \rotatebox[origin=c]{90}{Navigation}&
        \rotatebox[origin=c]{90}{Action Prediction}&
        \rotatebox[origin=c]{90}{Visual}&
        \rotatebox[origin=c]{90}{Proprioception}&
        \rotatebox[origin=c]{90}{Kinesthetic}&
        \rotatebox[origin=c]{90}{Tactile}&
        \rotatebox[origin=c]{90}{Primitive}&
        \rotatebox[origin=c]{90}{Complex}&
        \rotatebox[origin=c]{90}{Labels}&
        \rotatebox[origin=c]{90}{Demonstrations}&
        \rotatebox[origin=c]{90}{Trial and Error}&
        \rotatebox[origin=c]{90}{Heuristics}&
        \rotatebox[origin=c]{90}{Deterministic}&
        \rotatebox[origin=c]{90}{Probabilistic}&
        \rotatebox[origin=c]{90}{Planning}
        \\
        \hline
        
         \citet{ugur2011unsupervised} & Kurt3D &  &\checkmark&& \checkmark & & && \checkmark & \checkmark&&&&\checkmark&\checkmark&&\checkmark \\ \hline
         
         \citet{ugur2011goal} & Gifu hand &\checkmark  & &  &\checkmark  & & &  &  &\checkmark & & && \checkmark &\checkmark &  &\checkmark \\ \hline
         
         \citet{ugur2015staged} & Gifu hand  & \checkmark& & & \checkmark & & && \checkmark &&&&&\checkmark&\checkmark\\ \hline
        
        \citet{ugur2015bottom} & Kuka arm &\checkmark  & &  &\checkmark  & & &  &  &\checkmark & & && \checkmark &\checkmark &  &\checkmark \\ \hline
         
         \mbox{\citet{montesano2007affordances}}  &  Baltazar &\checkmark&& & \checkmark & \checkmark && & \checkmark&&&&&\checkmark&&\checkmark\\ \hline
        
         \citet{montesano2007modeling} & Baltazar & \checkmark& &  & \checkmark & & & & \checkmark&&&&&\checkmark&&\checkmark \\ \hline
         
         \mbox{\citet{montesano2008learning}} & Baltazar & \checkmark&& & \checkmark && &\checkmark  & \checkmark&&&&&\checkmark&&\checkmark \\ \hline
         
         \citet{stoytchev2005toward} & CRS \& A251 & \checkmark&& & \checkmark & &&  & \checkmark&&&&&\checkmark&\checkmark \\ \hline
         
         \citet{hermans2013decoupling} & PR2 & \checkmark&&& \checkmark &&&  & \checkmark&&&&&\checkmark&\checkmark \\ \hline
         
         \citet{bierbaum2009grasp} & FRH-4 & \checkmark&& & \checkmark &&& \checkmark & \checkmark&&&&&\checkmark&\checkmark \\ \hline
         
         \citet{fitzpatrick2003learning} & BabyBot \& Cog &\checkmark&& & \checkmark &  & & & \checkmark&&&&&\checkmark&\checkmark\\ \hline
         
         \citet{koppula2014physically} & None--theoretical work  &&& \checkmark & \checkmark & & &&& \checkmark &&&&\checkmark&&\checkmark\\ \hline
         
         \mbox{\citet{koppula2016anticipatory}} & PR2 &&& \checkmark & \checkmark & &&  &\checkmark & \checkmark&&&&\checkmark&&\checkmark&\checkmark \\ \hline
         
         \citet{koppula2016anticipating} & PR2 &&& \checkmark  &\checkmark &  &&  & \checkmark & \checkmark&&&&\checkmark&&\checkmark&\checkmark\\ \hline
         
         \citet{jiang2013hallucinated} & PR2 &&& \checkmark  &\checkmark & & &  & \checkmark & \checkmark&&&&\checkmark&\checkmark&\\ \hline
         
         \citet{ivaldi2012perception} & iCub & \checkmark & & &\checkmark & && &\checkmark&&&&&\checkmark&\checkmark&& \\ \hline
         
        \citet{liu2018physical} & None--theoretical work&&& \checkmark & \checkmark & & & & \checkmark & \checkmark &&&&\checkmark&\checkmark&\\ \hline
         
          
         \citet{pieropan2014recognizing}  & None--theoretical work &&& \checkmark & \checkmark & & & &  &\checkmark&&&&\checkmark&&\checkmark \\ \hline
         
        \citet{kroemer2012kernel}  &  Allegro hand & \checkmark&& &\checkmark & \checkmark && & \checkmark & \checkmark&&&&\checkmark&&\checkmark\\ \hline
        
        \citet{kaiser2015validation} & Armar III &\checkmark &\checkmark & & \checkmark &\checkmark & &&& \checkmark & &&&\checkmark&\checkmark\\ \hline
        
        \citet{kaiser2016towards} & Armar III &\checkmark & & & \checkmark &\checkmark & &&& \checkmark & &&&\checkmark&\checkmark&&\checkmark\\ \hline
        
    \end{tabular}
    
    \caption{Summary of methods that have no \textit{a priori} information on the affordance components relation to achieve a task. \label{tb:exploratory}}
\end{table*}

\begin{figure}[ht!]
   \centering
    \subfigure[\citet{kroemer2012kernel} ]{\label{fig:kroemer} \includegraphics[width=7cm]{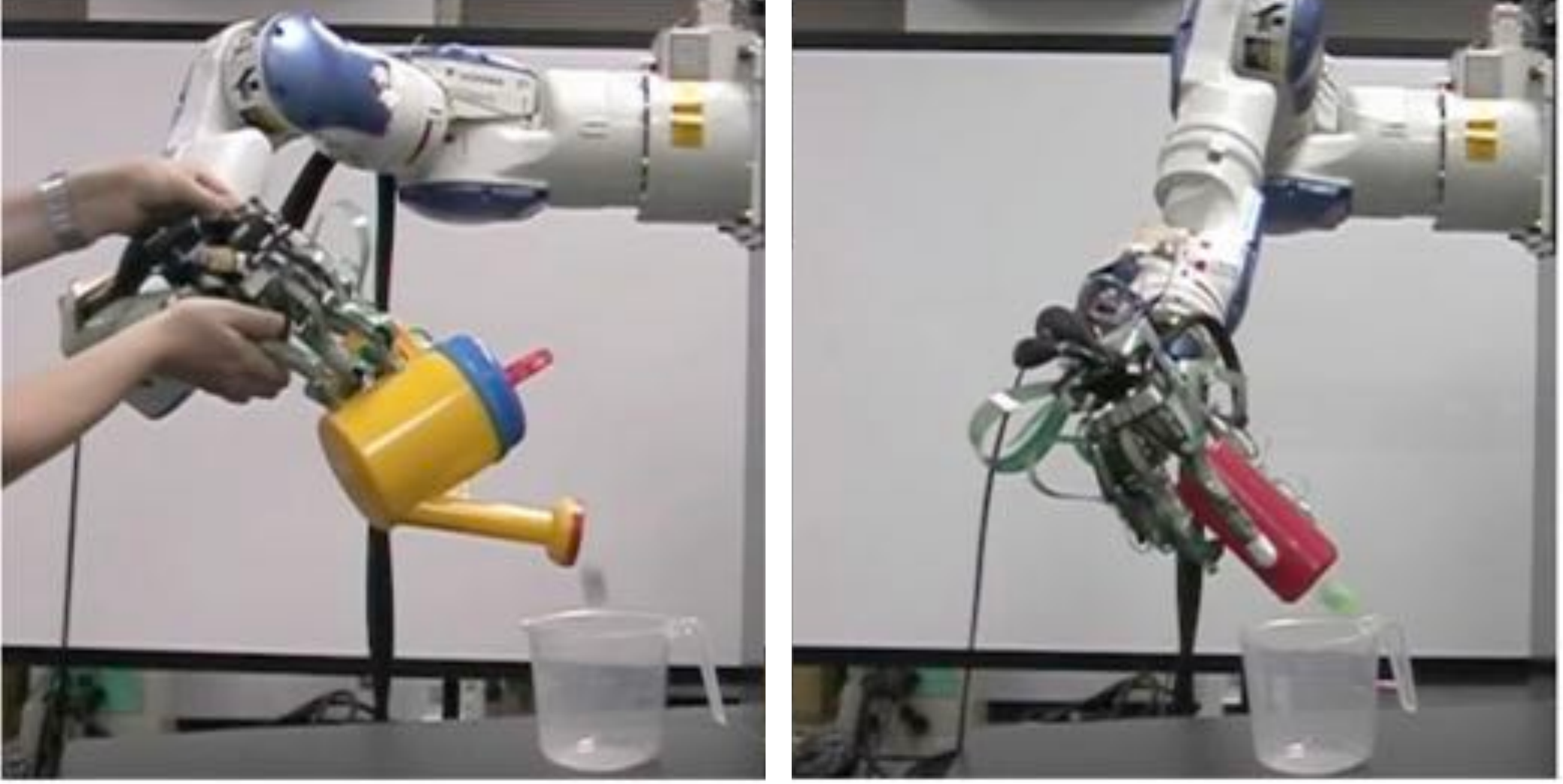}}
    
    \subfigure[\citet{ivaldi2012perception} ]{\label{fig:ivaldi} \includegraphics[width=7cm]{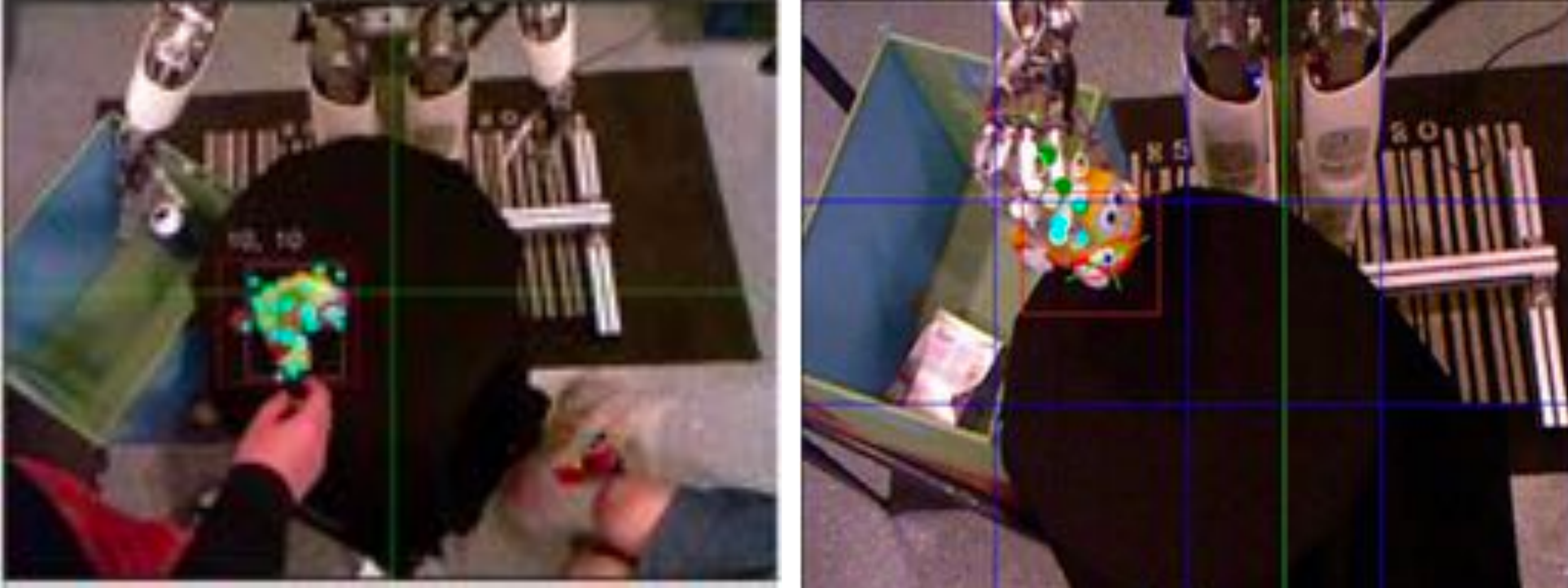}}

   \caption{Examples of methods that learn the affordance relation from heuristics and experiment on a demonstrated task. The images on the left column for (a) \citet{kroemer2012kernel} and (b) \citet{ivaldi2012perception} show a human demonstrating an affordance task to the robot. Once the system has trained and learned the proper policy for the affordance relation it replicates the task as shown in the images in the right column.
   \label{fig:exploratory_based}}
\end{figure}

\subsection{Using self-exploration to test affordance relations \label{ssc:RL}}

    The works in this section propose a framework that allows the robot to explore target objects, actions and effects to learn a model created from heuristic functions. \citet{ugur2015staged} present a behaviour development framework for a robot that learns progressively in stages. In the first stage, the robot is initialised with the basic motion primitives such as reach and enclose-on-contact movements to discover a set of action primitives by exploring on its own \cite{ugur2011goal}. In the next stage, the robot exercises the discovered actions on different novel objects and learns the action and effects relation \cite{ugur2011unsupervised}. Effectively, these learned relationships build a library of affordances that are used to bootstrap complex imitation and action learning with the help of a cooperative tutor \cite{ugur2015bottom,ugur2015staged}.
    
    \citet{kaiser2016towards} uses geometric primitives to extract the affordances of whole body motions such as support, lean and grasp. The final framework is also the result of work done in stages through \cite{kaiser2014extracting,kaiser2015validation} where heuristic rules are put in place to learn the affordance model. \citet{montesano2008learning} also build an affordance framework in stages with the goal of applying this approach to human-robot interaction tasks. In \cite{montesano2007affordances}, they start by proposing a formalism of affordance, as previously seen in Fig.~\ref{fig:montesano_d} and introduced in Section~\ref{sc:formalism}. They model the previously presented formalism using Bayesian networks to link target object features, actions and effects. Given that their learning is based on a purely probabilistic model they are able to generalise to new target objects \cite{montesano2007modeling}. These previous stages are then used to create a final framework that identifies reliable grasping points on novel objects \cite{montesano2008learning}. 

    The three aforementioned frameworks learn high-level behaviours, however, questions such as \textit{how does a robot learn to pull an object towards itself?} or \textit{how does the robot learn that spherical objects roll while a cube only slides when pushed?} concern learning of primitive actions at a control level. Some works learn the parameters to basic controller primitive actions to generalise to new affordance tasks by combining visual and tactile information and testing the heuristic model in a trial and error stage \cite{fitzpatrick2003learning,bierbaum2009grasp, stoytchev2005toward,hermans2013decoupling}.

\subsection{Using demonstrations to test affordance relations \label{ssc:lbd}}

    To achieve collaborative tasks, many works use demonstrations to predict actions from other agents using heuristics \cite{koppula2016anticipatory,pieropan2014recognizing,jiang2013hallucinated}. \citet{koppula2016anticipatory} present a framework that has been built in different sequential stages \cite{koppula2016anticipating,koppula2014physically,koppula2013learning}. 
    The work starts by extracting a descriptive labelling of the sequence of sub-activities being performed by a human and their interactions with the objects in the form of associated affordances \cite{koppula2013learning}. In \cite{koppula2014physically}, they create a state-space model that simulates a human's low-level kinematics and high-level intent which is then unified with the previously extracted labels of activities to create a complete framework that predicts human actions \cite{koppula2016anticipatory}.
    
    Using guidance from a tutor, \cite{liu2018physical,ivaldi2012perception,kroemer2012kernel} perform collaborative tasks among two agents. These works attempt to predict human demonstrations, object appearances and the effects of actions in order to understand the task. For example, \citet{liu2018physical} find the affordance of new objects through the decomposition of the object parts and use video frames to associate their effects when an action is applied. More examples are shown in Fig.~\ref{fig:exploratory_based}. 
    
    In general, methodologies in this category show interesting results in applications where the scenario is of a changing nature and the tasks are particularly collaborative. It is notable that fewer works fall into this categorisation and, as observed in the timeline presented in Fig.~\ref{fig:timeline}, it is still an emerging category.

%% file: sections/6_limitations.tex
\section{Limitations and open questions\label{sc:limitations}}

  Given that the concept of affordance derives strongly from the literature in psychology, it is natural that most of the research in robotics is aimed at trying to connect these conceptual abstractions with a mathematical formalism that enables implementation on robotic platforms.
  Many of the approaches have envisioned affordances perceived by robots as an imitation of how humans might learn about affordances during their development as infants \cite{min2016affordance}. Nonetheless, viewed primarily from the perspective of robotics tasks, there are still many aspects that hinder progress in this research area, which need to be standardised before moving forward as a developmental approach. Some of these aspects include: (i)~ambiguity regarding how affordances influence a task (\textit{Section~\ref{ssc:ambiguous}}), (ii)~lack of datasets that represent the affordance task components (\textit{Section~\ref{ssc:datasets}}), and (iii)~lack of standardised metrics to evaluate robotics tasks that require use of the concept of affordances (\textit{Section~\ref{ssc:metrics}}).
  Moreover, in this section we identify aspects of affordances that have been widely explored within robotics, as well as those that need more research so that their inclusion in models pushes the field forward.
  

\subsection{Ambiguous affordance tasks~\label{ssc:ambiguous}}

  It is still unclear how to generalise the relationship between affordance components: target object, actions and effects. As presented in Section~\ref{sc:formalism}, there are many formalisms in the field with only some of them agreeing on these previously mentioned three elements. Later in this survey, we observed how some methods have gone beyond the target, action and effect relation and have integrated into their models other associations such as object-object or agent-object-agent which is especially helpful when performing collaborative and human-robot interaction tasks that include some scene understanding \cite{jiang2013hallucinated,pieropan2014recognizing,pandey2013affordance,sun2014object}.
  
 Moreover, the scope in which affordances are included in the different robotic application tasks is not clear. For example, there are methods where affordances are used only to contextualise the task of other methodologies, such as for control. \citet{diana2013deformable} present a method for controlling a robotic swarm by identifying and matching the different shapes created with a piece of clay, which they refer to as affordances. Nonetheless, in their method, the affordance concept does not contribute further or differently than if using any other shape identification, which they recognise in their results. Other methodologies such as the one developed in \citet{ugur2007curiosity} use the concept of affordance to learn the traversability of objects, depending on their shape, in a navigation application. They explore how, for example, cylindrical and spherical objects roll, thus the vehicle learns to push them and continue navigating on the pre-planned path, while the same idea cannot be replicated with cubic objects. In this case, the navigation task is built around the concept of the traversability affordance of the obstacles. Thus, the scope for robotic application tasks that include affordances is not clear. This is one of the main issues that needs to be addressed to guarantee the progression of the field.

\subsection{Datasets\label{ssc:datasets}}

    \begin{table*}
        \centering
        \begin{tabular}{p{2cm}|l|l|l}
        \hline
        \textbf{Work} & \textbf{Robotics Task} & \textbf{Contains} & \textbf{Dataset location} \\ \hline
      
         \citet{do2017affordancenet} & Action Prediction &10 objects / 9 affordances& \url{http://sites.google.com/site/iitaffdataset/} \\ \hline
         \citet{myers2015affordance} & Manipulation& 105 tool-handles / 3 cluttered scenes& \url{http://users.umiacs.umd.edu/~fer/affordance/part-affordance-dataset/} \\ \hline
         \citet{jiang2013hallucinated} &Action Prediction& 47 objects / 4 scenes / 6 human poses& \url{http://pr.cs.cornell.edu/hallucinatinghumans/}  \\\hline
         \mbox{\citet{koppula2016anticipatory}} &Action Prediction& 10 activies with human poses& \url{http://pr.cs.cornell.edu/humanactivities/} \\ \hline
         
         \citet{ardon2019learning} &Manipulation&30 objects / 7 scenes / 14 affordances& \url{https://paolaardon.github.io/grasp_affordance_reasoning/} \\ \hline
         
          \citet{chu2019learning} &Manipulation& 10 objects / 7 affordances& \url{https://www.dropbox.com/s/ldapcpanzqdu7tc/models.zip} \\ \hline
         
        \end{tabular}
        \caption{Summary of available datasets for different robotics tasks that include affordances.\label{tb:datasets}}
    \end{table*}

 A robotic application that includes affordances commonly encapsulates different sub-tasks ranging from object recognition (either visual or through tactile sensors) to action deployment. Finding a suitable dataset that includes all the affordance components related to the task is challenging. As a consequence, the literature makes non-real-world assumptions resulting in the methods being deployable only in controlled environments. Some of these assumptions include: (i)~assigning a single affordance to an object, and (ii)~assuming this single affordance is true no matter the context of the object. Unlike other research fields which have many datasets available online, such as the case for grasping that has over $30$ online datasets as summarised in \cite{huang2016recent}, for the affordance task the available datasets are very few. Table~\ref{tb:datasets} shows a summary of the datasets and their online location.
  
   Works such as \cite{do2017affordancenet,chu2019learning} and \cite{myers2015affordance} present mapping labels from object categories to affordances. The dataset in \cite{do2017affordancenet} contains $10$ object categories and $9$ affordance classes, where each object is mapped to an affordance. \citet{chu2019learning} present a synthetic version of the University of Maryland dataset for manipulation. This dataset is collected for autogenerating annotated, synthetic input data of objects and segmenting their parts according to their affordances.
   \citet{myers2015affordance} provide \ac{RGB-D} images and ground truth affordance labels for $105$ kitchen, workshop and gardening tools in $3$ cluttered scenes. This dataset presents a large and diverse collection of everyday tools, where tools from different categories share the same affordance. They include $7$ affordances associated with tool parts alongside a ranking to choose the most likely affordance of each part. 
  
  In \citet{jiang2013hallucinated} and \citet{koppula2016anticipatory}, besides the object category information, the datasets contain human skeletons to indicate the affordance of the objects. This addition is useful when the task is partly concerned to predict actions from a second agent. The dataset in \cite{jiang2013hallucinated} has $20$ scenes from $3$ categories and $47$ objects of $19$ types. Their point-cloud has labels for human poses and object locations indicating the affordance of the objects. In \cite{koppula2016anticipatory}, the dataset has $120$ RGB-D videos of long daily activities mapped with $12$ object affordance labels with their corresponding tracked skeletons. 

  
  To answer questions such as \textit{how does the affordance change in different environments?} and \textit{how to relate different grasp regions to different affordances?} \citet{ardon2019learning} present a dataset that relates different grasping regions of an object with corresponding affordances correlated with different indoor scenes. This dataset contains: $30$ different objects, $7$ indoor scenes, $3$ object attributes with their corresponding categories and $14$ different affordances closely related to a grasping area. Nonetheless, this dataset has been designed for grasping applications and improvements to go across the navigation and human action prediction need to be done.
  
\subsection{Metrics\label{ssc:metrics}}

\begin{figure}[ht!]
  
      \centering
       \subfigure[\citet{aldoma2012supervised} ]{\label{fig:aldoma_analysis} \includegraphics[width=4cm]{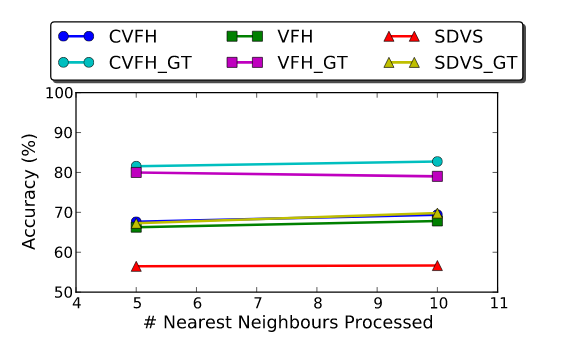}}
       \subfigure[\citet{bohg2010learning} ]{\label{fig:bohg_analysis} \includegraphics[width=4cm]{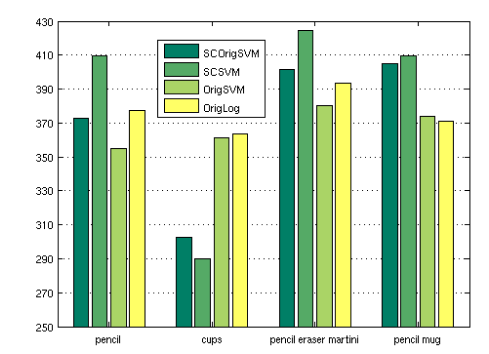}}
       \subfigure[\citet{fallon2015architecture} ]{\label{fig:fallon_analysis} \includegraphics[width=8cm]{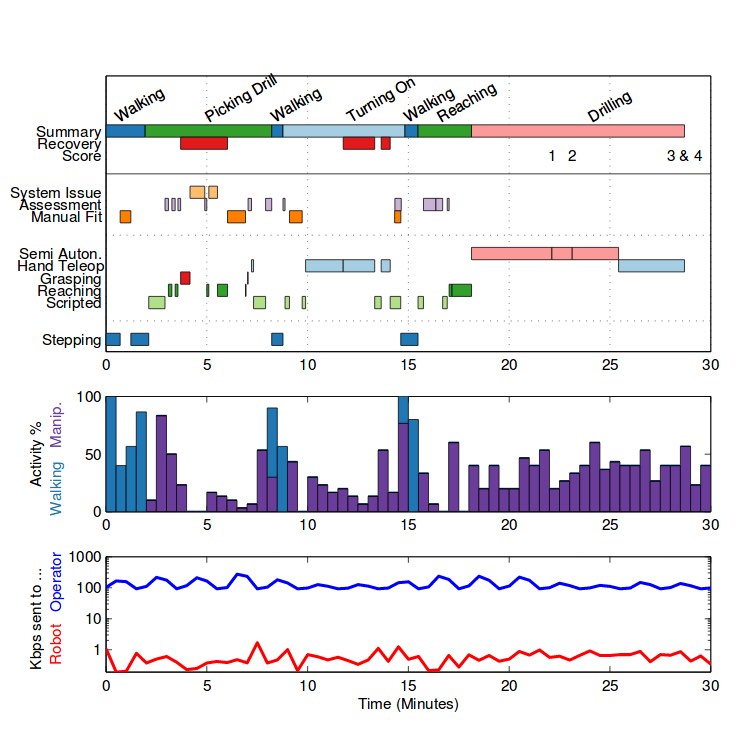}}
  
     \caption{Examples of intrinsic and extrinsic evaluations. (a)~\citet{aldoma2012supervised} evaluate the accuracy of the affordance detection using several descriptors and nearest neighbours. (b)~\citet{bohg2010learning} calculate the accuracy of detection of the best ten grasping points of each shape model on a picture test set in the task of learning grasping points from shape context. (c) \citet{fallon2015architecture} use the affordance concept on a drill task for the Darpa competition, as an example of extrinsic evaluation. The graph on the top shows the analysis for the operating time for the different sub-tasks. The graph in the centre shows the time the robot spent moving and manipulating the object. The bottom graph shows the data transmission rates from the robot to the operator (blue), and vice versa (red), which provides an estimation of the teleoperation time to complete a robotics task that includes affordances.}
     \label{fig:accuracy_of_detection_analysis}
   \end{figure}

  Affordance in robotics is a relatively young field. As highlighted in this survey, a wide variety of approaches address this challenge in different ways. Given that an affordance problem is highly correlated to the task, to date there exists little direct comparison for the different approaches. As a result of this variety, such approaches use ad-hoc metrics that fit their application needs. The vast majority use popular classification evaluation as the first step in recognising the target of the affordance task. Therefore, popular metrics seen in the field are confusion matrices, \ac{MSE}, and accuracy of classification metrics that reflect intrinsic assessments. Examples of such metrics used in affordance tasks are shown in Fig.~\ref{fig:aldoma_analysis} and \ref{fig:bohg_analysis}.

\begin{figure}[t!]
    \centering
     \subfigure[\citet{bohg2010learning} ]{\label{fig:bohg_table} \includegraphics[width=6cm]{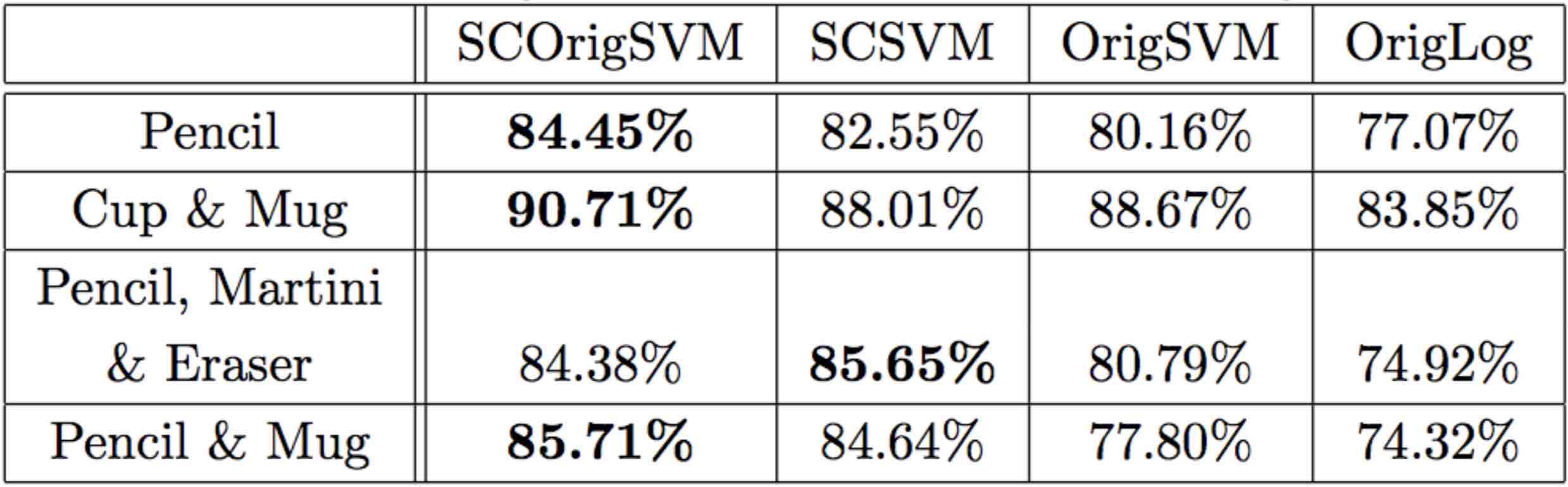}}

    \subfigure[\citet{kroemer2012kernel} ]{\label{fig:kroemer_analysis} \includegraphics[width=7cm]{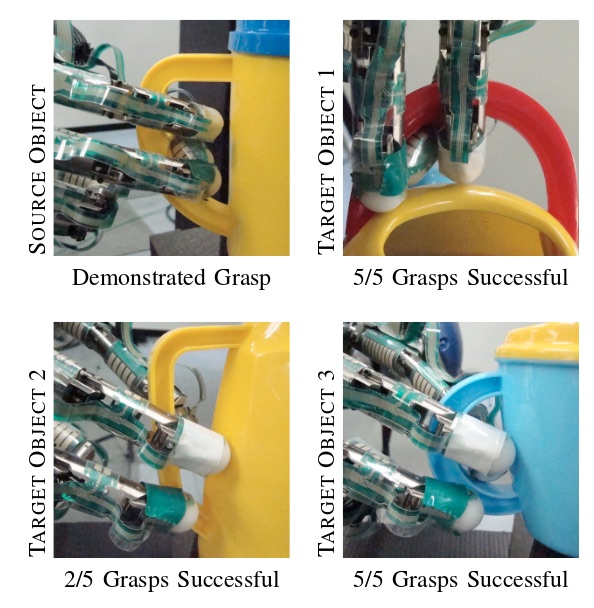}}

    \caption{Examples of evaluations for the grasp affordance task. (a)~\citet{bohg2010learning} evaluate the accuracy of the different models to classify grasping regions based on shape context. (b)~\citet{kroemer2012kernel} evaluate the grasp affordance generalisation of the learning method with familiar objects.}
    \label{fig:katz_and_kroemer_analysis}
  \end{figure}

 When it comes to the performance of the affordance task as a whole, approaches addressing applications such as navigation or action prediction for collaboration, tend to measure the quality of the method extrinsically. Namely, they judge the quality of the technique based on how it affects the completion of the task. Thus they measure the completion time and, usually, the percentage of collisions, as shown in Fig.~\ref{fig:fallon_analysis}. A similar scenario is presented for those focused on the grasping task, where a combination of intrinsic (i.e., based on the grasping detection and classification performance) and extrinsic evaluations are widely used. Examples of evaluation for the grasp affordance task are shown in Fig.~\ref{fig:katz_and_kroemer_analysis}.
  
  Nonetheless, the affordance task lacks a standard set of evaluation metrics, which further complicates the comparisons across algorithms and domains. Given the ambiguity of the affordance concept, it is not surprising the area lacks a set of benchmark methods. Metrics that reflect the performance of the robotics tasks when including affordances are fundamental for the progress of the field. Given the field's motivation lies in collaboration tasks with other agents and reaching human-like performance, an interesting approach would be if this metric can reflect and compare the similarity of the actions taken by the system with those a human would execute. Options such as the Hausdorff distance and the Kullback-Leibler divergence are interesting to explore. The Hausdorff distance measures how similar or close two sets of points are and the Kullback-Leibler divergence measures how one probability distribution is different from a second probability distribution. Including such evaluations would be a good assessment of the performance on the affordance task in relation to the ground truth data, regardless of the learning algorithm.

    \subsection{Study of aspects that influence affordances in robotics\label{ssc:summary}}
    
\begin{figure}[t!]
    \centering
    \includegraphics[width= 9.5cm]{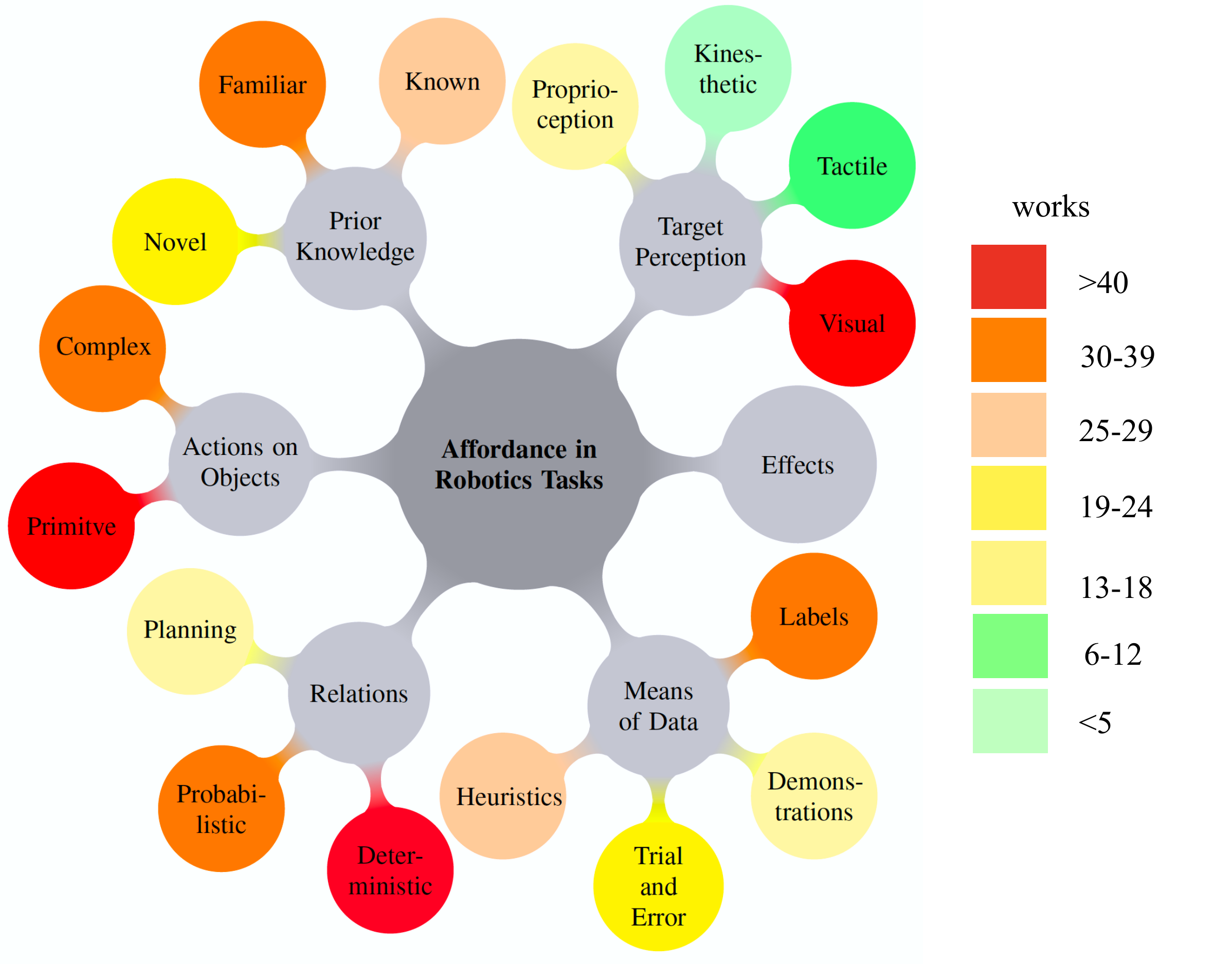}
    \caption{Population map of the different aspects that influence a robotics task that includes affordances, based on approaches reviewed from the literature. The warmer the colour (red) the more works that use that element to associate affordances, while the colder the colour (green), the less that element is used.}
    \label{fig:coverage}
\end{figure}

    Throughout this survey, we have explored the differences and similarities of how approaches in the literature represent affordances in their robotics tasks. The research spreads across different aspects as shown in Fig.~\ref{fig:intro}. Nonetheless, some of these aspects are very popular areas of study while others lack further research. From Tables~\ref{tb:known_data}, \ref{tb:combined_online_actions} and \ref{tb:exploratory} we can gather a population of methodologies across these elements as shown in Fig.~\ref{fig:coverage}, where the warmer the colour the more used the aspect is across the literature. For example,  most of the works emphasise the learning of primitive actions as affordances (i.e., push, poke, lift among others), using visual perception and labels from images to get an affordance per target object, building the affordance relation deterministically. On the opposite side, colder coloured elements, indicate there are very few works that exploit learning affordance trajectories in the form of motions (using kinesthetic sensing), as well as those that exploit a multi-step prediction to achieve the tasks in a planning manner. Certainly some of these components are highly dependent on hardware robustness more than others. For example perceiving a target using kinesthetic teaching or tactile sensing is highly reliable accurate sensor readings.
    Nonetheless, studying such aspects in greater depth would improve their inclusion in robotics tasks as well as provide valuable insights for collaboration activities and task replication across different agents. 
    
%% file: sections/open_questions.tex
\section{Final Notes\label{sc:openquestions}}

In this survey, we review methods that include the affordance concept in their robotics tasks. In contrast to previous reviews of affordance in the field that primarly focus on the influences from psychology, we propose a categorisation based on the level of prior knowledge to build the relation among affordance elements. We identify target object, action and effects to be the three main components historically considered by all formalisms.
We pinpoint several problems in the field such as the ambiguity of the usage of the affordance concept in robotics tasks, as well as the lack of standardised datasets and metrics to evaluate across applications. Moreover, we identify the areas where there has been extensive research and those where it has been notably sparse, thus providing suggestions on areas of improvement.
As illustrated in Section~\ref{sc:formalism} with Fig.~\ref{fig:timeline} and throughout this survey, it is clear that the subject of affordances in robotics started with a strong influence from psychology in an attempt to achieve human-like generalisation capabilities. The progress towards this generalisation is also notable, from the rise of works as presented in Section~\ref{sc:model} to those in Section~\ref{sc:exploratory}. 
On the downside, the lack of activities organised to help achieve standards in the field is also notable, such as the late appearance of the few available datasets as well as the rare workshops that motivate the gathering of researchers in the community.

%% file: sections/biography.tex

\begin{IEEEbiography}
    [{\includegraphics[width=1in,height=1.25in,clip,keepaspectratio]{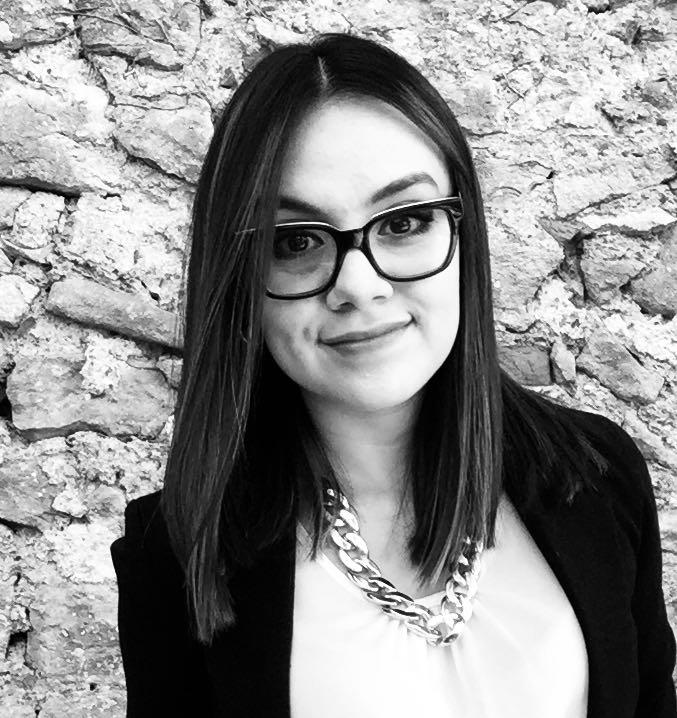}}]{Paola Ard\'on}
is a a Ph.D candidate at the Edinburgh Centre for Robotics (ECR) member of the Robust Autonomy and Decisions (Rad) at University of Edinburgh and Social robotics group at Heriot-Watt University. She holds an Erasmus Mundus M.Sc. in Computer Vision and Robotics (VIBOT) from the University of Burgundy (France), University of Girona (Spain) and Heriot-Watt University (UK) and an M.Sc. in Robotics and Autonomous Systems from The University of Edinburgh and Heriot-Watt University. Her research is about including the affordances in a robust autonomous framework for assistive robots specially for the application of grasping objects in indoor environments.
\end{IEEEbiography}

\begin{IEEEbiography}
    [{\includegraphics[width=1in,height=1.25in,clip,keepaspectratio]{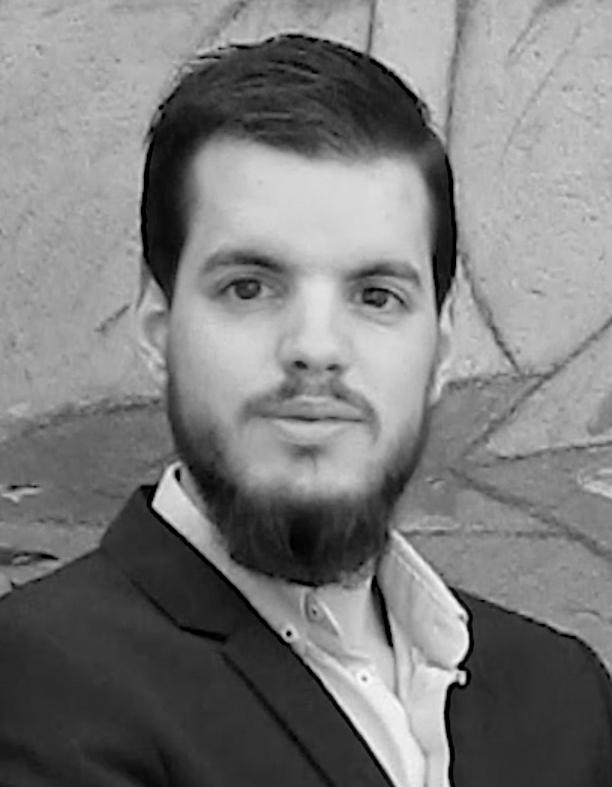}}]{\`Eric Pairet}
is a PhD candidate at the Edinburgh Centre for Robotics, a consortium between The University of Edinburgh and Heriot-Watt University. He holds an Erasmus Mundus M.Sc. in Computer Vision and Robotics (VIBOT) from the University of Burgundy (France), University of Girona (Spain) and Heriot-Watt University (UK) and an M.Sc. in Robotics and Autonomous Systems from The University of Edinburgh and Heriot-Watt University. His research interests lies in techniques capable of leveraging prior experiences to efficiently define robotics behaviour, particularly for motion planning applications such as object manipulation and robot navigation.
\end{IEEEbiography}

\begin{IEEEbiography}
    [{\includegraphics[width=1in,height=1.25in,clip,keepaspectratio]{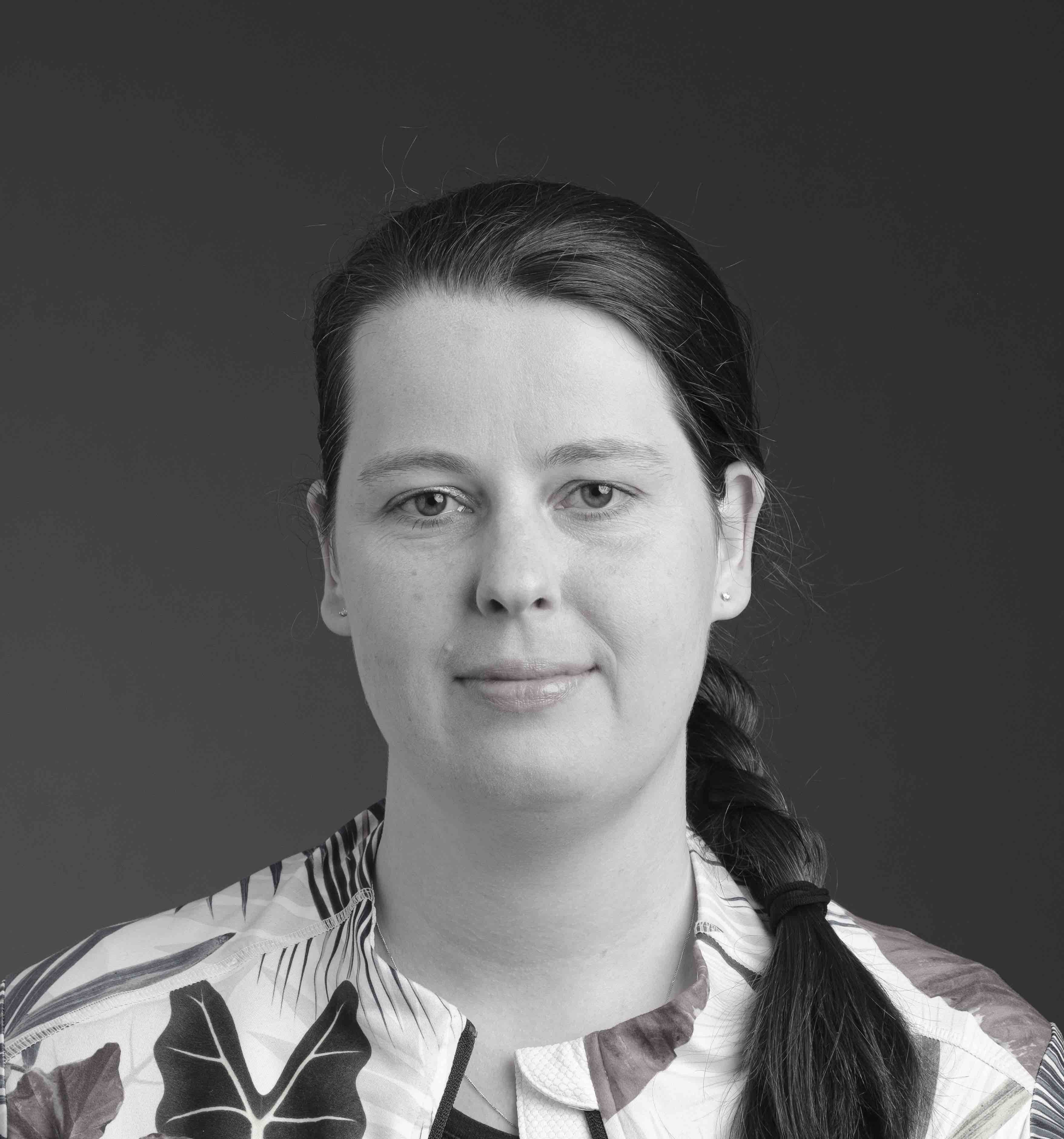}}]{Katrin Lohan}
is Professor for Robotics and Automation at the Institute for Development of Mechatronic Systems (EMS) at the University of Applied Sciences in Technology (NTB) in Buchs, Switzerland since 2019. She is an Associate Professor at the School of Mathematical and Computer Sciences at Heriot-Watt University where she joined in 2013. She is Director of the Social Robotics Group, and she is a member of the Edinburgh Centre for Robotics at Heriot-Watt University. Previously, she was working at the Italian Institute of Technology (IIT) as a Post Doc in the RobotDoc project funded by a Marie Curie RTN. 
She obtained her Ph.D. in Engineering from Bielefeld University, Germany in 2012, where she also worked as an RA in the ITALK project. 
Her main research interests are inspired by developmental robotics to allow for a natural interaction with a robot. From this natural interactions, robots shall learn semantics about objects, both through vision and speech.
\end{IEEEbiography}

\begin{IEEEbiography}
    [{\includegraphics[width=1in,height=1.25in,clip,keepaspectratio]{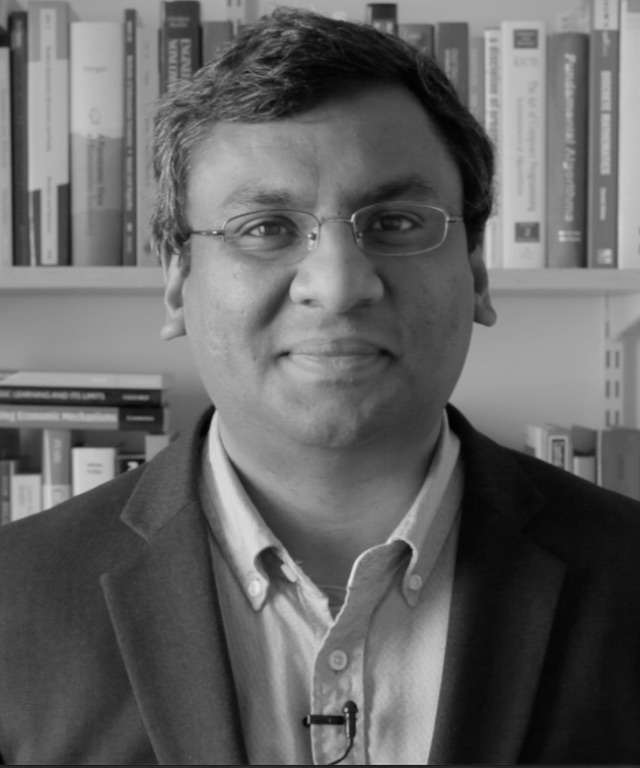}}]{Subramanian Ramamoorthy}
is a Reader (Associate Professor) in Robotics within the School of Informatics, University of Edinburgh, where he has been on the faculty since 2007. He is a Turing Fellow at the Alan Turing Insatitute and an Executive Committee Member for the Edinburgh Centre for Robotics. He received his PhD in Electrical and Computer Engineering from The University of Texas at Austin in 2007. He has been a Member of the Young Academy of Scotland at the Royal Society of Edinburgh, and has held Visiting Professor positions at the University of Rome ``La Sapienza'' and at Stanford University. He serves as Vice President - Prediction and Planning at FiveAI, a UK-based startup company focused on developing a technology platform for autonomous vehicles. His research focus is on robot learning and decision-making under uncertainty, with particular emphasis on achieving safe and robust autonomy in human-centered environments. 
\end{IEEEbiography}

\begin{IEEEbiography}
    [{\includegraphics[width=1in,height=1.25in,clip,keepaspectratio]{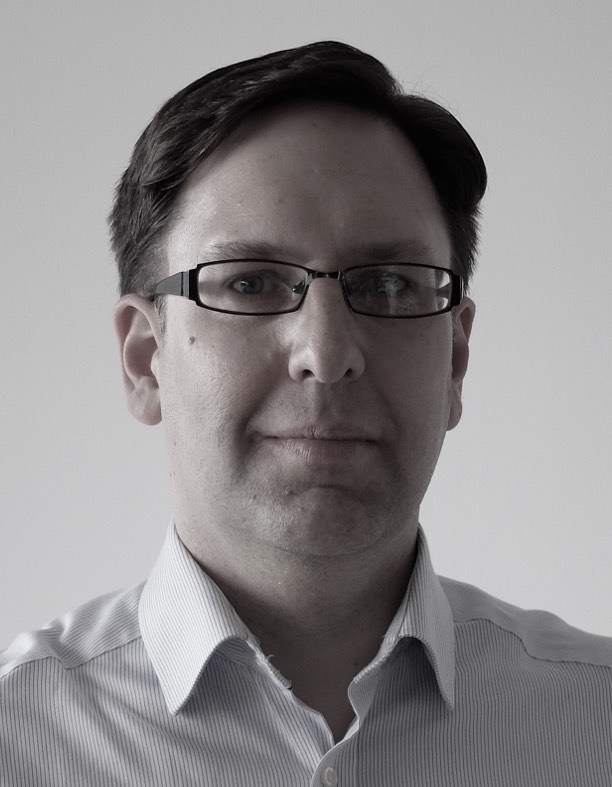}}]{Ronald P.A Petrick}
 is an Associate Professor in the School of Mathematical and Computer Sciences at Heriot-Watt University and a member of the Edinburgh Centre for Robotics. He received his PhD in Computer Science from the University of Toronto and, prior to joining Heriot-Watt, was a Research Fellow in the School of Informatics at the University of Edinburgh. He is currently an Honorary Fellow of the University of Edinburgh and recently held a Visiting Professor position at Sapienza University of Rome. He is a Co-Chair of the AAAI Symposium Series and Coordinator of the UK Planning and Scheduling Special Interest Group. His research interests include automated planning, cognitive robotics, knowledge representation and reasoning, action model learning, and human-robot interaction, with a focus on applications of planning in complex environments with humans and robots.
 \end{IEEEbiography}

%% file: main.bbl
\begin{thebibliography}{115}
\providecommand{\natexlab}[1]{#1}
\providecommand{\url}[1]{#1}
\csname url@samestyle\endcsname
\providecommand{\newblock}{\relax}
\providecommand{\bibinfo}[2]{#2}
\providecommand{\BIBentrySTDinterwordspacing}{\spaceskip=0pt\relax}
\providecommand{\BIBentryALTinterwordstretchfactor}{4}
\providecommand{\BIBentryALTinterwordspacing}{\spaceskip=\fontdimen2\font plus
\BIBentryALTinterwordstretchfactor\fontdimen3\font minus
  \fontdimen4\font\relax}
\providecommand{\BIBforeignlanguage}[2]{{%
\expandafter\ifx\csname l@#1\endcsname\relax
\typeout{** WARNING: IEEEtranSN.bst: No hyphenation pattern has been}%
\typeout{** loaded for the language `#1'. Using the pattern for}%
\typeout{** the default language instead.}%
\else
\language=\csname l@#1\endcsname
\fi
#2}}
\providecommand{\BIBdecl}{\relax}
\BIBdecl

\bibitem[Abelha et~al.(2016)Abelha, Guerin, and Schoeler]{abelha2016model}
P.~Abelha, F.~Guerin, and M.~Schoeler, ``A model-based approach to finding
  substitute tools in 3d vision data,'' in \emph{2016 IEEE International
  Conference on Robotics and Automation (ICRA)}.\hskip 1em plus 0.5em minus
  0.4em\relax IEEE, 2016, pp. 2471--2478.

\bibitem[Aksoy et~al.(2011)Aksoy, Abramov, D{\"o}rr, Ning, Dellen, and
  W{\"o}rg{\"o}tter]{aksoy2011learning}
E.~E. Aksoy, A.~Abramov, J.~D{\"o}rr, K.~Ning, B.~Dellen, and
  F.~W{\"o}rg{\"o}tter, ``Learning the semantics of object--action relations by
  observation,'' \emph{The International Journal of Robotics Research},
  vol.~30, no.~10, pp. 1229--1249, 2011.

\bibitem[Aksoy et~al.(2015)Aksoy, Tamosiunaite, and
  W{\"o}rg{\"o}tter]{aksoy2015model}
E.~E. Aksoy, M.~Tamosiunaite, and F.~W{\"o}rg{\"o}tter, ``Model-free
  incremental learning of the semantics of manipulation actions,''
  \emph{Robotics and Autonomous Systems}, vol.~71, pp. 118--133, 2015.

\bibitem[Aldoma et~al.(2012)Aldoma, Tombari, and Vincze]{aldoma2012supervised}
A.~Aldoma, F.~Tombari, and M.~Vincze, ``Supervised learning of hidden and
  non-hidden 0-order affordances and detection in real scenes,'' in
  \emph{Robotics and Automation (ICRA), 2012 IEEE International Conference
  on}.\hskip 1em plus 0.5em minus 0.4em\relax IEEE, 2012, pp. 1732--1739.

\bibitem[Antunes et~al.(2016)Antunes, Jamone, Saponaro, Bernardino, and
  Ventura]{antunes2016human}
A.~Antunes, L.~Jamone, G.~Saponaro, A.~Bernardino, and R.~Ventura, ``From human
  instructions to robot actions: Formulation of goals, affordances and
  probabilistic planning,'' in \emph{Robotics and Automation (ICRA), 2016 IEEE
  International Conference on}.\hskip 1em plus 0.5em minus 0.4em\relax IEEE,
  2016, pp. 5449--5454.

\bibitem[Ard{\'o}n et~al.(2019)Ard{\'o}n, Pairet, Petrick, Ramamoorthy, and
  Lohan]{ardon2019learning}
P.~Ard{\'o}n, {\`E}.~Pairet, R.~P.~A. Petrick, S.~Ramamoorthy, and K.~S. Lohan,
  ``Learning grasp affordance reasoning through semantic relations.''\hskip 1em
  plus 0.5em minus 0.4em\relax IEEE Robotics and Automation Letters, 2019.

\bibitem[Asada et~al.(2009)Asada, Hosoda, Kuniyoshi, Ishiguro, Inui, Yoshikawa,
  Ogino, and Yoshida]{asada2009cognitive}
M.~Asada, K.~Hosoda, Y.~Kuniyoshi, H.~Ishiguro, T.~Inui, Y.~Yoshikawa,
  M.~Ogino, and C.~Yoshida, ``Cognitive developmental robotics: A survey,''
  \emph{IEEE transactions on autonomous mental development}, vol.~1, no.~1, pp.
  12--34, 2009.

\bibitem[Baleia et~al.(2015)Baleia, Santana, and Barata]{baleia2015exploiting}
J.~Baleia, P.~Santana, and J.~Barata, ``On exploiting haptic cues for
  self-supervised learning of depth-based robot navigation affordances,''
  \emph{Journal of Intelligent \& Robotic Systems}, vol.~80, no. 3-4, pp.
  455--474, 2015.

\bibitem[Barck-Holst et~al.(2009)Barck-Holst, Ralph, Holmar, and
  Kragic]{barck2009learning}
C.~Barck-Holst, M.~Ralph, F.~Holmar, and D.~Kragic, ``Learning grasping
  affordance using probabilistic and ontological approaches,'' in
  \emph{Advanced Robotics, 2009. ICAR 2009. International Conference on}.\hskip
  1em plus 0.5em minus 0.4em\relax IEEE, 2009, pp. 1--6.

\bibitem[Bekiroglu et~al.(2013)Bekiroglu, Smith, Karayiannidis, Kragic,
  et~al.]{bekiroglu2013predicting}
Y.~Bekiroglu, C.~Smith, Y.~Karayiannidis, D.~Kragic \emph{et~al.}, ``Predicting
  slippage and learning manipulation affordances through gaussian process
  regression,'' in \emph{2013 13th IEEE-RAS International Conference on
  Humanoid Robots (Humanoids)}.\hskip 1em plus 0.5em minus 0.4em\relax IEEE,
  2013, pp. 462--468.

\bibitem[Bierbaum et~al.(2009)Bierbaum, Rambow, Asfour, and
  Dillmann]{bierbaum2009grasp}
A.~Bierbaum, M.~Rambow, T.~Asfour, and R.~Dillmann, ``Grasp affordances from
  multi-fingered tactile exploration using dynamic potential fields,'' in
  \emph{Humanoid Robots, 2009. Humanoids 2009. 9th IEEE-RAS International
  Conference on}.\hskip 1em plus 0.5em minus 0.4em\relax IEEE, 2009, pp.
  168--174.

\bibitem[Bohg and Kragic(2010)]{bohg2010learning}
J.~Bohg and D.~Kragic, ``Learning grasping points with shape context,''
  \emph{Robotics and Autonomous Systems}, vol.~58, no.~4, pp. 362--377, 2010.

\bibitem[Bohg et~al.(2017)Bohg, Hausman, Sankaran, Brock, Kragic, Schaal, and
  Sukhatme]{bohg2017interactive}
J.~Bohg, K.~Hausman, B.~Sankaran, O.~Brock, D.~Kragic, S.~Schaal, and G.~S.
  Sukhatme, ``Interactive perception: Leveraging action in perception and
  perception in action,'' \emph{IEEE Transactions on Robotics}, vol.~33, no.~6,
  pp. 1273--1291, 2017.

\bibitem[Castellini et~al.(2011)Castellini, Tommasi, Noceti, Odone, and
  Caputo]{castellini2011using}
C.~Castellini, T.~Tommasi, N.~Noceti, F.~Odone, and B.~Caputo, ``Using object
  affordances to improve object recognition,'' \emph{IEEE Transactions on
  Autonomous Mental Development}, vol.~3, no.~3, pp. 207--215, 2011.

\bibitem[Chan et~al.(2014)Chan, Kakiuchi, Okada, and
  Inaba]{chan2014determining}
W.~P. Chan, Y.~Kakiuchi, K.~Okada, and M.~Inaba, ``Determining proper grasp
  configurations for handovers through observation of object movement patterns
  and inter object interactions during usage,'' in \emph{2014 IEEE/RSJ
  International Conference on Intelligent Robots and Systems}.\hskip 1em plus
  0.5em minus 0.4em\relax IEEE, 2014, pp. 1355--1360.

\bibitem[Chemero(2003)]{chemero2003outline}
A.~Chemero, ``An outline of a theory of affordances,'' \emph{Ecological
  psychology}, vol.~15, no.~2, pp. 181--195, 2003.

\bibitem[Chemero and Turvey(2007)]{chemero2007gibsonian}
A.~Chemero and M.~T. Turvey, ``Gibsonian affordances for roboticists,''
  \emph{Adaptive Behavior}, vol.~15, no.~4, pp. 473--480, 2007.

\bibitem[Chu et~al.(2019)Chu, Xu, and Vela]{chu2019learning}
F.-J. Chu, R.~Xu, and P.~A. Vela, ``Learning affordance segmentation for
  real-world robotic manipulation via synthetic images,'' \emph{IEEE Robotics
  and Automation Letters}, vol.~4, no.~2, pp. 1140--1147, 2019.

\bibitem[Chu et~al.(2016)Chu, Fitzgerald, and Thomaz]{chu2016learning}
V.~Chu, T.~Fitzgerald, and A.~L. Thomaz, ``Learning object affordances by
  leveraging the combination of human-guidance and self-exploration,'' in
  \emph{2016 11th ACM/IEEE International Conference on Human-Robot Interaction
  (HRI)}.\hskip 1em plus 0.5em minus 0.4em\relax IEEE, 2016, pp. 221--228.

\bibitem[Cos-Aguilera et~al.(2004)Cos-Aguilera, Hayes, and
  Canamero]{cos2004using}
I.~Cos-Aguilera, G.~Hayes, and L.~Canamero, ``Using a sofm to learn object
  affordances,'' in \emph{Procs 5th Workshop of Physical Agents
  (WAF'04)}.\hskip 1em plus 0.5em minus 0.4em\relax University of Edinburgh,
  2004.

\bibitem[Cruz et~al.(2016)Cruz, Magg, Weber, and Wermter]{cruz2016training}
F.~Cruz, S.~Magg, C.~Weber, and S.~Wermter, ``Training agents with interactive
  reinforcement learning and contextual affordances,'' \emph{IEEE Transactions
  on Cognitive and Developmental Systems}, vol.~8, no.~4, pp. 271--284, 2016.

\bibitem[Cutsuridis and Taylor(2013)]{cutsuridis2013cognitive}
V.~Cutsuridis and J.~G. Taylor, ``A cognitive control architecture for the
  perception--action cycle in robots and agents,'' \emph{Cognitive
  Computation}, vol.~5, no.~3, pp. 383--395, 2013.

\bibitem[Dag et~al.(2010)Dag, Atil, Kalkan, and Sahin]{dag2010learning}
N.~Dag, I.~Atil, S.~Kalkan, and E.~Sahin, ``Learning affordances for
  categorizing objects and their properties,'' in \emph{2010 20th International
  Conference on Pattern Recognition}.\hskip 1em plus 0.5em minus 0.4em\relax
  IEEE, 2010, pp. 3089--3092.

\bibitem[de~Granville et~al.(2006)de~Granville, Southerland, and
  Fagg]{de2006learning}
C.~de~Granville, J.~Southerland, and A.~H. Fagg, ``Learning grasp affordances
  through human demonstration,'' in \emph{Proceedings of the International
  Conference on Development and Learning (ICDL’06)}, 2006.

\bibitem[Dehban et~al.(2016)Dehban, Jamone, Kampff, and
  Santos-Victor]{dehban2016denoising}
A.~Dehban, L.~Jamone, A.~R. Kampff, and J.~Santos-Victor, ``Denoising
  auto-encoders for learning of objects and tools affordances in continuous
  space,'' in \emph{2016 IEEE International Conference on Robotics and
  Automation (ICRA)}.\hskip 1em plus 0.5em minus 0.4em\relax IEEE, 2016, pp.
  4866--4871.

\bibitem[Detry et~al.(2011)Detry, Kraft, Kroemer, Bodenhagen, Peters,
  Kr{\"u}ger, and Piater]{detry2011learning}
R.~Detry, D.~Kraft, O.~Kroemer, L.~Bodenhagen, J.~Peters, N.~Kr{\"u}ger, and
  J.~Piater, ``Learning grasp affordance densities,'' \emph{Paladyn, Journal of
  Behavioral Robotics}, vol.~2, no.~1, pp. 1--17, 2011.

\bibitem[Diana et~al.(2013)Diana, de~la Croix, and
  Egerstedt]{diana2013deformable}
M.~Diana, J.-P. de~la Croix, and M.~Egerstedt, ``Deformable-medium affordances
  for interacting with multi-robot systems,'' in \emph{Intelligent Robots and
  Systems (IROS), 2013 IEEE/RSJ International Conference on}.\hskip 1em plus
  0.5em minus 0.4em\relax IEEE, 2013, pp. 5252--5257.

\bibitem[Do et~al.(2017)Do, Nguyen, Reid, Caldwell, and
  Tsagarakis]{do2017affordancenet}
T.-T. Do, A.~Nguyen, I.~Reid, D.~G. Caldwell, and N.~G. Tsagarakis,
  ``Affordancenet: An end-to-end deep learning approach for object affordance
  detection,'' \emph{arXiv preprint arXiv:1709.07326}, 2017.

\bibitem[Dogar et~al.(2007)Dogar, Cakmak, Ugur, and Sahin]{dogar2007primitive}
M.~R. Dogar, M.~Cakmak, E.~Ugur, and E.~Sahin, ``From primitive behaviors to
  goal-directed behavior using affordances,'' in \emph{2007 IEEE/RSJ
  International Conference on Intelligent Robots and Systems}.\hskip 1em plus
  0.5em minus 0.4em\relax IEEE, 2007, pp. 729--734.

\bibitem[Erkan et~al.(2010)Erkan, Kroemer, Detry, Altun, Piater, and
  Peters]{erkan2010learning}
A.~N. Erkan, O.~Kroemer, R.~Detry, Y.~Altun, J.~Piater, and J.~Peters,
  ``Learning probabilistic discriminative models of grasp affordances under
  limited supervision,'' in \emph{2010 IEEE/RSJ International Conference on
  Intelligent Robots and Systems}.\hskip 1em plus 0.5em minus 0.4em\relax IEEE,
  2010, pp. 1586--1591.

\bibitem[Fallon et~al.(2015)Fallon, Kuindersma, Karumanchi, Antone, Schneider,
  Dai, D'Arpino, Deits, DiCicco, Fourie, et~al.]{fallon2015architecture}
M.~Fallon, S.~Kuindersma, S.~Karumanchi, M.~Antone, T.~Schneider, H.~Dai, C.~P.
  D'Arpino, R.~Deits, M.~DiCicco, D.~Fourie \emph{et~al.}, ``An architecture
  for online affordance-based perception and whole-body planning,''
  \emph{Journal of Field Robotics}, vol.~32, no.~2, pp. 229--254, 2015.

\bibitem[Fitzpatrick et~al.(2003)Fitzpatrick, Metta, Natale, Rao, and
  Sandini]{fitzpatrick2003learning}
P.~Fitzpatrick, G.~Metta, L.~Natale, S.~Rao, and G.~Sandini, ``Learning about
  objects through action-initial steps towards artificial cognition,'' in
  \emph{Robotics and Automation, 2003. Proceedings. ICRA'03. IEEE International
  Conference on}, vol.~3.\hskip 1em plus 0.5em minus 0.4em\relax IEEE, 2003,
  pp. 3140--3145.

\bibitem[Fritz et~al.(2006)Fritz, Paletta, Breithaupt, Rome, and
  Dorffner]{fritz2006learning}
G.~Fritz, L.~Paletta, R.~Breithaupt, E.~Rome, and G.~Dorffner, ``Learning
  predictive features in affordance based robotic perception systems,'' in
  \emph{Intelligent Robots and Systems, 2006 IEEE/RSJ International Conference
  on}.\hskip 1em plus 0.5em minus 0.4em\relax IEEE, 2006, pp. 3642--3647.

\bibitem[Gaver(1991)]{gaver1991technology}
W.~W. Gaver, ``Technology affordances,'' in \emph{Proceedings of the SIGCHI
  conference on Human factors in computing systems}.\hskip 1em plus 0.5em minus
  0.4em\relax ACM, 1991, pp. 79--84.

\bibitem[Gibson(2014)]{gibson2014ecological}
J.~J. Gibson, \emph{The ecological approach to visual perception: classic
  edition}.\hskip 1em plus 0.5em minus 0.4em\relax Psychology Press, 2014.

\bibitem[Gijsberts et~al.(2010)Gijsberts, Tommasi, Metta, and
  Caputo]{gijsberts2010object}
A.~Gijsberts, T.~Tommasi, G.~Metta, and B.~Caputo, ``Object recognition using
  visuo-affordance maps,'' in \emph{2010 IEEE/RSJ International Conference on
  Intelligent Robots and Systems}.\hskip 1em plus 0.5em minus 0.4em\relax IEEE,
  2010, pp. 1572--1578.

\bibitem[Gon{\c{c}}alves et~al.(2014)Gon{\c{c}}alves, Abrantes, Saponaro,
  Jamone, and Bernardino]{gonccalves2014learning}
A.~Gon{\c{c}}alves, J.~Abrantes, G.~Saponaro, L.~Jamone, and A.~Bernardino,
  ``Learning intermediate object affordances: Towards the development of a tool
  concept,'' in \emph{4th International Conference on Development and Learning
  and on Epigenetic Robotics}.\hskip 1em plus 0.5em minus 0.4em\relax IEEE,
  2014, pp. 482--488.

\bibitem[Griffith et~al.(2011)Griffith, Sinapov, Sukhoy, and
  Stoytchev]{griffith2011behavior}
S.~Griffith, J.~Sinapov, V.~Sukhoy, and A.~Stoytchev, ``A behavior-grounded
  approach to forming object categories: Separating containers from
  noncontainers,'' \emph{IEEE Transactions on Autonomous Mental Development},
  vol.~4, no.~1, pp. 54--69, 2011.

\bibitem[Hart et~al.(2015)Hart, Dinh, and Hambuchen]{Hart2015TheAT}
S.~Hart, P.~Dinh, and K.~A. Hambuchen, ``The affordance template ros package
  for robot task programming,'' \emph{2015 IEEE International Conference on
  Robotics and Automation (ICRA)}, pp. 6227--6234, 2015.

\bibitem[Hermans et~al.(2011)Hermans, Rehg, and Bobick]{hermans2011affordance}
T.~Hermans, J.~M. Rehg, and A.~Bobick, ``Affordance prediction via learned
  object attributes,'' in \emph{IEEE International Conference on Robotics and
  Automation (ICRA): Workshop on Semantic Perception, Mapping, and
  Exploration}.\hskip 1em plus 0.5em minus 0.4em\relax Citeseer, 2011, pp.
  181--184.

\bibitem[Hermans et~al.(2013{\natexlab{b}})Hermans, Li, Rehg, and
  Bobick]{hermans2013learning}
T.~Hermans, F.~Li, J.~M. Rehg, and A.~F. Bobick, ``Learning contact locations
  for pushing and orienting unknown objects,'' in \emph{2013 13th IEEE-RAS
  international conference on humanoid robots (humanoids)}.\hskip 1em plus
  0.5em minus 0.4em\relax IEEE, 2013, pp. 435--442.

\bibitem[Hermans et~al.(2013{\natexlab{a}})Hermans, Rehg, and
  Bobick]{hermans2013decoupling}
T.~Hermans, J.~M. Rehg, and A.~F. Bobick, ``Decoupling behavior, perception,
  and control for autonomous learning of affordances,'' in \emph{Robotics and
  Automation (ICRA), 2013 IEEE International Conference on}.\hskip 1em plus
  0.5em minus 0.4em\relax IEEE, 2013, pp. 4989--4996.

\bibitem[Horton et~al.(2012)Horton, Chakraborty, and
  Amant]{horton2012affordances}
T.~E. Horton, A.~Chakraborty, and R.~S. Amant, ``Affordances for robots: a
  brief survey,'' \emph{AVANT. Pismo Awangardy Filozoficzno-Naukowej}, vol.~2,
  pp. 70--84, 2012.

\bibitem[Huang et~al.(2016)Huang, Bianchi, Liarokapis, and
  Sun]{huang2016recent}
Y.~Huang, M.~Bianchi, M.~Liarokapis, and Y.~Sun, ``Recent data sets on object
  manipulation: A survey,'' \emph{Big data}, vol.~4, no.~4, pp. 197--216, 2016.

\bibitem[Ivaldi et~al.(2012)Ivaldi, Lyubova, G{\'e}rardeaux-Viret, Droniou,
  Anzalone, Chetouani, Filliat, and Sigaud]{ivaldi2012perception}
S.~Ivaldi, N.~Lyubova, D.~G{\'e}rardeaux-Viret, A.~Droniou, S.~M. Anzalone,
  M.~Chetouani, D.~Filliat, and O.~Sigaud, ``Perception and human interaction
  for developmental learning of objects and affordances,'' in \emph{Humanoid
  Robots (Humanoids), 2012 12th IEEE-RAS International Conference on}.\hskip
  1em plus 0.5em minus 0.4em\relax IEEE, 2012, pp. 248--254.

\bibitem[Jamone et~al.(2018)Jamone, Ugur, Cangelosi, Fadiga, Bernardino,
  Piater, and Santos-Victor]{jamone2018affordances}
L.~Jamone, E.~Ugur, A.~Cangelosi, L.~Fadiga, A.~Bernardino, J.~Piater, and
  J.~Santos-Victor, ``Affordances in psychology, neuroscience, and robotics: A
  survey,'' \emph{IEEE Transactions on Cognitive and Developmental Systems},
  vol.~10, no.~1, pp. 4--25, 2018.

\bibitem[Jiang et~al.(2013)Jiang, Koppula, and Saxena]{jiang2013hallucinated}
Y.~Jiang, H.~Koppula, and A.~Saxena, ``Hallucinated humans as the hidden
  context for labeling 3d scenes,'' in \emph{Proceedings of the IEEE Conference
  on Computer Vision and Pattern Recognition}, 2013, pp. 2993--3000.

\bibitem[Kaiser et~al.(2014)Kaiser, Gonzalez-Aguirre, Sch{\"u}ltje, Borras,
  Vahrenkamp, and Asfour]{kaiser2014extracting}
P.~Kaiser, D.~Gonzalez-Aguirre, F.~Sch{\"u}ltje, J.~Borras, N.~Vahrenkamp, and
  T.~Asfour, ``Extracting whole-body affordances from multimodal exploration,''
  in \emph{2014 IEEE-RAS International Conference on Humanoid Robots}.\hskip
  1em plus 0.5em minus 0.4em\relax IEEE, 2014, pp. 1036--1043.

\bibitem[Kaiser et~al.(2015)Kaiser, Grotz, Aksoy, Do, Vahrenkamp, and
  Asfour]{kaiser2015validation}
P.~Kaiser, M.~Grotz, E.~E. Aksoy, M.~Do, N.~Vahrenkamp, and T.~Asfour,
  ``Validation of whole-body loco-manipulation affordances for pushability and
  liftability,'' in \emph{2015 IEEE-RAS 15th International Conference on
  Humanoid Robots (Humanoids)}.\hskip 1em plus 0.5em minus 0.4em\relax IEEE,
  2015, pp. 920--927.

\bibitem[Kaiser et~al.(2016)Kaiser, Aksoy, Grotz, and
  Asfour]{kaiser2016towards}
P.~Kaiser, E.~E. Aksoy, M.~Grotz, and T.~Asfour, ``Towards a hierarchy of
  loco-manipulation affordances,'' in \emph{2016 IEEE/RSJ International
  Conference on Intelligent Robots and Systems (IROS)}.\hskip 1em plus 0.5em
  minus 0.4em\relax IEEE, 2016, pp. 2839--2846.

\bibitem[Katz et~al.(2014)Katz, Venkatraman, Kazemi, Bagnell, and
  Stentz]{katz2014perceiving}
D.~Katz, A.~Venkatraman, M.~Kazemi, J.~A. Bagnell, and A.~Stentz, ``Perceiving,
  learning, and exploiting object affordances for autonomous pile
  manipulation,'' \emph{Autonomous Robots}, vol.~37, no.~4, pp. 369--382, 2014.

\bibitem[Kim and Sukhatme(2014)]{kim2014semantic}
D.~I. Kim and G.~S. Sukhatme, ``Semantic labeling of 3d point clouds with
  object affordance for robot manipulation,'' in \emph{Robotics and Automation
  (ICRA), 2014 IEEE International Conference on}.\hskip 1em plus 0.5em minus
  0.4em\relax Citeseer, 2014, pp. 5578--5584.

\bibitem[Kim and Sukhatme(2015)]{kim2015interactive}
------, ``Interactive affordance map building for a robotic task,'' in
  \emph{2015 IEEE/RSJ International Conference on Intelligent Robots and
  Systems (IROS)}.\hskip 1em plus 0.5em minus 0.4em\relax IEEE, 2015, pp.
  4581--4586.

\bibitem[Kjellstr{\"o}m et~al.(2011)Kjellstr{\"o}m, Romero, and
  Kragi{\'c}]{kjellstrom2011visual}
H.~Kjellstr{\"o}m, J.~Romero, and D.~Kragi{\'c}, ``Visual object-action
  recognition: Inferring object affordances from human demonstration,''
  \emph{Computer Vision and Image Understanding}, vol. 115, no.~1, pp. 81--90,
  2011.

\bibitem[Koppula and Saxena(2014)]{koppula2014physically}
H.~S. Koppula and A.~Saxena, ``Physically grounded spatio-temporal object
  affordances,'' in \emph{European Conference on Computer Vision}.\hskip 1em
  plus 0.5em minus 0.4em\relax Springer, 2014, pp. 831--847.

\bibitem[Koppula and Saxena(2016)]{koppula2016anticipating}
------, ``Anticipating human activities using object affordances for reactive
  robotic response,'' \emph{IEEE transactions on pattern analysis and machine
  intelligence}, vol.~38, no.~1, pp. 14--29, 2016.

\bibitem[Koppula et~al.(2016)Koppula, Jain, and
  Saxena]{koppula2016anticipatory}
H.~S. Koppula, A.~Jain, and A.~Saxena, ``Anticipatory planning for human-robot
  teams,'' in \emph{Experimental Robotics}.\hskip 1em plus 0.5em minus
  0.4em\relax Springer, 2016, pp. 453--470.

\bibitem[Koppula et~al.(2013)Koppula, Gupta, and Saxena]{koppula2013learning}
H.~S. Koppula, R.~Gupta, and A.~Saxena, ``Learning human activities and object
  affordances from rgb-d videos,'' \emph{The International Journal of Robotics
  Research}, vol.~32, no.~8, pp. 951--970, 2013.

\bibitem[Kostavelis et~al.(2012)Kostavelis, Nalpantidis, and
  Gasteratos]{kostavelis2012collision}
I.~Kostavelis, L.~Nalpantidis, and A.~Gasteratos, ``Collision risk assessment
  for autonomous robots by offline traversability learning,'' \emph{Robotics
  and Autonomous Systems}, vol.~60, no.~11, pp. 1367--1376, 2012.

\bibitem[Kraft et~al.(2009)Kraft, Detry, Pugeault, Ba{\c{s}}eski, Piater, and
  Kr{\"u}ger]{kraft2009learning}
D.~Kraft, R.~Detry, N.~Pugeault, E.~Ba{\c{s}}eski, J.~Piater, and
  N.~Kr{\"u}ger, ``Learning objects and grasp affordances through autonomous
  exploration,'' in \emph{International Conference on Computer Vision
  Systems}.\hskip 1em plus 0.5em minus 0.4em\relax Springer, 2009, pp.
  235--244.

\bibitem[Kroemer and Peters(2011)]{kroemer2011flexible}
O.~Kroemer and J.~Peters, ``A flexible hybrid framework for modeling complex
  manipulation tasks,'' in \emph{2011 IEEE International Conference on Robotics
  and Automation}.\hskip 1em plus 0.5em minus 0.4em\relax IEEE, 2011, pp.
  1856--1861.

\bibitem[Kroemer et~al.(2012)Kroemer, Ugur, Oztop, and
  Peters]{kroemer2012kernel}
O.~Kroemer, E.~Ugur, E.~Oztop, and J.~Peters, ``A kernel-based approach to
  direct action perception,'' in \emph{Robotics and Automation (ICRA), 2012
  IEEE International Conference on}.\hskip 1em plus 0.5em minus 0.4em\relax
  IEEE, 2012, pp. 2605--2610.

\bibitem[Kr{\"u}ger et~al.(2011)Kr{\"u}ger, Geib, Piater, Petrick, Steedman,
  W{\"o}rg{\"o}tter, Ude, Asfour, Kraft, Omr{\v{c}}en,
  et~al.]{kruger2011object}
N.~Kr{\"u}ger, C.~Geib, J.~Piater, R.~Petrick, M.~Steedman,
  F.~W{\"o}rg{\"o}tter, A.~Ude, T.~Asfour, D.~Kraft, D.~Omr{\v{c}}en
  \emph{et~al.}, ``Object--action complexes: Grounded abstractions of
  sensory--motor processes,'' \emph{Robotics and Autonomous Systems}, vol.~59,
  no.~10, pp. 740--757, 2011.

\bibitem[Lewis et~al.(2005)Lewis, Lee, and Patla]{lewis2005foot}
M.~A. Lewis, H.-K. Lee, and A.~Patla, ``Foot placement selection using
  non-geometric visual properties,'' \emph{The International Journal of
  Robotics Research}, vol.~24, no.~7, pp. 553--561, 2005.

\bibitem[Liu et~al.(2018)Liu, Freeman, Tenenbaum, and Wu]{liu2018physical}
Z.~Liu, W.~T. Freeman, J.~B. Tenenbaum, and J.~Wu, ``Physical primitive
  decomposition,'' \emph{arXiv preprint arXiv:1809.05070}, 2018.

\bibitem[Lopes et~al.(2007)Lopes, Melo, and Montesano]{lopes2007affordance}
M.~Lopes, F.~S. Melo, and L.~Montesano, ``Affordance-based imitation learning
  in robots,'' in \emph{2007 IEEE/RSJ International Conference on Intelligent
  Robots and Systems}.\hskip 1em plus 0.5em minus 0.4em\relax IEEE, 2007, pp.
  1015--1021.

\bibitem[Mar et~al.(2015)Mar, Tikhanoff, Metta, and Natale]{mar2015multi}
T.~Mar, V.~Tikhanoff, G.~Metta, and L.~Natale, ``Multi-model approach based on
  3d functional features for tool affordance learning in robotics,'' in
  \emph{Humanoid Robots (Humanoids), 2015 IEEE-RAS 15th International
  Conference on}.\hskip 1em plus 0.5em minus 0.4em\relax IEEE, 2015, pp.
  482--489.

\bibitem[McGrenere and Ho(2000)]{mcgrenere2000affordances}
J.~McGrenere and W.~Ho, ``Affordances: Clarifying and evolving a concept,'' in
  \emph{Graphics interface}, vol. 2000, 2000, pp. 179--186.

\bibitem[Min et~al.(2016)Min, Yi, Luo, Zhu, and Bi]{min2016affordance}
H.~Min, C.~Yi, R.~Luo, J.~Zhu, and S.~Bi, ``Affordance research in
  developmental robotics: a survey,'' \emph{IEEE Transactions on Cognitive and
  Developmental Systems}, vol.~8, no.~4, pp. 237--255, 2016.

\bibitem[Moldovan and De~Raedt(2014)]{moldovan2014occluded}
B.~Moldovan and L.~De~Raedt, ``Occluded object search by relational
  affordances,'' in \emph{Robotics and Automation (ICRA), 2014 IEEE
  International Conference on}.\hskip 1em plus 0.5em minus 0.4em\relax IEEE,
  2014, pp. 169--174.

\bibitem[Moldovan et~al.(2012)Moldovan, Moreno, van Otterlo, Santos-Victor, and
  De~Raedt]{moldovan2012learning}
B.~Moldovan, P.~Moreno, M.~van Otterlo, J.~Santos-Victor, and L.~De~Raedt,
  ``Learning relational affordance models for robots in multi-object
  manipulation tasks,'' in \emph{Robotics and Automation (ICRA), 2012 IEEE
  International Conference on}.\hskip 1em plus 0.5em minus 0.4em\relax IEEE,
  2012, pp. 4373--4378.

\bibitem[Montesano and Lopes(2009)]{montesano2009learning}
L.~Montesano and M.~Lopes, ``Learning grasping affordances from local visual
  descriptors,'' in \emph{Development and Learning, 2009. ICDL 2009. IEEE 8th
  International Conference on}.\hskip 1em plus 0.5em minus 0.4em\relax IEEE,
  2009, pp. 1--6.

\bibitem[Montesano et~al.(2007{\natexlab{a}})Montesano, Lopes, Bernardino, and
  Santos-Victor]{montesano2007affordances}
L.~Montesano, M.~Lopes, A.~Bernardino, and J.~Santos-Victor, ``Affordances,
  development and imitation,'' in \emph{Development and Learning, 2007. ICDL
  2007. IEEE 6th International Conference on}.\hskip 1em plus 0.5em minus
  0.4em\relax IEEE, 2007, pp. 270--275.

\bibitem[Montesano et~al.(2007{\natexlab{b}})Montesano, Lopes, Bernardino, and
  Santos-Victor]{montesano2007modeling}
------, ``Modeling affordances using bayesian networks,'' in \emph{Intelligent
  Robots and Systems, 2007. IROS 2007. IEEE/RSJ International Conference
  on}.\hskip 1em plus 0.5em minus 0.4em\relax IEEE, 2007, pp. 4102--4107.

\bibitem[Montesano et~al.(2008)Montesano, Lopes, Bernardino, and
  Santos-Victor]{montesano2008learning}
------, ``Learning object affordances: from sensory--motor coordination to
  imitation,'' \emph{IEEE Transactions on Robotics}, vol.~24, no.~1, pp.
  15--26, 2008.

\bibitem[Myers et~al.(2015)Myers, Teo, Ferm{\"u}ller, and
  Aloimonos]{myers2015affordance}
A.~Myers, C.~L. Teo, C.~Ferm{\"u}ller, and Y.~Aloimonos, ``Affordance detection
  of tool parts from geometric features,'' in \emph{2015 IEEE International
  Conference on Robotics and Automation (ICRA)}.\hskip 1em plus 0.5em minus
  0.4em\relax IEEE, 2015, pp. 1374--1381.

\bibitem[Nguyen et~al.(2017)Nguyen, Kanoulas, Caldwell, and
  Tsagarakis]{nguyen2017object}
A.~Nguyen, D.~Kanoulas, D.~G. Caldwell, and N.~G. Tsagarakis, ``Object-based
  affordances detection with convolutional neural networks and dense
  conditional random fields,'' in \emph{Intelligent Robots and Systems (IROS),
  2017 IEEE/RSJ International Conference on}.\hskip 1em plus 0.5em minus
  0.4em\relax IEEE, 2017, pp. 5908--5915.

\bibitem[Nishide et~al.(2009)Nishide, Nakagawa, Ogata, Tani, Takahashi, and
  Okuno]{nishide2009modeling}
S.~Nishide, T.~Nakagawa, T.~Ogata, J.~Tani, T.~Takahashi, and H.~G. Okuno,
  ``Modeling tool-body assimilation using second-order recurrent neural
  network,'' in \emph{2009 IEEE/RSJ International Conference on Intelligent
  Robots and Systems}.\hskip 1em plus 0.5em minus 0.4em\relax IEEE, 2009, pp.
  5376--5381.

\bibitem[Norman(1988)]{norman1988psychology}
D.~A. Norman, ``The psychology of everyday things.(the design of everyday
  things),'' 1988.

\bibitem[Omr{\v{c}}en et~al.(2009)Omr{\v{c}}en, B{\"o}ge, Asfour, Ude, and
  Dillmann]{omrvcen2009autonomous}
D.~Omr{\v{c}}en, C.~B{\"o}ge, T.~Asfour, A.~Ude, and R.~Dillmann, ``Autonomous
  acquisition of pushing actions to support object grasping with a humanoid
  robot,'' in \emph{2009 9th IEEE-RAS International Conference on Humanoid
  Robots}.\hskip 1em plus 0.5em minus 0.4em\relax IEEE, 2009, pp. 277--283.

\bibitem[Pandey and Alami(2013)]{pandey2013affordance}
A.~K. Pandey and R.~Alami, ``Affordance graph: A framework to encode
  perspective taking and effort based affordances for day-to-day human-robot
  interaction,'' in \emph{Intelligent Robots and Systems (IROS), 2013 IEEE/RSJ
  International Conference on}.\hskip 1em plus 0.5em minus 0.4em\relax IEEE,
  2013, pp. 2180--2187.

\bibitem[Pieropan et~al.(2014)Pieropan, Ek, and
  Kjellstr{\"o}m]{pieropan2014recognizing}
A.~Pieropan, C.~H. Ek, and H.~Kjellstr{\"o}m, ``Recognizing object affordances
  in terms of spatio-temporal object-object relationships,'' in
  \emph{International Conference on Humanoid Robots, November 18-20th 2014,
  Madrid, Spain}.\hskip 1em plus 0.5em minus 0.4em\relax IEEE conference
  proceedings, 2014, pp. 52--58.

\bibitem[Price et~al.(2016)Price, Balakirsky, Bobick, and
  Christensen]{price2016affordance}
A.~Price, S.~Balakirsky, A.~Bobick, and H.~Christensen, ``Affordance-feasible
  planning with manipulator wrench spaces,'' in \emph{2016 IEEE International
  Conference on Robotics and Automation (ICRA)}.\hskip 1em plus 0.5em minus
  0.4em\relax IEEE, 2016, pp. 3979--3986.

\bibitem[Ridge and Ude(2013)]{ridge2013action}
B.~Ridge and A.~Ude, ``Action-grounded push affordance bootstrapping of unknown
  objects,'' in \emph{2013 IEEE/RSJ International Conference on Intelligent
  Robots and Systems}.\hskip 1em plus 0.5em minus 0.4em\relax IEEE, 2013, pp.
  2791--2798.

\bibitem[Ruiz and Mayol-Cuevas(2018)]{ruiz2018can}
E.~Ruiz and W.~Mayol-Cuevas, ``Where can i do this? geometric affordances from
  a single example with the interaction tensor,'' in \emph{f}.\hskip 1em plus
  0.5em minus 0.4em\relax IEEE, 2018, pp. 2192--2199.

\bibitem[{\c{S}}ahin et~al.(2007){\c{S}}ahin, {\c{C}}akmak, Do{\u{g}}ar,
  U{\u{g}}ur, and {\"U}{\c{c}}oluk]{csahin2007afford}
E.~{\c{S}}ahin, M.~{\c{C}}akmak, M.~R. Do{\u{g}}ar, E.~U{\u{g}}ur, and
  G.~{\"U}{\c{c}}oluk, ``To afford or not to afford: A new formalization of
  affordances toward affordance-based robot control,'' \emph{Adaptive
  Behavior}, vol.~15, no.~4, pp. 447--472, 2007.

\bibitem[Saxena et~al.(2014)Saxena, Jain, Sener, Jami, Misra, and
  Koppula]{saxena2014robobrain}
A.~Saxena, A.~Jain, O.~Sener, A.~Jami, D.~K. Misra, and H.~S. Koppula,
  ``Robobrain: Large-scale knowledge engine for robots,'' \emph{arXiv preprint
  arXiv:1412.0691}, 2014.

\bibitem[Song et~al.(2010)Song, Huebner, Kyrki, and Kragic]{song2010learning}
D.~Song, K.~Huebner, V.~Kyrki, and D.~Kragic, ``Learning task constraints for
  robot grasping using graphical models,'' in \emph{Intelligent Robots and
  Systems (IROS), 2010 IEEE/RSJ International Conference on}.\hskip 1em plus
  0.5em minus 0.4em\relax IEEE, 2010, pp. 1579--1585.

\bibitem[Song et~al.(2011)Song, Ek, Huebner, and Kragic]{song2011embodiment}
D.~Song, C.~H. Ek, K.~Huebner, and D.~Kragic, ``Embodiment-specific
  representation of robot grasping using graphical models and latent-space
  discretization,'' in \emph{2011 IEEE/RSJ International Conference on
  Intelligent Robots and Systems}.\hskip 1em plus 0.5em minus 0.4em\relax IEEE,
  2011, pp. 980--986.

\bibitem[Song et~al.(2013)Song, Kyriazis, Oikonomidis, Papazov, Argyros,
  Burschka, and Kragic]{song2013predicting}
D.~Song, N.~Kyriazis, I.~Oikonomidis, C.~Papazov, A.~Argyros, D.~Burschka, and
  D.~Kragic, ``Predicting human intention in visual observations of hand/object
  interactions,'' in \emph{2013 IEEE International Conference on Robotics and
  Automation}.\hskip 1em plus 0.5em minus 0.4em\relax IEEE, 2013, pp.
  1608--1615.

\bibitem[Song et~al.(2015{\natexlab{a}})Song, Ek, Huebner, and
  Kragic]{song2015task}
D.~Song, C.~H. Ek, K.~Huebner, and D.~Kragic, ``Task-based robot grasp planning
  using probabilistic inference,'' \emph{IEEE transactions on robotics},
  vol.~31, no.~3, pp. 546--561, 2015.

\bibitem[Song et~al.(2015{\natexlab{b}})Song, Fritz, Goehring, and
  Darrell]{song2015learning}
H.~O. Song, M.~Fritz, D.~Goehring, and T.~Darrell, ``Learning to detect visual
  grasp affordance,'' \emph{IEEE Transactions on Automation Science and
  Engineering}, vol.~13, no.~2, pp. 798--809, 2015.

\bibitem[Stark et~al.(2008)Stark, Lies, Zillich, Wyatt, and
  Schiele]{stark2008functional}
M.~Stark, P.~Lies, M.~Zillich, J.~Wyatt, and B.~Schiele, ``Functional object
  class detection based on learned affordance cues,'' in \emph{International
  conference on computer vision systems}.\hskip 1em plus 0.5em minus
  0.4em\relax Springer, 2008, pp. 435--444.

\bibitem[Steedman(2002)]{steedman2002formalizing}
M.~Steedman, ``Formalizing affordance,'' in \emph{Proceedings of the Annual
  Meeting of the Cognitive Science Society}, vol.~24, no.~24, 2002.

\bibitem[Stoffregen(2003)]{stoffregen2003affordances}
T.~A. Stoffregen, ``Affordances as properties of the animal-environment
  system,'' \emph{Ecological psychology}, vol.~15, no.~2, pp. 115--134, 2003.

\bibitem[Stoytchev(2005)]{stoytchev2005toward}
A.~Stoytchev, ``Behavior-grounded representation of tool affordances,'' in
  \emph{Proceedings of the 2005 ieee international conference on robotics and
  automation}.\hskip 1em plus 0.5em minus 0.4em\relax IEEE, 2005, pp.
  3060--3065.

\bibitem[Sun et~al.(2010)Sun, Moore, Bobick, and Rehg]{sun2010learning}
J.~Sun, J.~L. Moore, A.~Bobick, and J.~M. Rehg, ``Learning visual object
  categories for robot affordance prediction,'' \emph{The International Journal
  of Robotics Research}, vol.~29, no. 2-3, pp. 174--197, 2010.

\bibitem[Sun et~al.(2014)Sun, Ren, and Lin]{sun2014object}
Y.~Sun, S.~Ren, and Y.~Lin, ``Object--object interaction affordance learning,''
  \emph{Robotics and Autonomous Systems}, vol.~62, no.~4, pp. 487--496, 2014.

\bibitem[Sweeney and Grupen(2007)]{sweeney2007model}
J.~D. Sweeney and R.~Grupen, ``A model of shared grasp affordances from
  demonstration,'' in \emph{2007 7th IEEE-RAS International Conference on
  Humanoid Robots}.\hskip 1em plus 0.5em minus 0.4em\relax IEEE, 2007, pp.
  27--35.

\bibitem[Szedmak et~al.(2014)Szedmak, Ugur, and Piater]{szedmak2014knowledge}
S.~Szedmak, E.~Ugur, and J.~Piater, ``Knowledge propagation and relation
  learning for predicting action effects,'' in \emph{2014 IEEE/RSJ
  International Conference on Intelligent Robots and Systems}.\hskip 1em plus
  0.5em minus 0.4em\relax IEEE, 2014, pp. 623--629.

\bibitem[Thomaz and Cakmak(2009)]{thomaz2009learning}
A.~L. Thomaz and M.~Cakmak, ``Learning about objects with human teachers,'' in
  \emph{Proceedings of the 4th ACM/IEEE international conference on Human robot
  interaction}.\hskip 1em plus 0.5em minus 0.4em\relax ACM, 2009, pp. 15--22.

\bibitem[Tikhanoff et~al.(2013)Tikhanoff, Pattacini, Natale, and
  Metta]{tikhanoff2013exploring}
V.~Tikhanoff, U.~Pattacini, L.~Natale, and G.~Metta, ``Exploring affordances
  and tool use on the icub,'' in \emph{Humanoid Robots (Humanoids), 2013 13th
  IEEE-RAS International Conference on}.\hskip 1em plus 0.5em minus 0.4em\relax
  IEEE, 2013, pp. 130--137.

\bibitem[Turvey(1992)]{turvey1992affordances}
M.~T. Turvey, ``Affordances and prospective control: An outline of the
  ontology,'' \emph{Ecological psychology}, vol.~4, no.~3, pp. 173--187, 1992.

\bibitem[Ugur and Piater(2015)]{ugur2015bottom}
E.~Ugur and J.~Piater, ``Bottom-up learning of object categories, action
  effects and logical rules: From continuous manipulative exploration to
  symbolic planning,'' in \emph{2015 IEEE International Conference on Robotics
  and Automation (ICRA)}.\hskip 1em plus 0.5em minus 0.4em\relax IEEE, 2015,
  pp. 2627--2633.

\bibitem[Ugur et~al.(2007{\natexlab{b}})Ugur, Dogar, Cakmak, and
  Sahin]{ugur2007curiosity}
E.~Ugur, M.~R. Dogar, M.~Cakmak, and E.~Sahin, ``Curiosity-driven learning of
  traversability affordance on a mobile robot,'' in \emph{Development and
  Learning, 2007. ICDL 2007. IEEE 6th International Conference}.\hskip 1em plus
  0.5em minus 0.4em\relax IEEE, 2007, pp. 13--18.

\bibitem[Ugur et~al.(2007{\natexlab{a}})Ugur, Dogar, Cakmak, and
  Sahin]{ugur2007learning}
------, ``The learning and use of traversability affordance using range images
  on a mobile robot,'' in \emph{Robotics and Automation, 2007 IEEE
  International Conference}.\hskip 1em plus 0.5em minus 0.4em\relax IEEE, 2007,
  pp. 1721--1726.

\bibitem[Ugur et~al.(2009)Ugur, Sahin, and Oztop]{ugur2009affordance}
E.~Ugur, E.~Sahin, and E.~Oztop, ``Affordance learning from range data for
  multi-step planning,'' in \emph{Int. Conf. on Epigenetic Robotics}, 2009.

\bibitem[Ugur et~al.(2011{\natexlab{b}})Ugur, Oztop, and Sahin]{ugur2011goal}
E.~Ugur, E.~Oztop, and E.~Sahin, ``Goal emulation and planning in perceptual
  space using learned affordances,'' \emph{Robotics and Autonomous Systems},
  vol.~59, no. 7-8, pp. 580--595, 2011.

\bibitem[Ugur et~al.(2011{\natexlab{a}})Ugur, {\c{S}}ahin, and
  Oztop]{ugur2011unsupervised}
E.~Ugur, E.~{\c{S}}ahin, and E.~Oztop, ``Unsupervised learning of object
  affordances for planning in a mobile manipulation platform,'' in
  \emph{Robotics and Automation (ICRA), 2011 IEEE International Conference
  on}.\hskip 1em plus 0.5em minus 0.4em\relax IEEE, 2011, pp. 4312--4317.

\bibitem[Ugur et~al.(2015)Ugur, Nagai, Sahin, and Oztop]{ugur2015staged}
E.~Ugur, Y.~Nagai, E.~Sahin, and E.~Oztop, ``Staged development of robot
  skills: Behavior formation, affordance learning and imitation with
  motionese,'' \emph{IEEE Transactions on Autonomous Mental Development},
  vol.~7, no.~2, pp. 119--139, 2015.

\bibitem[Varadarajan and Vincze(2012)]{varadarajan2012afrob}
K.~M. Varadarajan and M.~Vincze, ``Afrob: The affordance network ontology for
  robots,'' in \emph{2012 IEEE/RSJ International Conference on Intelligent
  Robots and Systems}.\hskip 1em plus 0.5em minus 0.4em\relax IEEE, 2012, pp.
  1343--1350.

\bibitem[Wang et~al.(2013)Wang, Hindriks, and Babuska]{wang2013robot}
C.~Wang, K.~V. Hindriks, and R.~Babuska, ``Robot learning and use of
  affordances in goal-directed tasks,'' in \emph{Intelligent Robots and Systems
  (IROS), 2013 IEEE/RSJ International Conference on}.\hskip 1em plus 0.5em
  minus 0.4em\relax IEEE, 2013, pp. 2288--2294.

\bibitem[Ye et~al.(2017)Ye, Yang, Mao, Ferm{\"u}ller, and Aloimonos]{ye2017can}
C.~Ye, Y.~Yang, R.~Mao, C.~Ferm{\"u}ller, and Y.~Aloimonos, ``What can i do
  around here? deep functional scene understanding for cognitive robots,'' in
  \emph{2017 IEEE International Conference on Robotics and Automation
  (ICRA)}.\hskip 1em plus 0.5em minus 0.4em\relax IEEE, 2017, pp. 4604--4611.

\bibitem[Zech et~al.(2017)Zech, Haller, Lakani, Ridge, Ugur, and
  Piater]{zech2017computational}
P.~Zech, S.~Haller, S.~R. Lakani, B.~Ridge, E.~Ugur, and J.~Piater,
  ``Computational models of affordance in robotics: a taxonomy and systematic
  classification,'' \emph{Adaptive Behavior}, vol.~25, no.~5, pp. 235--271,
  2017.

\bibitem[Zhu et~al.(2014)Zhu, Fathi, and Fei-Fei]{zhu2014reasoning}
Y.~Zhu, A.~Fathi, and L.~Fei-Fei, ``Reasoning about object affordances in a
  knowledge base representation,'' in \emph{European conference on computer
  vision}.\hskip 1em plus 0.5em minus 0.4em\relax Springer, 2014, pp. 408--424.

\end{thebibliography}
